\documentclass[lettersize,journal]{IEEEtran}
\usepackage{amsmath,amsfonts}
\usepackage{algorithmic}
\usepackage{algorithm}
\usepackage{array}
\usepackage[caption=false,font=normalsize,labelfont=sf,textfont=sf]{subfig}
\usepackage{textcomp}
\usepackage{stfloats}
\usepackage{url}
\usepackage{verbatim}
\usepackage{graphicx}
\usepackage{cite}
\usepackage{orcidlink}
\usepackage{booktabs}
\usepackage{multirow}
\usepackage{caption}

\usepackage{bm}
\usepackage{amsmath}
\usepackage{amssymb}
\usepackage{multirow}
\usepackage{graphicx}
\usepackage{subcaption}
\usepackage{cleveref}
\usepackage{siunitx}
\usepackage{mathrsfs}

\begin{document}

\title{Neuro-Symbolic Operator for Interpretable and Generalizable Characterization of Complex Piezoelectric Systems}

\author{Abhishek Chandra \orcidlink{0000-0003-2319-6221},~\IEEEmembership{Graduate Student Member,~IEEE}, Taniya Kapoor \orcidlink{0000-0002-6361-446X}, \\ Mitrofan Curti \orcidlink{0000-0002-0084-4372},~\IEEEmembership{Member,~IEEE}, Koen Tiels \orcidlink{0000-0001-9279-110X},~\IEEEmembership{Member,~IEEE}, Elena A. Lomonova \orcidlink{0000-0002-2515-1441},~\IEEEmembership{Senior Member,~IEEE}
\thanks{This work has been submitted to the IEEE for possible publication. Copyright may be transferred without notice, after which this version may no longer be accessible. (Corresponding author: Abhishek Chandra)}
\thanks{A. Chandra, M. Curti, E. A. Lomonova are with the Department of Electrical Engineering, Eindhoven University of Technology, The Netherlands. (e-mail: a.chandra@tue.nl; m.curti@tue.nl; e.lomonova@tue.nl).}
\thanks{T. Kapoor is with the Department of Engineering Structures, Delft University of Technology, The Netherlands. (e-mail: t.kapoor@tudelft.nl).}
\thanks{K. Tiels is with the Department of Mechanical Engineering, Eindhoven University of Technology, The Netherlands. (e-mail: k.tiels@tue.nl).}
}

\markboth{}%
{Shell \MakeLowercase{\textit{Chandra et al.}}: NSO for Interpretable and Generalizable Characterization of Complex Piezoelectric Systems}

\maketitle

\begin{abstract}
Complex piezoelectric systems are foundational in industrial applications. Their performance, however, is challenged by the nonlinear voltage-displacement hysteretic relationships. Efficient characterization methods are, therefore, essential for reliable design, monitoring, and maintenance. Recently proposed neural operator methods serve as surrogates for system characterization but face two pressing issues: interpretability and generalizability. State-of-the-art (SOTA) neural operators are black-boxes, providing little insight into the learned operator. Additionally, generalizing them to novel voltages and predicting displacement profiles beyond the training domain is challenging, limiting their practical use. To address these limitations, this paper proposes a neuro-symbolic operator (NSO) framework that derives the analytical operators governing hysteretic relationships. NSO first learns a Fourier neural operator mapping voltage fields to displacement profiles, followed by a library-based sparse model discovery method, generating white-box parsimonious models governing the underlying hysteresis. These models enable accurate and interpretable prediction of displacement profiles across varying and out-of-distribution voltage fields, facilitating generalizability. The potential of NSO is demonstrated by accurately predicting voltage-displacement hysteresis, including butterfly-shaped relationships. Moreover, NSO predicts displacement profiles even for noisy and low-fidelity voltage data, emphasizing its robustness. The results highlight the advantages of NSO compared to SOTA neural operators and model discovery methods on several evaluation metrics. Consequently, NSO contributes to characterizing complex piezoelectric systems while improving the interpretability and generalizability of neural operators, essential for design, monitoring, maintenance, and other real-world scenarios.
\end{abstract}

\begin{IEEEkeywords}
Neuro-symbolic, neural operators, hysteresis, piezoelectric.
\end{IEEEkeywords}

\section{Introduction}
\IEEEPARstart{C}{omplex} piezoelectric systems, such as sensors and actuators, are essential in diverse engineering industries, including process control \cite{qian2020piezoelectric}, structures \cite{chen2019piezoelectric}, aerospace \cite{elahi2020review}, and biomedical \cite{wang2025applications}, among others \cite{sekhar2023review}. These systems leverage piezoelectric materials that convert mechanical energy into electrical energy, known as the direct piezoelectric effect, or vice versa, known as the inverse piezoelectric effect. This paper focuses on the inverse piezoelectric effect\textemdash where the applied voltage induces mechanical displacement\textemdash which is fundamental for tasks like nanopositioning and automation. Given their widespread industrial application, it is essential to characterize and predict the performance of these systems. However, characterizing these systems is challenging due to the complex, history-dependent, and nonlinear behavior inherent in piezoelectric materials \cite{damjanovic2006hysteresis}.

For instance, a piezoelectric actuator under a varying voltage field does not exhibit a linearly varying displacement profile. Instead, the actuator exhibits a nonlinear history-dependent hysteretic relationship \cite{zhang2018adaptive}. Hysteresis makes it challenging to accurately identify and predict the actuator positioning and induces energy losses during device operation \cite{robert2001piezoelectric}. Prior knowledge of the actuator position and energy losses under a varying arbitrary voltage field is vital for industry operations, signifying the necessity of robust methodologies that accurately characterize the material-specific hysteresis relationship. These relationships are traditionally characterized through phenomenological models like the Preisach \cite{preisach1935magnetische} and Bouc--Wen-based models \cite{ottosen2005mechanics}. Although successful, these models' stringent data requirements and computational costs have led to developing deep learning-based methods for modeling hysteresis \cite{chandra2024magnetic, hassani2014survey}.

As the underlying characteristics of hysteresis are history-dependent, it forms a conceptual overlap with recurrent neural networks \cite{lecun2015deep}. Due to this overlap, several recurrent neural architectures like long short-term memory \cite{hochreiter1997long} and gated recurrent unit \cite{cho2014learning} have been utilized to model hysteresis\cite{zhang2024adaptive}, considering the previous system states to predict future states \cite{chandra2025generalizable}. Although these architectures provide accurate predictions within the training domain, they face challenges in predicting the displacement profiles for novel voltage fields, which are not used while training the model \cite{chandra2024characterizing}. This challenge of predicting displacement profiles for novel input voltage fields, known as generalization, limits the applicability of the trained model in real-world scenarios where the deployment conditions differ from those while training the model. 

A possible way of mitigating the challenge of generalization and predicting the displacement profiles for novel input voltage fields is by mapping the voltage and displacement function spaces. This mapping would enable predicting displacement profiles for novel input voltage fields, and represent the underlying operator governing the hysteresis phenomenon in the material. One such class of neural network-based methods, which learns the underlying operator, mapping function spaces, are neural operator architectures \cite{kovachki2023neural, azizzadenesheli2024neural}. Neural operators learn a neural surrogate of the underlying black-box hysteresis phenomenon and facilitate inferring for novel input functions in fractions of seconds \cite{chandra2024characterizing}.

However, state-of-the-art (SOTA) neural operators face limitations in terms of interpretability and generalizability. While these models can efficiently predict displacement profiles for previously unseen voltage fields within the training domain \cite{lu2021learning, lifourier, raonic2023convolutional}, they offer limited insight into the underlying dynamics they have learned and are not interpretable. In particular, SOTA neural operator frameworks cannot recover an analytical form of the governing equations inferred from data. The availability of interpretable, white-box models enables engineers to analyze better, optimize, and monitor system performance, ultimately contributing to improved maintenance, monitoring, and the design of more efficient systems \cite{chandra2023discovery, lin2022identification}. This work seeks to advance neural operator methodologies by enabling the analytical inference of the learned governing models, thereby providing an understanding of the physical laws learned by the neural operator.

Additionally, SOTA neural operators require substantial data from each training domain and fail to generalize the learned operator to another domain, as shown in this work. Obtaining labeled data samples for different input fields is often challenging in industrial settings \cite{wang2021learning}, where generating the measurement data are challenging and expensive, or the corresponding simulation is computationally expensive \cite{kapoor2025beyond}. Moreover, neural operators require data for multiple input-output function pairs, possibly hundreds or thousands of functions, to learn the underlying operator accurately \cite{viswanath2023neural}. Obtaining high-fidelity data for hundreds or thousands of functions for each input field under consideration makes it practically challenging to employ neural operators in modeling piezoelectric materials in real-world settings.

Hence, there is a need for interpretable neural operators that are generalizable to different and diverse input fields, even when trained on a particular input field. In response, this paper proposes a neuro-symbolic operator (NSO), which first employs a  Fourier neural operator \cite{lifourier}, on a specific input field to learn the underlying operator in black-box form. Subsequently, the learned operator is used to generate predictions in the trained input field. NSO further uses these predictions to train a library-based model discovery method \cite{brunton2016discovering} to discover the governing operator in a white-box form. This white-box model learned by NSO is generalizable to different input fields, not used to train the models, owing to the generalizable characteristics of symbolic models \cite{cranmer2020discovering}. 

The efficacy of the proposed NSO framework is demonstrated through learning complex hysteresis behaviors between voltage fields and corresponding displacement profiles. In particular, the framework's performance is tested on capturing differently-shaped hysteresis loops, including butterfly-shaped hysteresis loops \cite{drinvcic2011some}, illustrating its broad applicability. The experimental evaluation includes data from multiple differential hysteresis models to showcase learning of diverse hysteretic dynamics. Experiments showcase model-agnosticism and non-biased characteristics of NSO towards the training data. Additional ablation studies further assess the performance under noisy displacement fields and generalization to high-fidelity data even when trained on low-fidelity datasets, showcasing the robustness of the proposed method.

The main contributions of this paper are as follows: \mbox{1.) This paper proposes} a novel operator learning framework, NSO, and showcases its applicability in modeling piezoelectric hysteresis; 2.) NSO learns interpretable differential models addressing the black-box nature of SOTA neural operators; 3.) NSO is demonstrated to be generalizable across a wide range of input fields and hysteresis models—a limitation in SOTA neural operators; 4.) NSO is evaluated on extensive experiments, including complex butterfly-shaped hysteresis, showcasing its applicability to unseen inputs and cross-model generalization; 5.) Experiments showcase that NSO is robust and provides advantages over both SOTA neural operators and traditional model discovery methods on several metrics, even for noisy and low-fidelity data.

The rest of the paper is organized as follows. Section~\ref{sec2} introduces the concepts of piezoelectric hysteresis modeling, neural operators, and model discovery methods. Section~\ref{sec3} formulates the problem tackled in this paper. Section~\ref{sec4} introduces the proposed NSO framework. Section~\ref{sec5} presents the numerical experiments to show the performance of NSO. Section~\ref{sec6} discusses the obtained results and compares neural operators on various characteristics. The main conclusions drawn from this study are collated in Section~\ref{sec7}.

\section{Preliminaries}
\label{sec2}
This section presents the concepts of piezoelectric hysteresis modeling, operator learning, and data-driven model discovery.

\subsection{Hysteresis Modeling for Piezoelectric Material}
Hysteresis in piezoelectric materials, such as actuators, is characterized by a nonlinear, history-dependent relationship between the applied voltage \( v(t) \) and the resulting displacement \( d(t)\) \cite{damjanovic2006hysteresis}. Here $v(t)$, $d(t)$ are temporal signals defined over a finite time interval $t\! \in\! [0, \mathrm{T}]$. The piezoelectric material produces distinct displacement profiles when subjected to different voltage fields. Hence, characterizing them necessitates modeling displacement as a functional of the voltage field \cite{damjanovic2006hysteresis}, i.e., \( d(t)\! =\! \mathcal{P}[v](t) \), where \( \mathcal{P} \) is a nonlinear, causal operator defined on an appropriate function space. 

Specifically, let \( v\! \in\! \mathcal{V}\! \subset\! C([0, \mathrm{T}]; \mathbb{R}) \) denote the space of admissible input voltage signals, where \(C([0, \mathrm{T}]; \mathbb{R})\) denotes the space of continuous real-valued functions on the interval \([0, \mathrm{T}]\). Then the goal is to identify an operator \( \mathcal{P}\! :\! \mathcal{V}\! \to\! C([0, \mathrm{T}]; \mathbb{R}) \) satisfying $d(t) = \mathcal{P}[v](t), \forall t\! \in\! [0, \mathrm{T}]$. This operator $\mathcal{P}$ is path-dependent such that \( d(t) \) depends on the entire history of the voltage field \( \{v(s) : 0 \leq s \leq t\} \). 

\subsection{Neural Operators}
Neural operators learn the mapping between infinite-dimensional function spaces \cite{kovachki2023neural}. Unlike traditional neural networks that map between finite-dimensional tensors, neural operators learn function-to-function mappings. Given an input function \( a \in \mathcal{A} \) defined on a domain \( \Omega \subset \mathbb{R} \), and an output function \( u \in \mathcal{U} \), the task is to learn an operator
$\mathcal{G}:\! \mathcal{A}\! \to\! \mathcal{U}$, such that $u\! =\! \mathcal{G}(a)$. Here \( \mathcal{A} \) and \( \mathcal{U} \) are infinite-dimensional function spaces, such as \( L^2(\Omega) \) or \( C(\Omega) \). Several neural operators have been proposed, including the deep operator network (DON) \cite{lu2021learning}, Fourier neural operator (FNO) \cite{lifourier}, and convolutional neural operator (CNO) \cite{raonic2023convolutional}, among others. 

Despite architectural differences, these neural operators, $\mathcal{N}$, aim to approximate the unknown operator \( \mathcal{G} \) using data pairs \( \{ a_i, u_i \}_{i=1}^N \), where \( u_i\!\approx\!\mathcal{G}(a_i) \), and $N$ is the number of function samples. The approximation of the unknown operator is achieved through minimizing the loss functional over a parameterized space of learnable parameters ($\theta$), $\min_{\theta} \sum_{i=1}^N \| \mathcal{N}_\theta(a_i) - u_i \|_{\mathcal{U}}^2$, where \( \| \cdot \|_{\mathcal{U}} \) is an appropriate norm on the output space. Once trained, the neural operator generalizes to unseen input functions within \( \mathcal{A} \), allowing the prediction of output functions.

However, despite their success in various scientific \cite{yang2024fourier} and engineering tasks \cite{borrel2024sound}, neural operators face two major limitations. First, they lack interpretability, as the learned mapping is embedded within high-dimensional latent representations, making extracting symbolic insights or physical laws challenging. Second, their generalizability is challenging outside the training distribution $\mathcal{A}$. 

\subsection{Model Discovery Methods}
Model discovery methods aim to learn parsimonious symbolic models from data. These methods assume that a sparse combination of candidate basis functions can approximate the underlying system dynamics. Several sparse regression-based approaches have been proposed recently, including but not limited to sparse identification of nonlinear dynamics \cite{brunton2016discovering} for discovering ordinary differential equations (ODEs) and PDE-FIND \cite{rudy2017data} for discovering partial differential equations. The approaches aim to describe the system dynamics while being interpretable and possibly consistent with classical physics or engineering laws.

Abstractly, let \( \mathrm{X}(t) \in \mathbb{R}^n \) represent the state of a dynamical system, where $n$ represents the number of time steps of data collection, and assume its time evolution satisfies an ODE ($\mathcal{S}$) of the form, $\mathcal{S}:=\frac{d\mathrm{X}}{dt} = \mathrm{f}(\mathrm{X})$, where \( \mathrm{f} \) is an unknown function. Model discovery methods construct a library \( \boldsymbol{\Theta}(\mathrm{X}) \in \mathbb{R}^{n \times p} \) of $p$ candidate basis functions (e.g., polynomials, trigonometric functions), and solve a sparse regression problem to approximate $\frac{d\mathrm{X}}{dt} \approx \boldsymbol{\Theta}(\mathrm{X}) \xi$.

Here, \( \xi\! \in\! \mathbb{R}^p \) is a sparse coefficient vector selecting the most relevant terms from the library. The learning is accomplished via a sparsity-promoting norm leading to few terms in the final representation \cite{tibshirani1996regression}. However, these methods face challenges for noisy and low-fidelity data, such as computing derivatives, leading to inaccurate or unstable model estimates \cite{fasel2022ensemble, wentz2023derivative}.

\section{Problem Formulation}
\label{sec3}
Modeling hysteresis in piezoelectric systems can be framed as a neural operator learning problem. As introduced in Section~\ref{sec2}, the system exhibits a nonlinear, history-dependent relationship between the input voltage \(v(t)\!\in\!\mathcal{V}\!\subset\!C([0, \mathrm{T}]; \mathbb{R})\) and the output displacement \(d(t)\!\in\!\mathcal{D}\!\subset\!C([0, \mathrm{T}]; \mathbb{R}) \). The goal is to learn an operator \(\mathcal{P}\!:\!\mathcal{V}\!\to\!\mathcal{D}\) such that $d(t)\!=\!\mathcal{P}[v](t),\! \forall t\! \in\! [0, \mathrm{T}]$, which aligns with the general neural operator framework \( u\!=\!\mathcal{G}(a) \), where the input function \( a \) corresponds to the voltage \( v \), the output function \( u \) corresponds to the displacement \( d \), and the operator $\mathcal{G}$ corresponds to $\mathcal{P}$. 

Mathematically, given a set of input–output pairs \( \{v_i(t), d_i(t)\}_{i=1}^N \), an operator \( \hat{\mathcal{P}} \) is sought, such that \( \hat{\mathcal{P}}[v_i](t) \approx d_i(t) \) with high accuracy. Therefore, hysteresis modeling amounts to neural operator learning where the learned mapping should satisfy some properties: (i) history preservation, ensuring modeling of memory effects; (ii) generalization, that is, the operator performs well on inputs \( v(t) \notin \text{span}\{v_i(t)\} \); and (iii) stability and fidelity, particularly under noisy or low-fidelity sampling conditions. 

However, neural operators, $\mathcal{N_\theta}$, are not interpretable, and their generalizability is challenging outside $\mathcal{V}$. These limitations motivate the need to distill neural operators into interpretable, symbolic representations. One potential solution is to extract white-box models from the learned operator \( \mathcal{N}_\theta \), enabling symbolic model discovery that approximates \( \mathcal{P} \) using parsimonious equations. Formally, the objective becomes to identify a symbolic operator \( \widehat{\mathcal{S}} \) such that: $\widehat{\mathcal{S}} \approx \mathcal{N}_\theta = \widehat{\mathcal{P}} \approx \mathcal{P}$. This formulation aims to bridge the gap between black-box operator learning and interpretable model discovery.

\section{Proposed Framework: NSO}
\label{sec4}
This section presents the proposed neuro-symbolic operator (NSO) framework that combines neural operators with symbolic regression to discover interpretable hysteresis models. The framework comprises of two stages: first, an operator FNO \cite{lifourier} is trained to predict displacement profiles \( \widehat{d}_i \) for input voltage fields \( v_i \); second, the predictions of FNO, \( \widehat{d}_i \), are used to extract sparse, low-dimensional hysteresis models (in parametric space) using the sequential threshold least-squares (STLSQ) algorithm \cite{brunton2016discovering}. While the proposed framework is general and, in principle, can be applied with any neural operator architecture and symbolic regression technique, FNO is employed due to its demonstrated efficacy in learning operators across a range of physical systems \cite{chandra2024magnetic, chandra2024characterizing, guo2025dynamic}. Similarly, the STLSQ algorithm is chosen for its ability to discover parsimonious and interpretable hysteresis models with high accuracy \cite{chandra2023discovery}.

\subsection{Stage I: Operator Learning via FNO}
In the first stage, FNO is employed to learn the mapping between the function spaces, $\widehat{\mathcal{P}}\! =\! \mathcal{N}_\theta:\! \mathcal{V}\! \mapsto\! \mathcal{D}$, where \( \theta \) denotes the trainable parameters of the network. FNO leverages convolutional neural network framework and takes as input the normalized voltage fields \( v_i\!\in\!\mathcal{V} \), and concatenates it with the time array to form a two-channel input tensor \( \mathrm{I}\! :=\! [v_i, t] \). The tensor $\mathrm{I}$ is projected into a high-dimensional space via a learnable projection \( \mathrm{P} \), increasing the number of channels to \( N_\text{f} \) with initial latent tensor \( u^0 = \mathrm{P} \mathrm{I} \). This latent representation is passed through \( K \) Fourier layers. The transformation at the \( k^\text{th} \) Fourier layer, $0 \leq k \leq K-1$, is given by:
\begin{equation}
u^{(k+1)} = \sigma\left(\mathbf{W}^{(k)} u^{(k)} + \mathcal{F}^{-1} \left( \mathbf{R}^{(k)} \mathcal{F}(u^{(k)}) \right) \right).
\label{eqn4}
\end{equation}

Here \( \mathcal{F} \) and \( \mathcal{F}^{-1} \) denote the fast Fourier transform and its inverse, \( \mathbf{W}^{(k)} \) is a learnable local transformation, and \( \mathbf{R}^{(k)} \) is a learnable tensor that modulates the first \( n_m \) frequency modes in the complex domain. The activation function \( \sigma \) introduces nonlinearity. This process is repeated for \( K \) Fourier layers. After \( K \) such transformations, the final latent representation \( u^{(K)} \) is projected back to the physical domain through two projection tensors, $\mathbf{Q}$ and $\widehat{\mathbf{Q}}$, to generate the predicted displacements \( \hat{d}_i \) as, $\widehat{d}_i = \widehat{\mathbf{Q}} \, \sigma(\mathbf{Q} u^{(K)})$. Finally, the model is trained using a supervised loss over \( N_\text{train} \) training samples:
\begin{equation}
\mathcal{L}(\theta) = \frac{1}{N_\text{train}} \sum_{i=1}^{N_\text{train}} \left( \widehat{d}_i(\theta) - d_i \right)^2.
\label{eqn6}
\end{equation}

Here, \( d_i \) is the measured displacement corresponding to input \( v_i \). Training is performed via standard gradient-based optimization algorithms, such as ADAM \cite{kingma2017adammethodstochasticoptimization}.

\subsection{Stage II: Sparse Model Discovery via STLSQ}
The trained FNO, $\mathcal{N_{\theta}}$, is used for predicting $\widehat{d_i}$ for novel $v_i\! \in\! \mathcal{V}$, not used while training. As discussed in Section~\ref{sec2}, symbolic model discovery methods typically require high-fidelity, low-noise data to compute temporal derivatives and identify governing equations accurately. In the proposed framework, FNO helps overcome this challenge as $\mathcal{N_{\theta}}$ acts as a denoising and upsampling surrogate model, owing to the properties of neural operators. Specifically, trained on noisy, low-fidelity input-output pairs \( \{v_i, d_i\}_{i=1}^{N_\text{train}} \), the neural operator \( \mathcal{N}_\theta \) can produce noise-robust and fidelity-invariant functional predictions \( \{ \widehat{d}_j = \mathcal{N}_\theta(v_j) \}_{j=1}^{N_\text{test}} \), which are then suitable for symbolic regression. The predicted outputs \( \widehat{d}_j(t) \) are used for symbolic model identification. Specifically, the predicted displacement profiles \( \{ \widehat{d}_j(t) \}_{j=1}^{N_\text{test}} \) and their corresponding inputs \( \{ v_j(t) \}_{j=1}^{N_\text{test}} \) are collected to form column vectors:
\[
\mathrm{D} := \begin{bmatrix}
\widehat{d}_1(t)^\top & \widehat{d}_2(t)^\top & \cdots & \widehat{d}_{N_\text{test}}(t)^\top
\end{bmatrix}^\top \in \mathbb{R}^{N_\text{test} \cdot n \times 1},
\]
\[
\mathrm{V} := \begin{bmatrix}
\widehat{v}_1(t)^\top & \widehat{v}_2(t)^\top & \cdots & \widehat{v}_{N_\text{test}}(t)^\top
\end{bmatrix}^\top \in \mathbb{R}^{N_\text{test} \cdot n \times 1},
\]
where \( n \) is the number of time steps for each sample and the superscript \( (\cdot)^\top \) denotes transpose for conciseness. The derivative of D with respect to $t$ is denoted as: 
\[
\dot{\mathrm{D}} := \begin{bmatrix}
\dot{\widehat{d}}_1(t)^\top & \dot{\widehat{d}}_2(t)^\top & \cdots & \dot{\widehat{d}}_{N_\text{test}}(t)^\top
\end{bmatrix}^\top \in \mathbb{R}^{N_\text{test} \cdot n \times 1}.
\]

To identify the underlying dynamics, a candidate library of $p$ basis functions is constructed as:
\[
\boldsymbol{\Theta}\! =\!\! \begin{pmatrix}
\vline & \cdots & \vline \\
\theta_1(\mathrm{D}, \mathrm{V}, f(\mathrm{V})) & \cdots & \theta_p(\mathrm{D}, \mathrm{V}, f(\mathrm{V})) \\
\vline & \cdots & \vline
\end{pmatrix}\!\!\! \in\! \mathbb{R}^{N_\text{test} \cdot n \times p},
\]
where each basis function \( \theta_j(\cdot) \), $1\! \leq j\! \leq p$ is a nonlinear combination of the predicted displacements, their time derivatives, voltage inputs, and $f(\mathrm{V})$ denoting terms depending on $\mathrm{V}$. These functions are inspired by known hysteresis models such as Bouc--Wen and generic polynomial and trigonometric terms. Subsequently, the governing model is assumed to be a sparse linear combination of columns from the library:
\begin{equation}
    \widehat{\mathcal{S}} = \dot{\mathrm{D}} = \boldsymbol{\Theta}(\mathrm{D}, \mathrm{V}, f(\mathrm{V})) \, \xi,
\end{equation}
\begin{algorithm}[t]
\caption{\textbf{NSO Framework}}
\label{alg1}
\begin{algorithmic}[1]

\STATE \textbf{Input:} Training data $\{(v_i, d_i)\}_{i=1}^{N_\text{train}}$, test inputs $\{v_j\}_{j=1}^{N_\text{test}}$
\STATE \textbf{Output:} Sparse model coefficients $\xi^*$, trained operator $\mathcal{N}_\theta$

\STATE \textbf{Stage I: Train FNO} 
\STATE Define $\mathcal{N}_\theta: \mathcal{V} \to \mathcal{D}$ with parameters $\theta$
\STATE Form input tensor $\mathrm{I} := [v_i, t]$
\STATE Initialize latent $u^0 = \mathrm{P} \mathrm{I}$
\FOR{$k=0$ to $K-1$}
    \STATE $u^{(k+1)} = \sigma\big(\mathbf{W}^{(k)} u^{(k)} + \mathcal{F}^{-1}(\mathbf{R}^{(k)} \mathcal{F}(u^{(k)}))\big)$
\ENDFOR
\STATE Predict $\widehat{d}_i = \widehat{\mathbf{Q}}\,\sigma(\mathbf{Q} u_i^{(K)})$
\STATE Train FNO ($\mathcal{N}_\theta$): minimize $\mathcal{L}(\theta) = \frac{1}{N_\text{train}} \sum_{i=1}^{N_\text{train}} \| \widehat{d}_i - d_i \|^2_2$

\STATE \textbf{Stage II: Sparse Model Discovery via STLSQ}
\STATE Predict $\widehat{d}_j = \mathcal{N}_\theta(v_j)$ for $j=1,\ldots,N_\text{test}$
\STATE Assemble $\mathrm{D}$, $\dot{\mathrm{D}}$, $\mathrm{V}$, and $f(\mathrm{V})$
\STATE Construct candidate library $\boldsymbol{\Theta}(\mathrm{D}, \mathrm{V}, f(\mathrm{V}))$
\STATE Solve $\dot{\mathrm{D}} = \boldsymbol{\Theta} \xi, \quad \xi^* = \text{STLSQ}(\boldsymbol{\Theta}, \dot{\mathrm{D}}, \lambda)$
\STATE \textbf{return} $\mathcal{N}_\theta, \xi^*$

\end{algorithmic}
\end{algorithm}
where \( \xi \in \mathbb{R}^{p \times 1} \) is a sparse coefficient vector representing the active terms. To enforce sparsity STLSQ algorithm is applied, which iteratively performs least-squares fits and zeros out coefficients below a given threshold ($\lambda$) to yield a parsimonious and interpretable model as, $\xi^* = \text{STLSQ}(\boldsymbol{\Theta}, \dot{\mathrm{D}}, \lambda)$.

The threshold parameter, $\lambda$, is selected based on the desired balance between model sparsity and accuracy. Most entries in \( \xi^* \) result in zero, and the remaining nonzero entries identify the dominant terms governing the hysteresis behavior.

\subsection{Summary of the NSO Framework}
The proposed two-stage framework leverages the strengths of both neural operators and symbolic model discovery. The operator, FNO, learns a resolution-invariant mapping \( v_i \mapsto \widehat{d}_i \), facilitating denoising and upsampling, and STLSQ assists in extracting sparse, interpretable equations from the predicted trajectories \( \widehat{d}_i \). The steps are also briefly presented in Algorithm~\ref{alg1}. The resulting NSO framework aims to provide predictive accuracy and physical insight, offering an interpretable and generalizable approach for modeling hysteresis.

\begin{table*}[t]
\setlength{\tabcolsep}{2.8pt}
\begin{center}
\caption{Performance of neural operators across experiments under different error metrics. The models are trained on the Sine kernel and tested on Sine, RBF, and Matern kernels.}
\label{tbl1}
\begin{tabular}{llcccccccccccc}
\toprule
\textbf{Experiment} & \textbf{Kernel} 
& \multicolumn{3}{c}{\textbf{DON}} 
& \multicolumn{3}{c}{\textbf{FNO}} 
& \multicolumn{3}{c}{\textbf{CNO}} 
& \multicolumn{3}{c}{\textbf{NSO}} \\
\cmidrule(lr){3-5} \cmidrule(lr){6-8} \cmidrule(lr){9-11} \cmidrule(lr){12-14}
 & & $\mathcal{R}$ & RMSE & MAE & $\mathcal{R}$ & RMSE & MAE & $\mathcal{R}$ & RMSE & MAE & $\mathcal{R}$ & RMSE & MAE \\
\midrule
\multirow{3}{*}{Exp 1} 
 & Sine      & 8.36e-01 & 8.63e-02 & 6.55e-02 & \textbf{7.95e-04} & \textbf{8.21e-05} & \textbf{6.29e-05} & 9.50e-03 & 9.82e-04 & 8.84e-04 & 8.11e-03 & 8.38e-04 & 5.72e-04 \\
 & RBF       & 3.40e+00 & 8.63e-02 & 3.42e-01 & 9.03e-01 & 1.11e-01 & 9.00e-02 & 4.64e-01 & 5.72e-02 & 4.41e-02 & \textbf{1.01e-02} & \textbf{1.25e-03} & \textbf{9.97e-04} \\
 & Matern52  & 3.21e+00 & 8.63e-02 & 3.33e-01 & 8.87e-01 & 1.13e-01 & 9.17e-02 & 4.34e-01 & 5.52e-02 & 4.32e-02 & \textbf{9.75e-03} & \textbf{1.24e-03} & \textbf{1.01e-03} \\
\midrule
\multirow{3}{*}{Exp 2} 
 & Sine      & 3.65e-02 & 6.14e-02 & 4.42e-02 & 1.12e-02 & 1.88e-02 & 1.49e-02 & \textbf{1.11e-03} & \textbf{1.86e-03} & \textbf{1.44e-03} & 1.24e-02 & 2.09e-02 & 1.76e-02 \\
 & RBF       & 1.21e+00 & 6.14e-02 & 2.10e+00 & 9.21e-01 & 1.99e+00 & 1.64e+00 & 7.64e-01 & 1.65e+00 & 1.33e+00 & \textbf{1.05e-02} & \textbf{2.28e-02} & \textbf{2.03e-02} \\
 & Matern52  & 1.18e+00 & 6.14e-02 & 1.98e+00 & 8.94e-01 & 1.86e+00 & 1.53e+00 & 7.54e-01 & 1.57e+00 & 1.26e+00 & \textbf{1.08e-02} & \textbf{2.24e-02} & \textbf{2.00e-02} \\
\midrule
\multirow{3}{*}{Exp 3} 
 & Sine      & 1.67e-01 & 4.02e-02 & 2.78e-02 & \textbf{1.16e-03} & \textbf{2.78e-04} & \textbf{2.15e-04} & 8.78e-03 & 2.11e-03 & 1.73e-03 & 4.28e-03 & 1.03e-03 & 7.38e-04 \\
 & RBF       & 1.49e+00 & 4.02e-02 & 2.95e-01 & 6.86e-01 & 1.69e-01 & 1.26e-01 & 6.17e-01 & 1.52e-01 & 1.09e-01 & \textbf{4.80e-03} & \textbf{1.19e-03} & \textbf{8.62e-04} \\
 & Matern32  & 1.66e+00 & 4.02e-02 & 2.89e-01 & 7.78e-01 & 1.70e-01 & 1.27e-01 & 6.30e-01 & 1.38e-01 & 1.01e-01 & \textbf{4.88e-03} & \textbf{1.07e-03} & \textbf{7.92e-04} \\
\midrule
\multirow{3}{*}{Exp 4} 
 & Sine      & 5.50e-01 & 3.11e-01 & 2.24e-01 & \textbf{5.72e-04} & \textbf{3.23e-04} & \textbf{2.41e-04} & 1.66e-03 & 9.40e-04 & 7.23e-04 & 3.91e-02 & 2.21e-02 & 1.63e-02 \\
 & RBF       & 1.04e+00 & 3.11e-01 & 5.05e-01 & 8.03e-01 & 4.99e-01 & 3.57e-01 & 7.60e-01 & 4.73e-01 & 3.58e-01 & \textbf{3.69e-02} & \textbf{2.29e-02} & \textbf{1.70e-02} \\
 & Matern32  & 1.08e+00 & 3.11e-01 & 4.45e-01 & 7.31e-01 & 3.90e-01 & 2.81e-01 & 7.87e-01 & 4.20e-01 & 3.15e-01 & \textbf{3.98e-02} & \textbf{2.12e-02} & \textbf{1.56e-02} \\
\bottomrule
\end{tabular}
\end{center}
\end{table*}

\section{Numerical Experiments}
\label{sec5}
The following subsections present seven numerical experiments to evaluate the performance of NSO. Experiments (Exp) 1-4 showcase the advantages of NSO compared with baseline neural operators. Additionally, ablation studies are conducted to investigate the robustness of NSO under noisy (Exp 5a) and low-fidelity data conditions (Exp 5b) and to examine the influence of the sparsity threshold ($\lambda$) on model learning (Exp 6). The experiments are structured progressively, starting with standard hysteresis responses and advancing to more complex butterfly-shaped hysteresis behavior. 

\subsection{Experiment Setting}
This subsection describes the baselines, evaluation metrics, and hyperparameters used in the paper.

\subsubsection{Baselines}
The proposed framework NSO is compared with three neural operators, DON \cite{lu2021learning}, FNO \cite{lifourier}, and CNO \cite{raonic2023convolutional}. These neural operators are widely used for infinite-dimensional function-to-function mapping. NSO is also compared with model discovery methods, Lasso \cite{tibshirani1996regression}, and SINDy \cite{brunton2016discovering}. NSO is evaluated against these baseline methods, and the corresponding advantages, such as interpretability and generalizability, are observed in the following experiments.

\subsubsection{Evaluation Metrics}
The methods are evaluated on qualitative figures and error metrics: relative L2 error ($\mathcal{R}$), root mean square error (RMSE), and mean absolute error (MAE). For $\hat{d}_j$ denoting displacement predictions and ${d}_j$ denoting the true displacements, the metrics are defined as $\mathcal{R} = \frac{\sqrt{\frac{1}{N_{\text{test}}} \sum_{j=1}^{N_{\text{test}}} (\hat{d}_j - d_j)^2}}{\sqrt{\frac{1}{N_{\text{test}}} \sum_{j=1}^{N_{\text{test}}} d_j^2}}$; $\text{RMSE} = \sqrt{\frac{1}{N_{\text{test}}} \sum_{j=1}^{N_{\text{test}}} (\hat{d}_j - d_j)^2}$; and $\text{MAE} = \frac{1}{N_{\text{test}}} \sum_{j=1}^{N_{\text{test}}} |\hat{d}_j - d_j|$.

\subsubsection{Hyperparameters} 
The hyperparameters that remain consistent across experiments are as follows. For each experiment, the models are trained only on functions sampled from the Sine kernel and tested on Sine, RBF, and Matern kernel functions \cite{williams2006gaussian}. The sampling strategy is described later for each experiment in their respective subsections. The baseline methods DON, FNO, CNO, Lasso, and SINDy utilize 1000 functions for training ($N_\textrm{train}$) in the Sine kernel and 1000 functions for testing the model ($N_\textrm{test}$) for each kernel. As the second stage of NSO utilizes the testing data of FNO for the Sine kernel case, it is trained and tested on 500 functions based on the predictions of FNO on the Sine kernel. The same 1000 functions are used across all methods for testing on other kernels. For all experiments except 5b, the input and output functions are sampled on 100-time points, \textit{i.e.,} $n\!=\!100$. For experiment 5b, the low fidelity samples correspond to $n\!=\!20$. 

The neural operator DON employs a consistent architecture and training setup across all experiments. The model consists of a branch and a trunk network, each composed of five fully connected hidden layers with 40 neurons per layer and a tanh activation function. The branch network processes 100-dimensional ($n$) input functions ($v_i$), while the trunk network takes a single temporal coordinate ($t$) as input. Both networks output 100-dimensional representations, which are combined using an inner product to produce the final prediction ($\widehat{d_i}$). The networks are trained for 5000 iterations using the Adam optimizer with a learning rate of 5e-5.

The neural operator FNO consists of four spectral convolution layers ($K=4$) with ReLU activation function. The input to the network comprises two channels\textemdash $v_i$ and $t$\textemdash lifted to $N_f\!=\!64$ hidden channels through a fully connected layer. The number of Fourier modes used in the spectral convolution is fixed at $n_m\!=\!32$. The training uses the Adam optimizer to optimize the mean square loss with a learning rate of 0.001, batch size of 100, and over 500 epochs. 

\begin{figure*} 
\centering
\includegraphics[width=0.30\columnwidth]{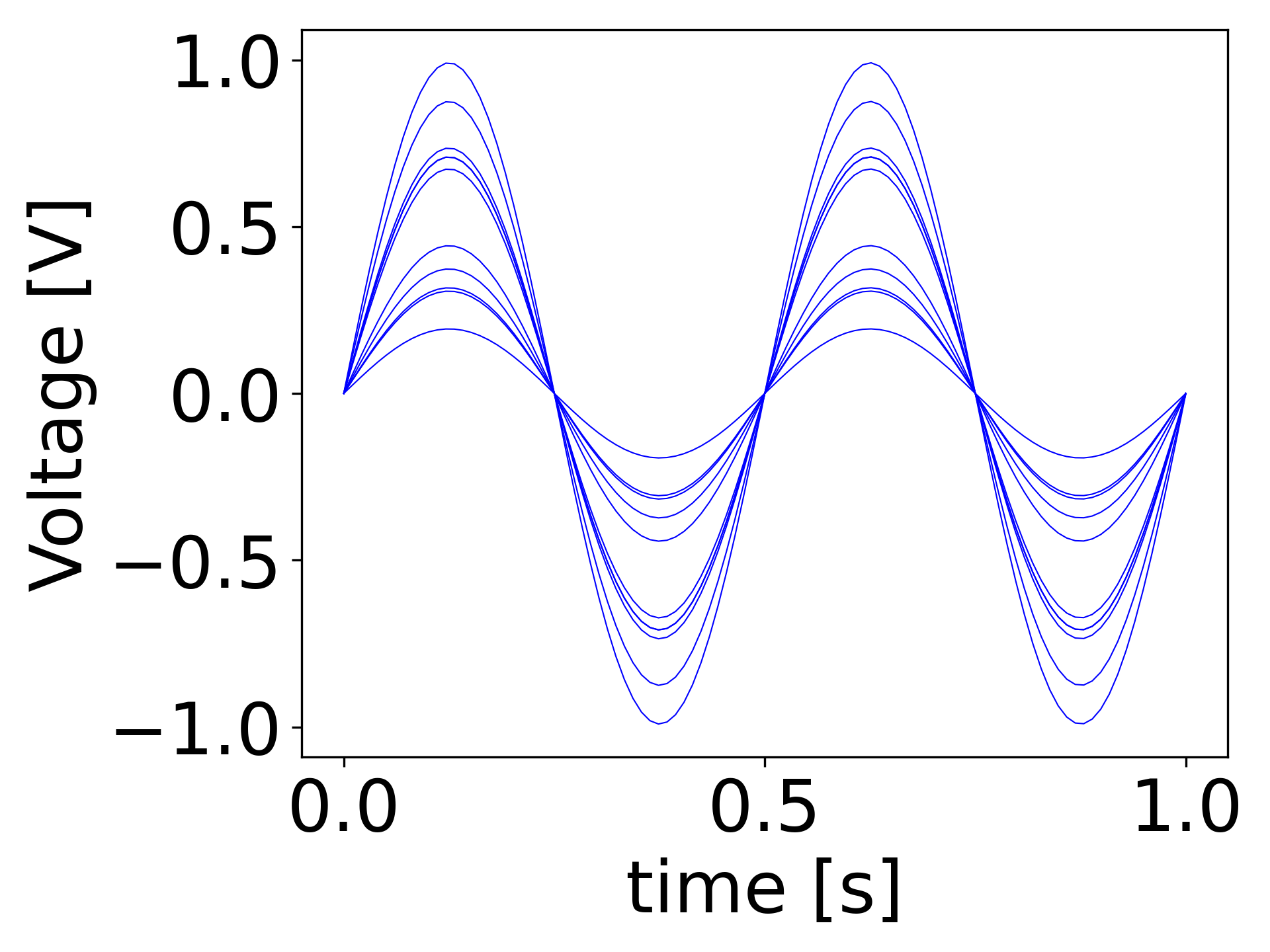}
\includegraphics[width=0.30\columnwidth]{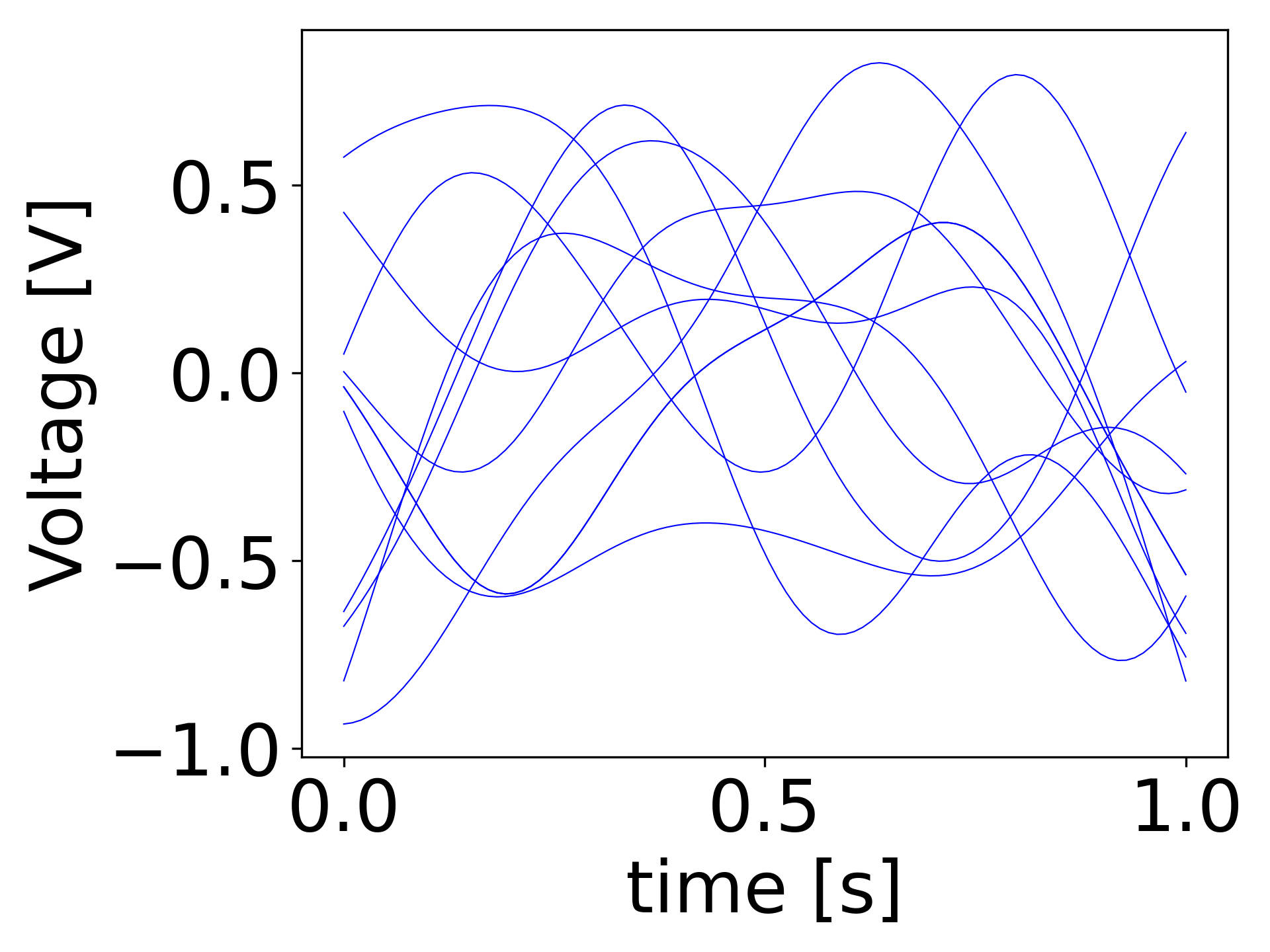}
\includegraphics[width=0.30\columnwidth]{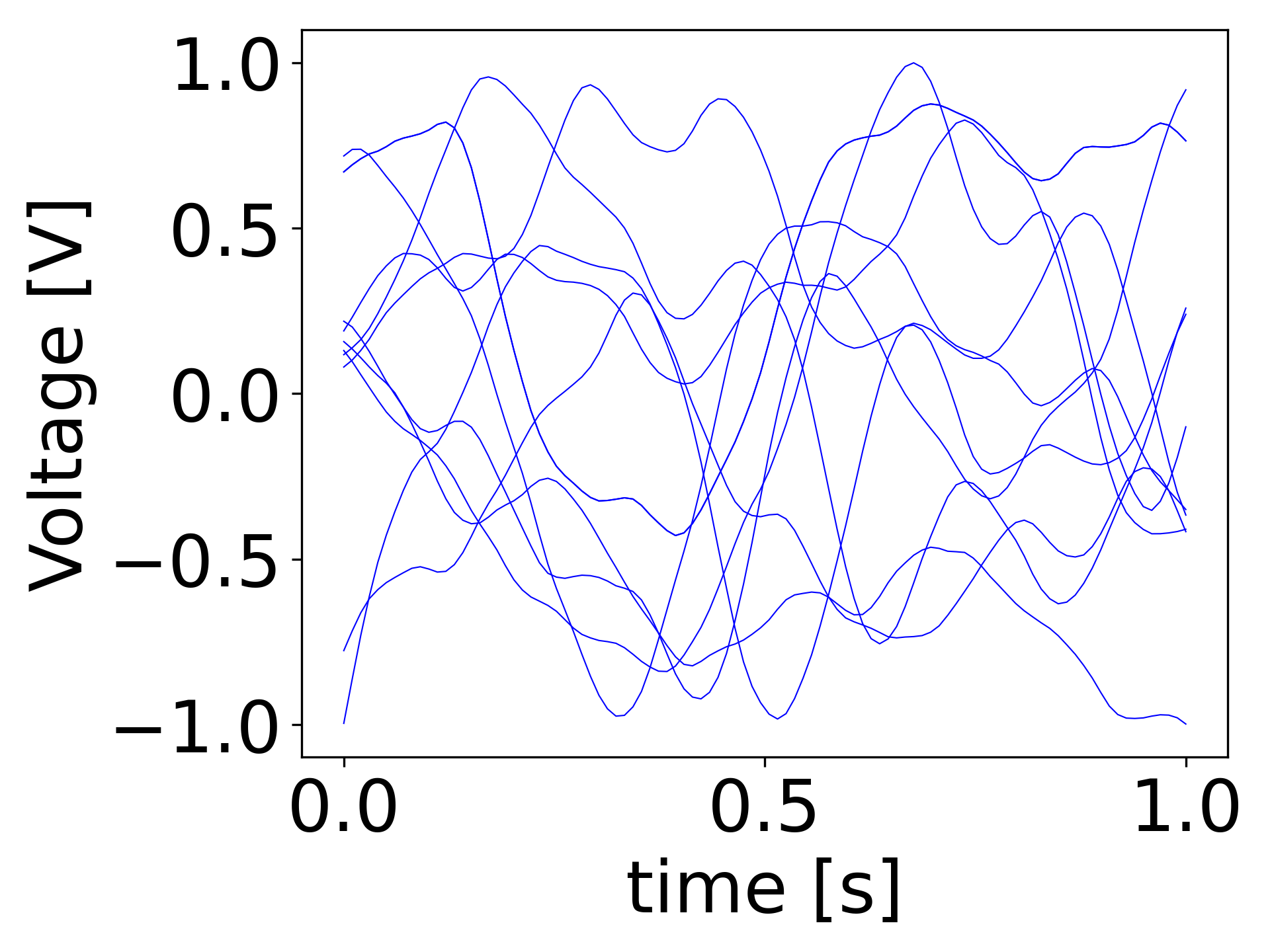}
\includegraphics[width=0.30\columnwidth]{Figures/Exp1/Exp1_Sine.png}
\includegraphics[width=0.30\columnwidth]{Figures/Exp1/Exp1_RBF.png}
\includegraphics[width=0.30\columnwidth]{Figures/Exp1/Exp1_Matern52.png}

\includegraphics[width=0.30\columnwidth]{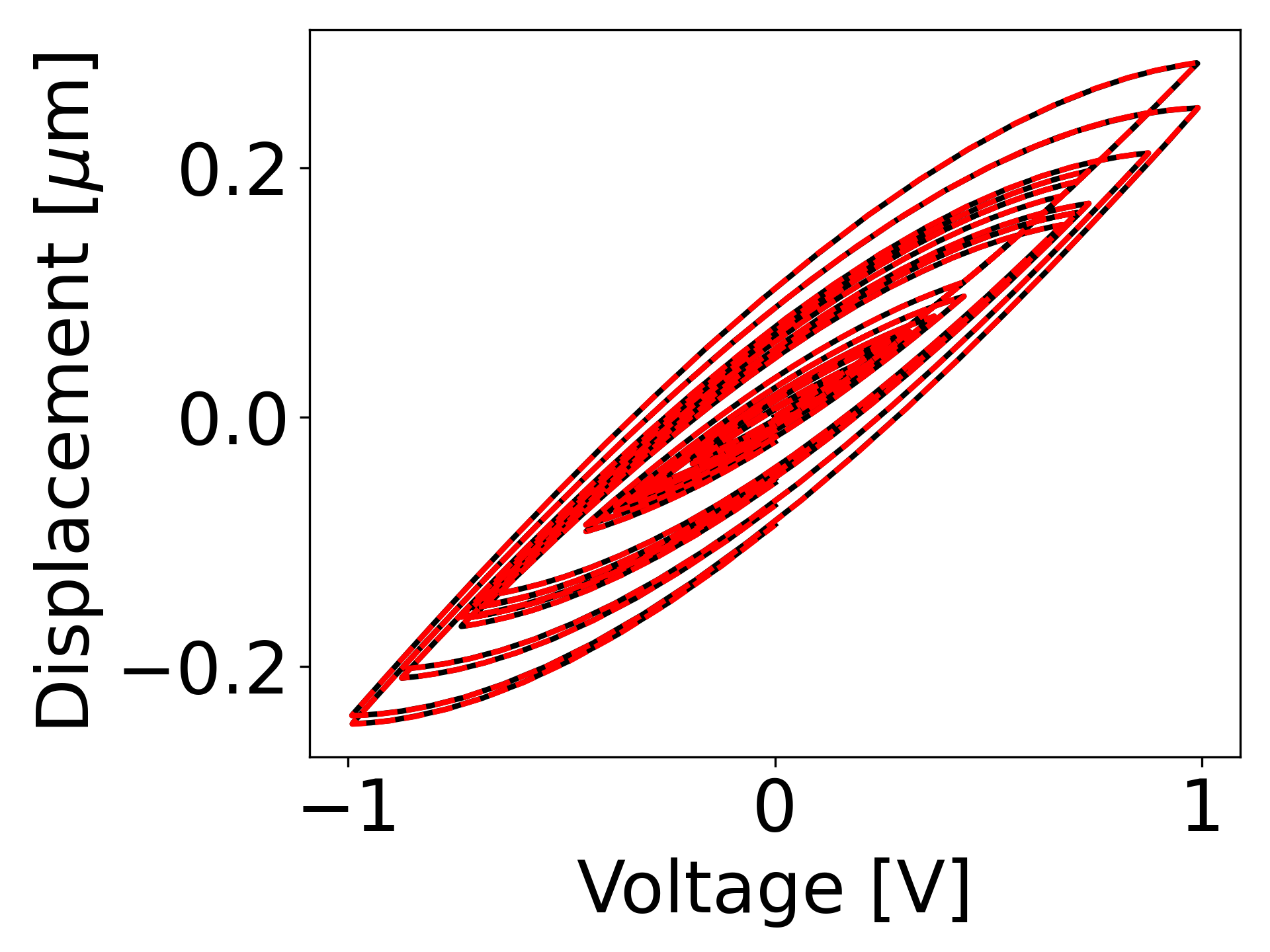}
\includegraphics[width=0.30\columnwidth]{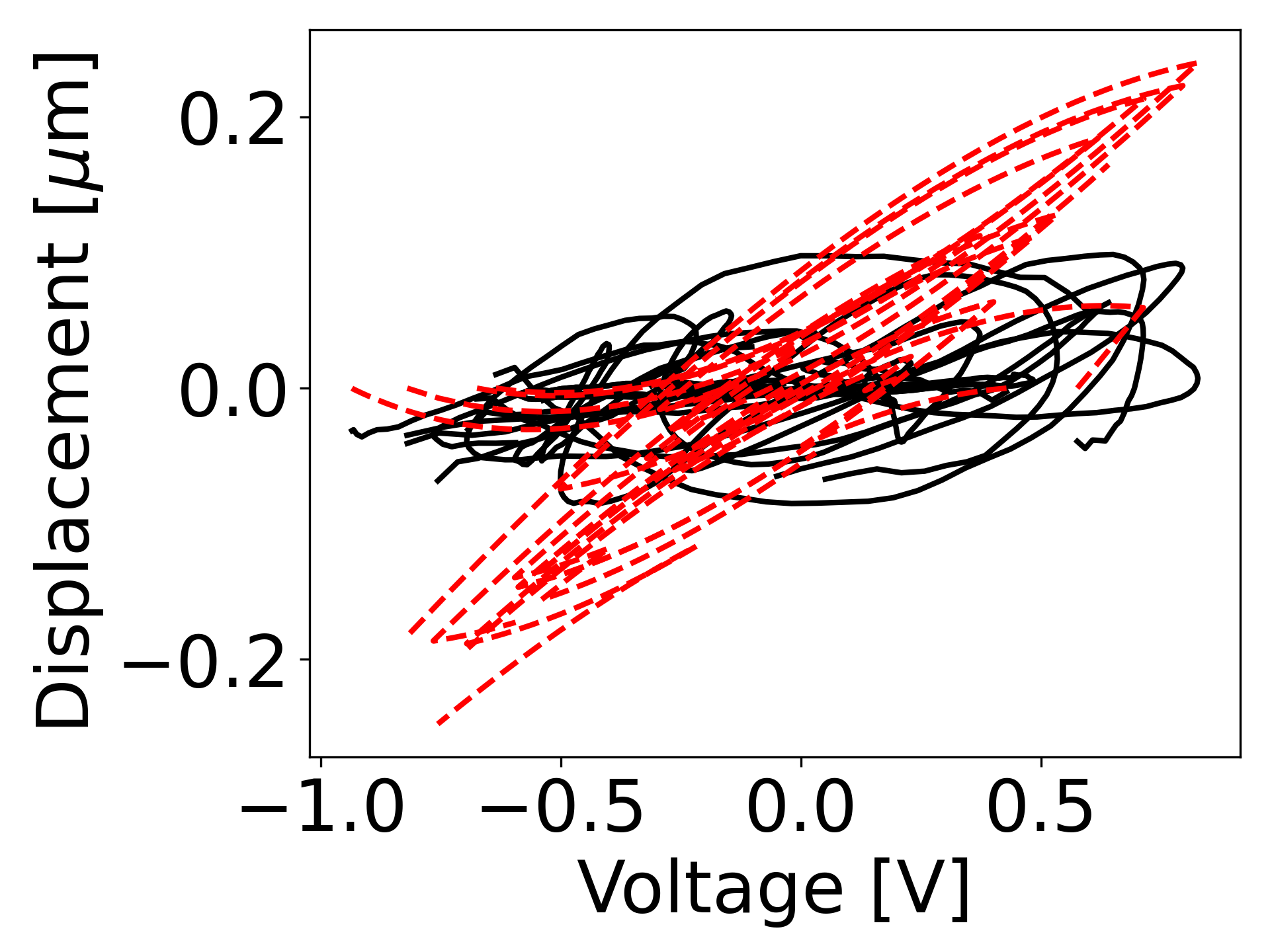}
\includegraphics[width=0.30\columnwidth]{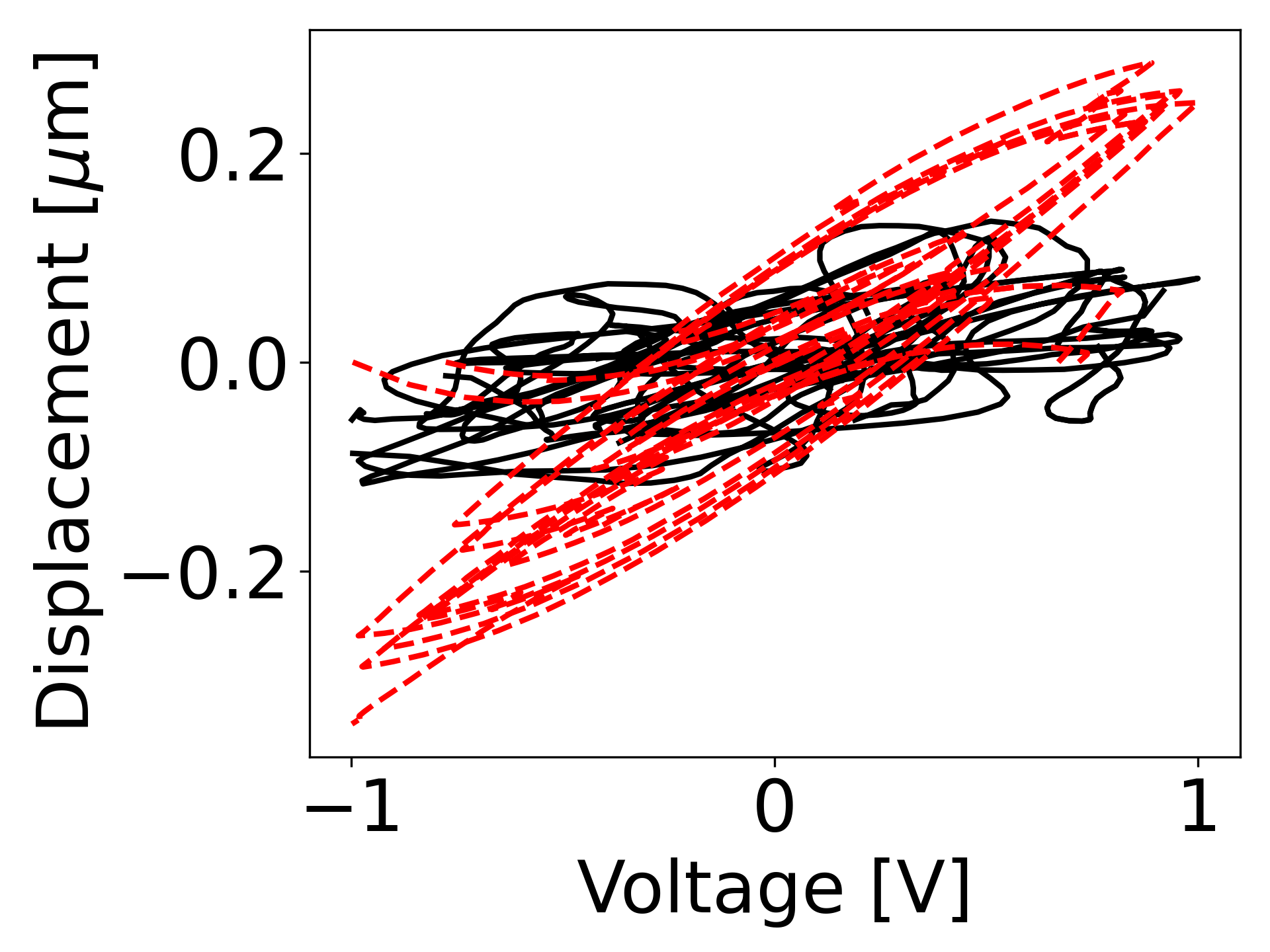}
\includegraphics[width=0.30\columnwidth]{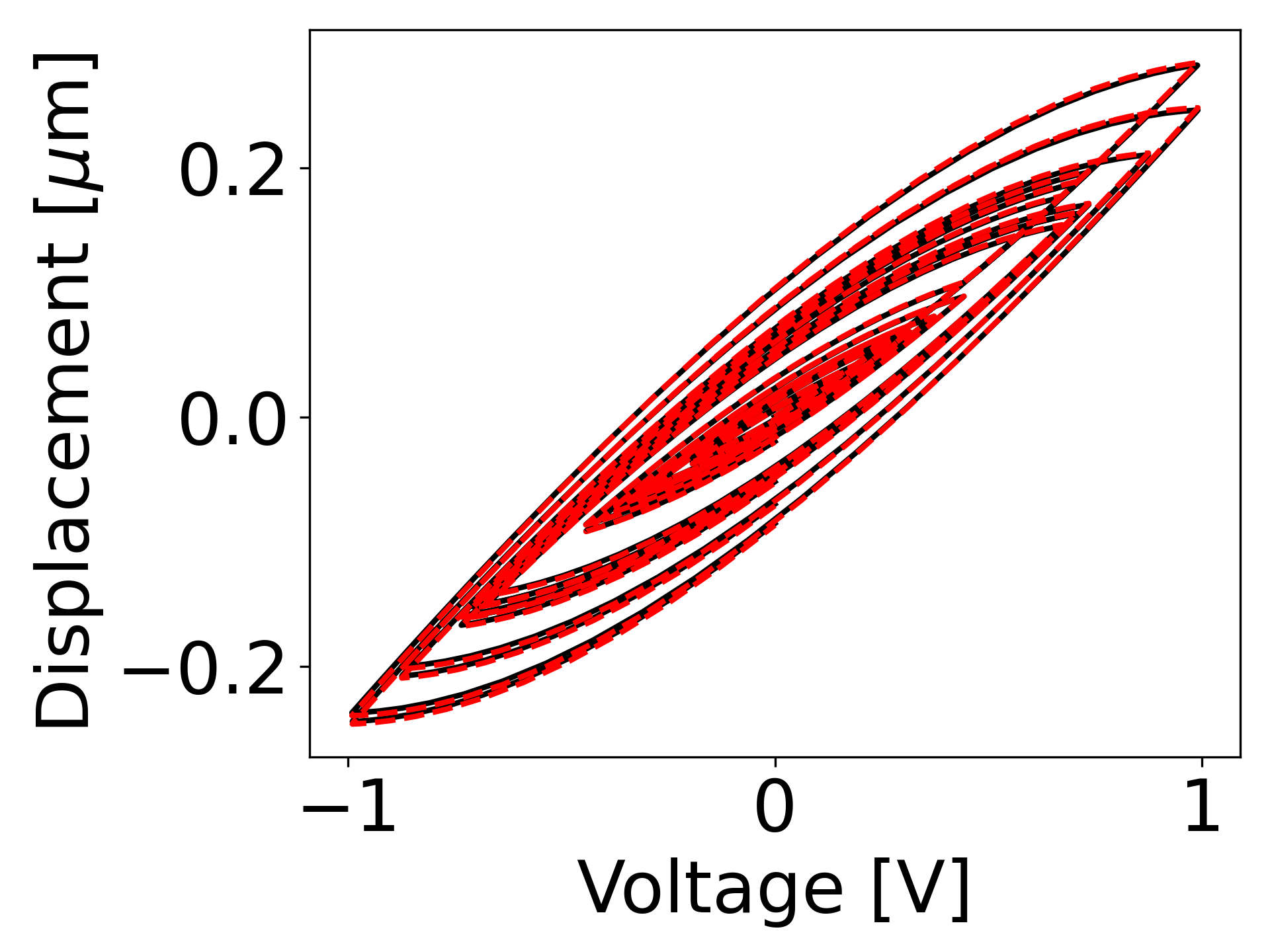}
\includegraphics[width=0.30\columnwidth]{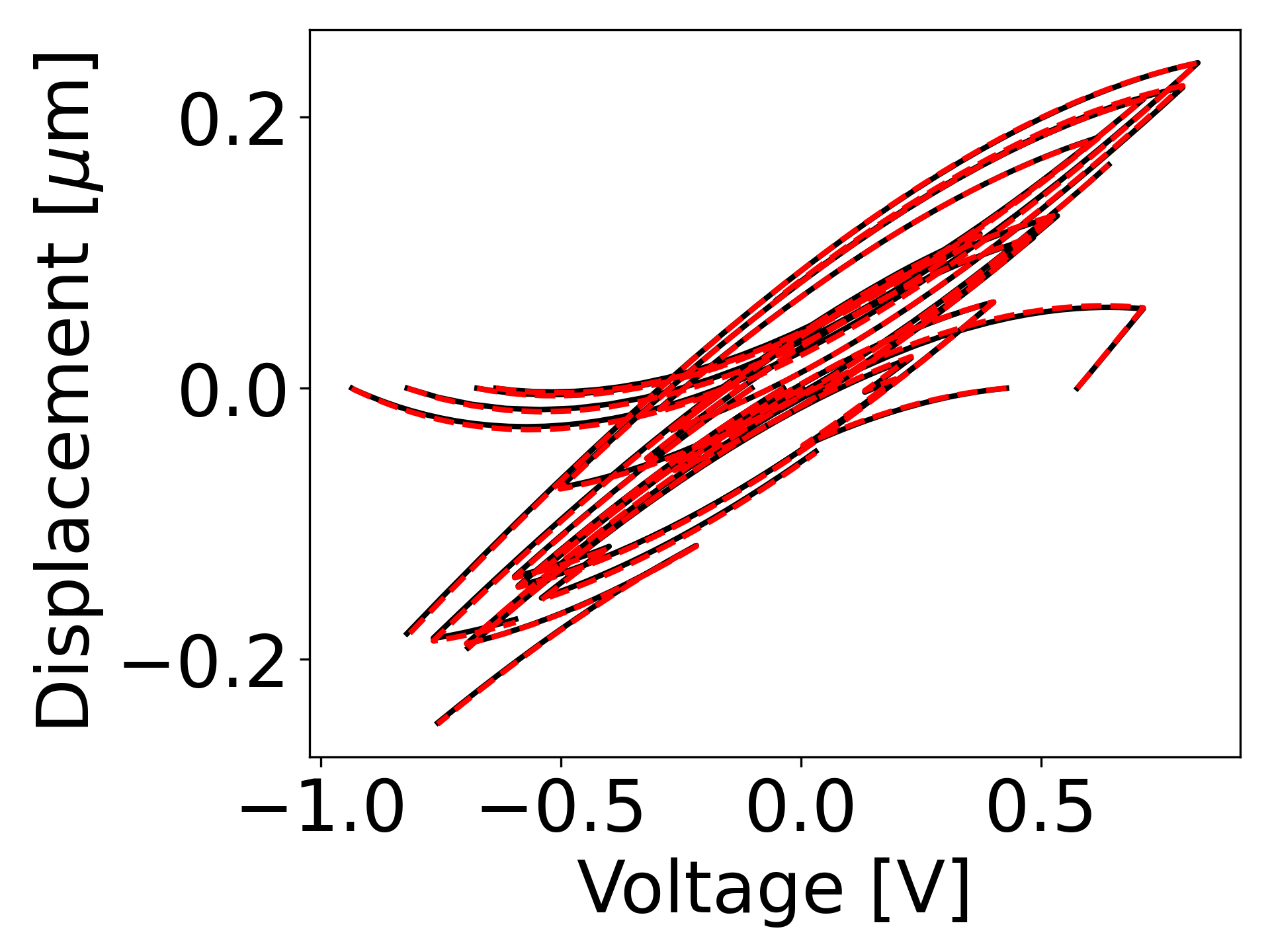}
\includegraphics[width=0.30\columnwidth]{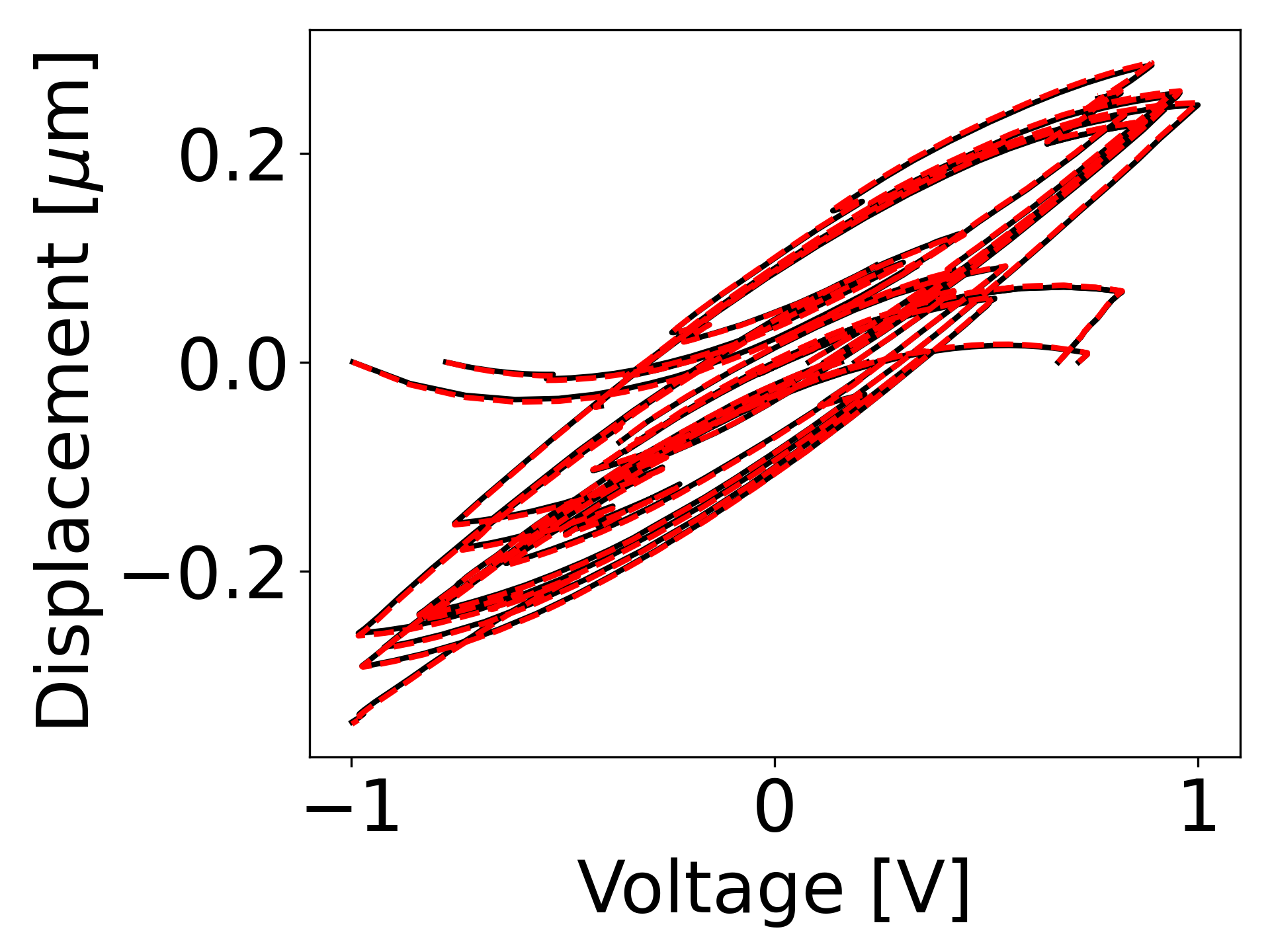}
\caption{Results for Experiment 1: \textbf{Top row: }Blue curves represent the testing voltage fields sampled from Sine, RBF, and Matern52 kernels for FNO (first three columns, from left to right) and NSO (last three columns, from left to right). \textbf{Bottom row: }Hysteresis responses for FNO (first three columns, from left to right) and NSO (last three columns, from left to right) corresponding to the voltage fields on the Top row. Red dashed curves represent the ground truth, while Black curves show model predictions. The models are trained on the Sine kernel's voltages and tested on Sine, RBF, and Matern52 kernels' voltages. The colors, directions (from left to right), and training-testing criteria remain the same for all the figures.}
\label{Nfig1}
\end{figure*}

\begin{figure*} 
\centering
\includegraphics[width=0.30\columnwidth]{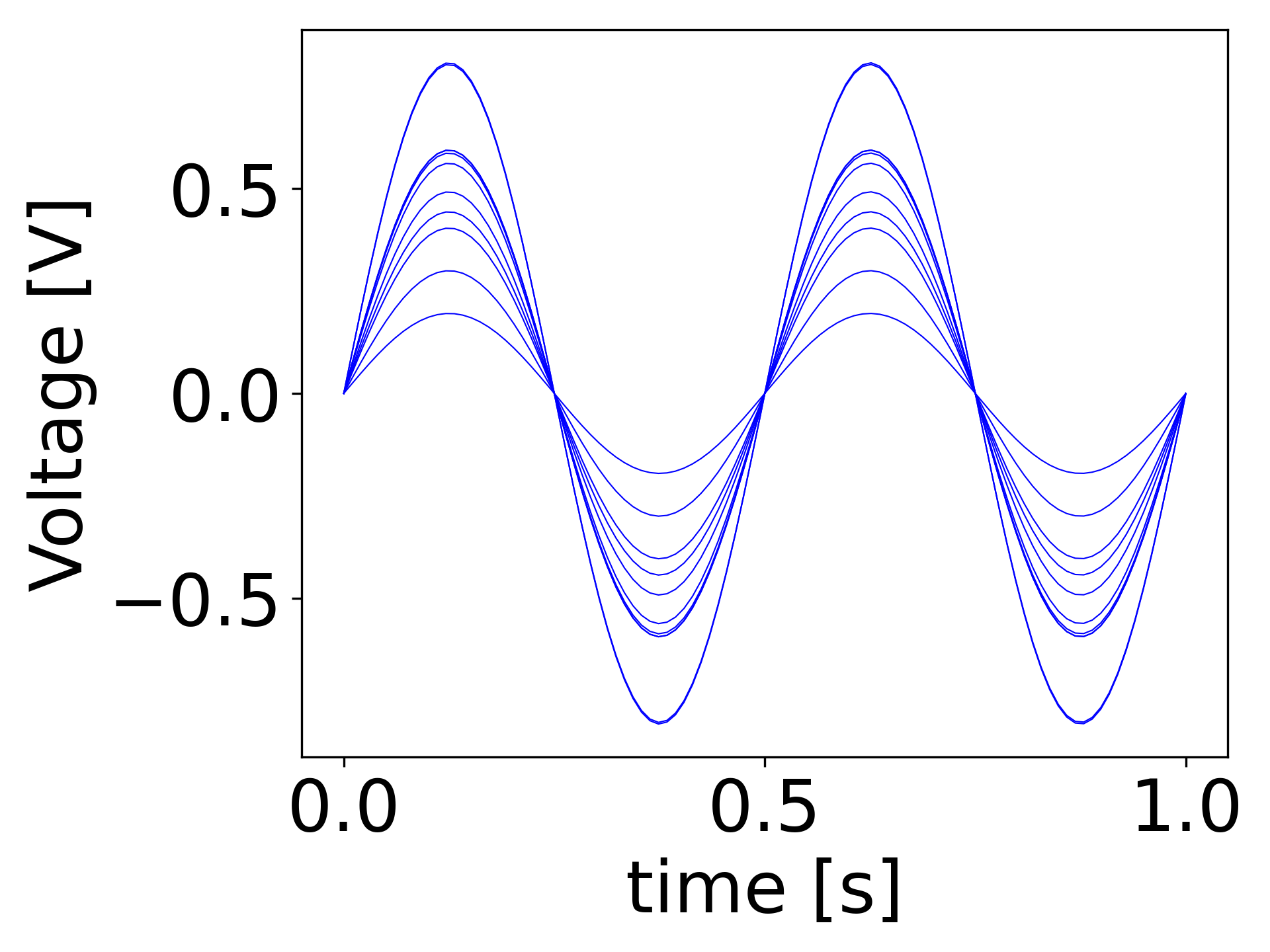}
\includegraphics[width=0.30\columnwidth]{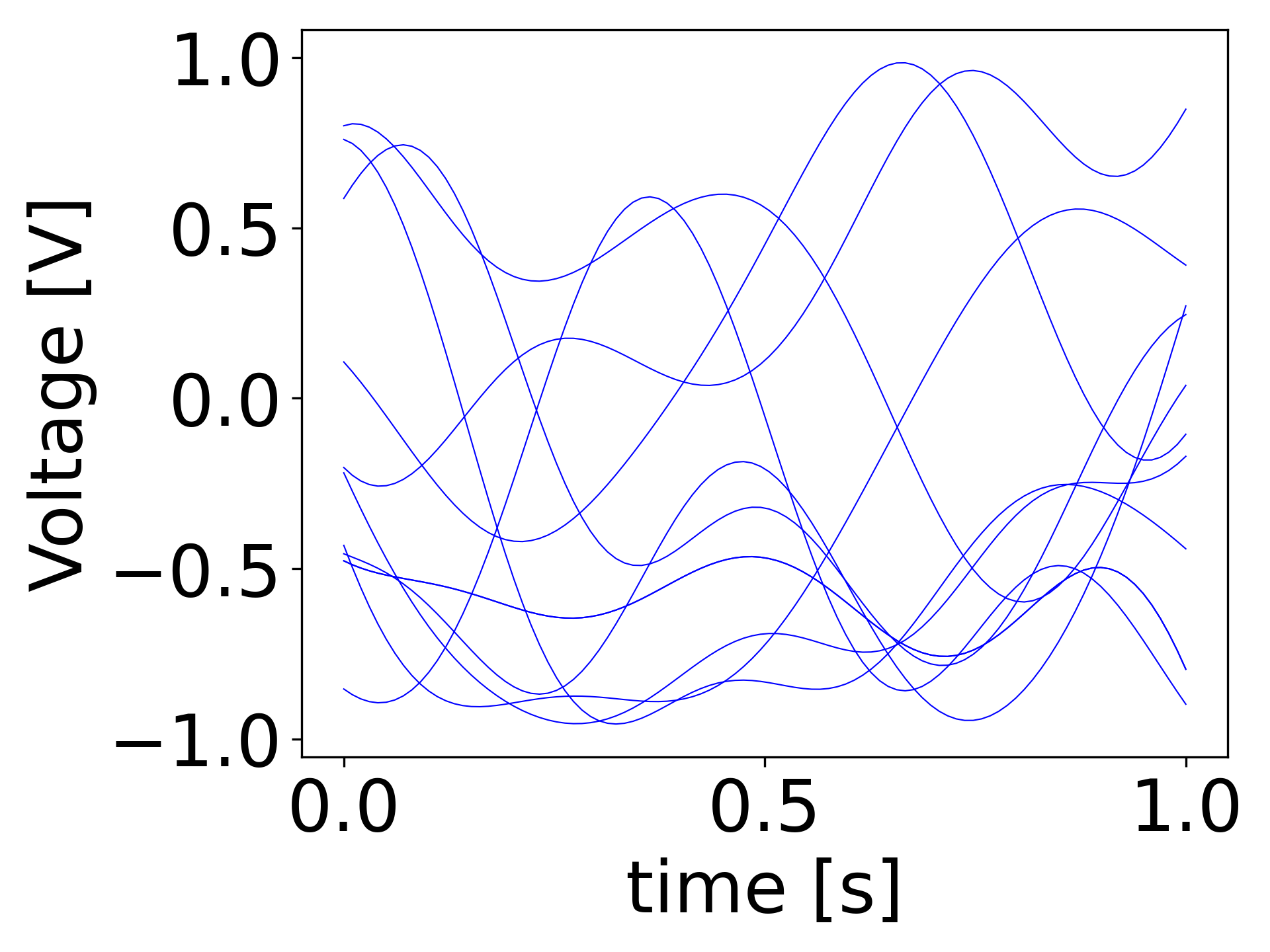}
\includegraphics[width=0.30\columnwidth]{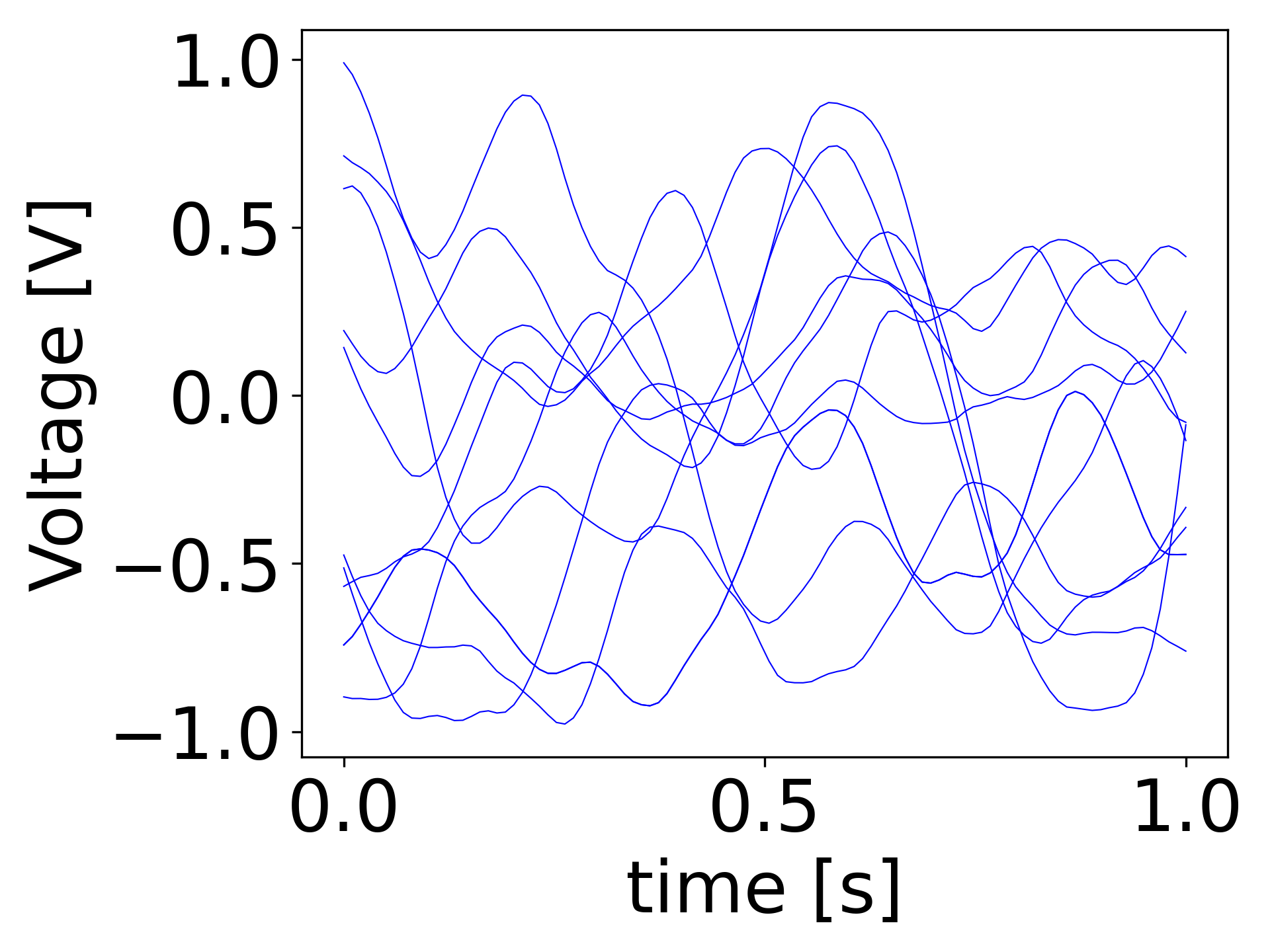}
\includegraphics[width=0.30\columnwidth]{Figures/Exp2/Exp2_Sine.png}
\includegraphics[width=0.30\columnwidth]{Figures/Exp2/Exp2_RBF.png}
\includegraphics[width=0.30\columnwidth]{Figures/Exp2/Exp2_Matern52.png}

\includegraphics[width=0.30\columnwidth]{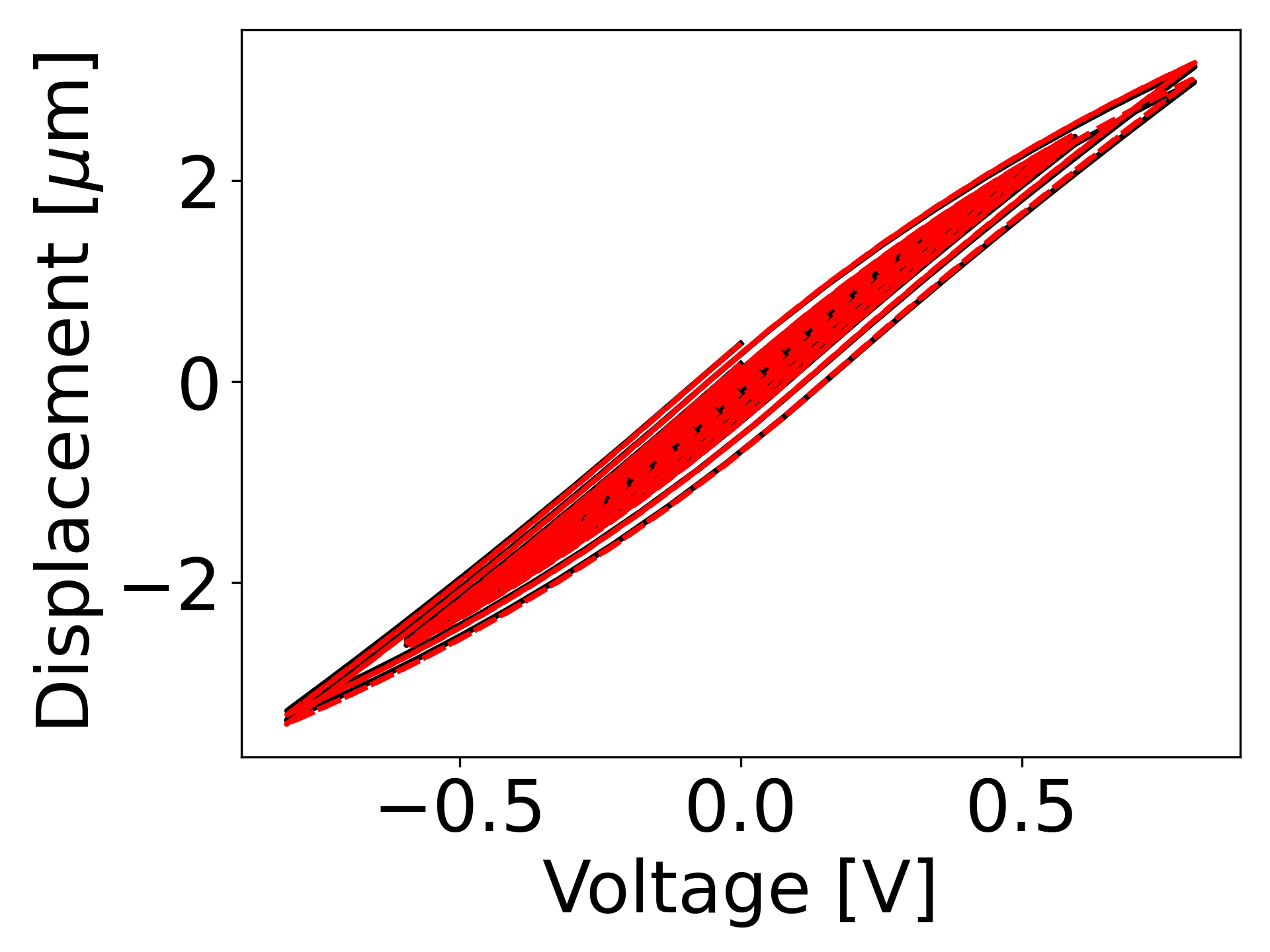}
\includegraphics[width=0.30\columnwidth]{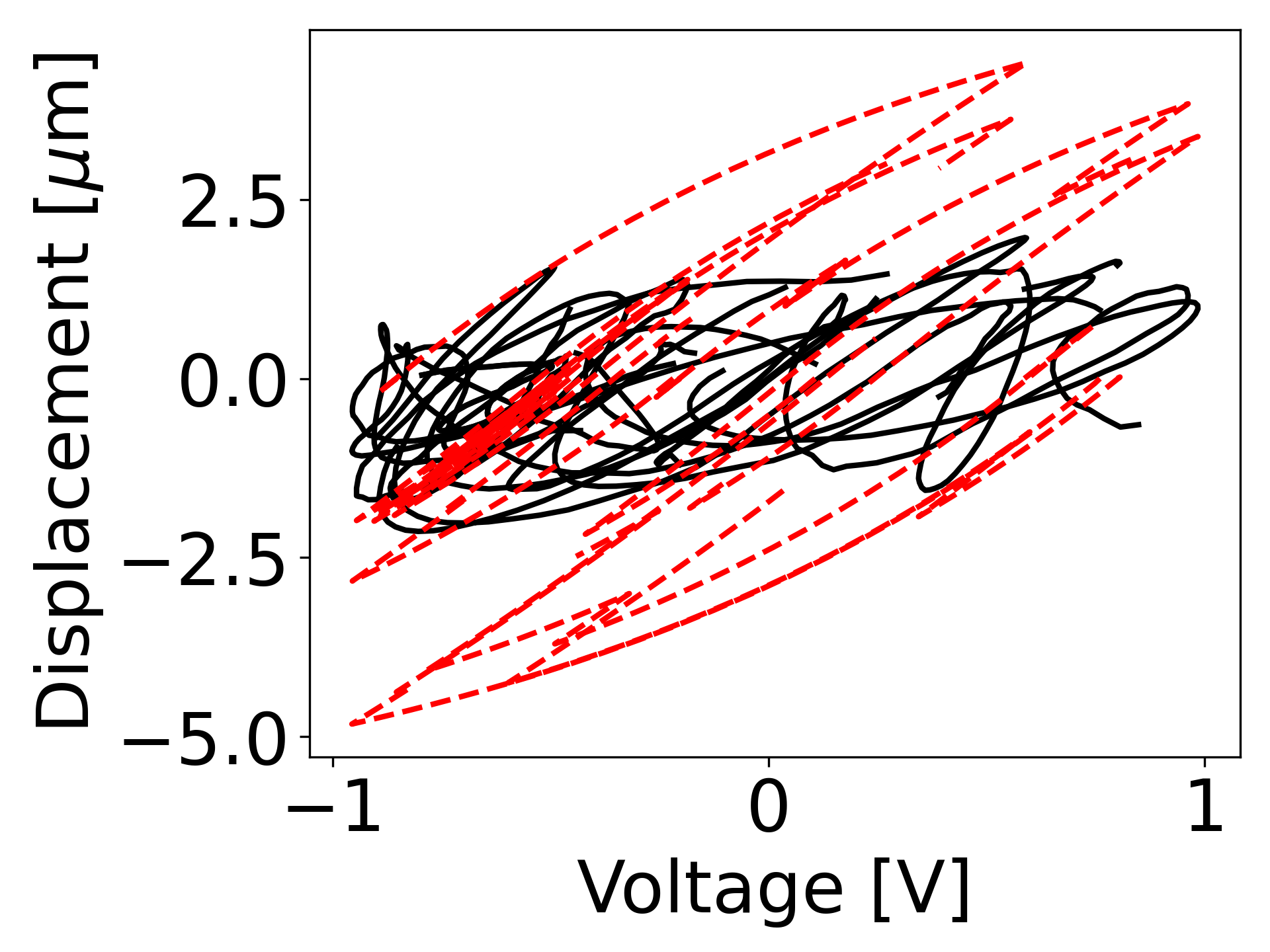}
\includegraphics[width=0.30\columnwidth]{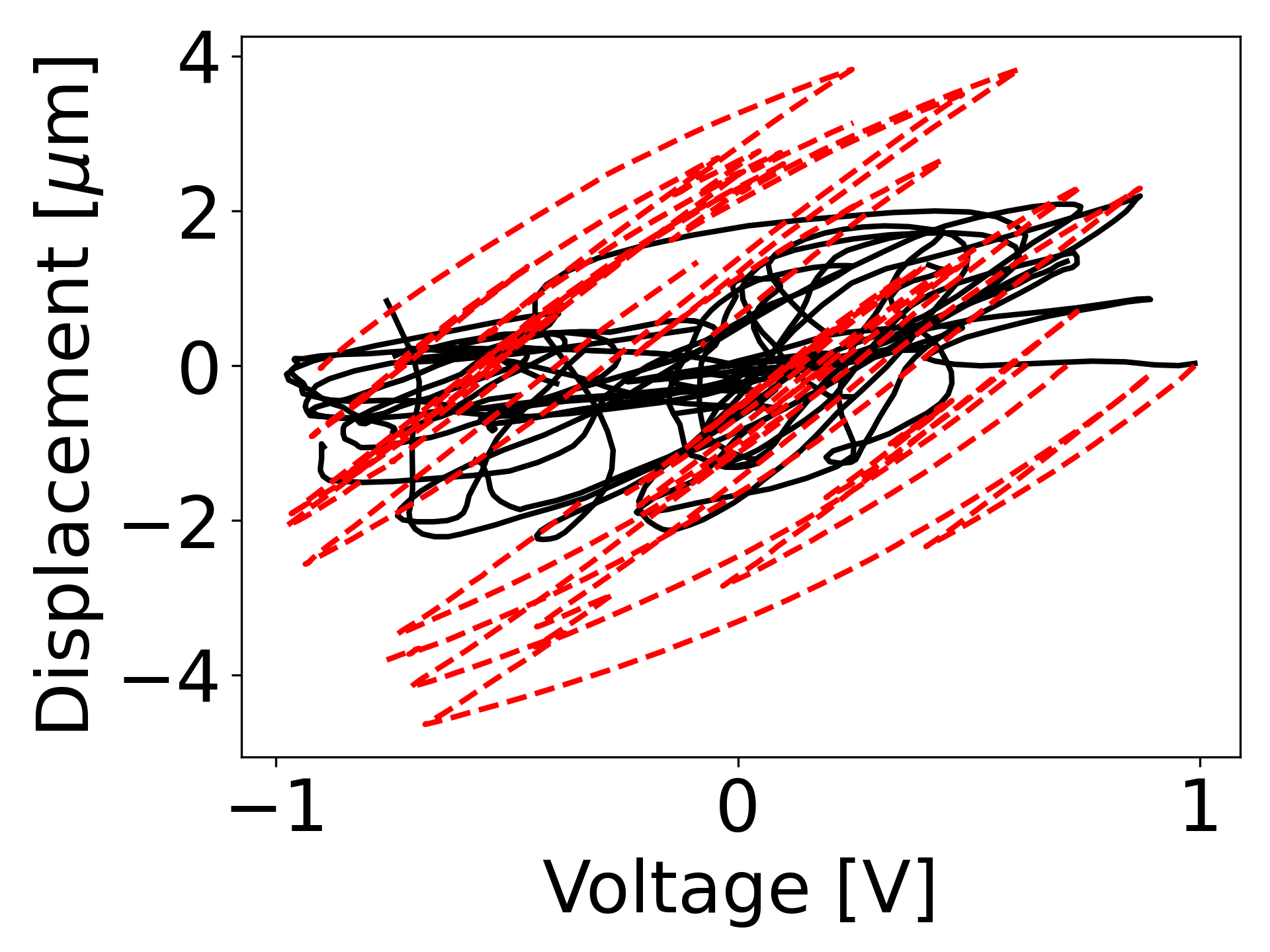}
\includegraphics[width=0.30\columnwidth]{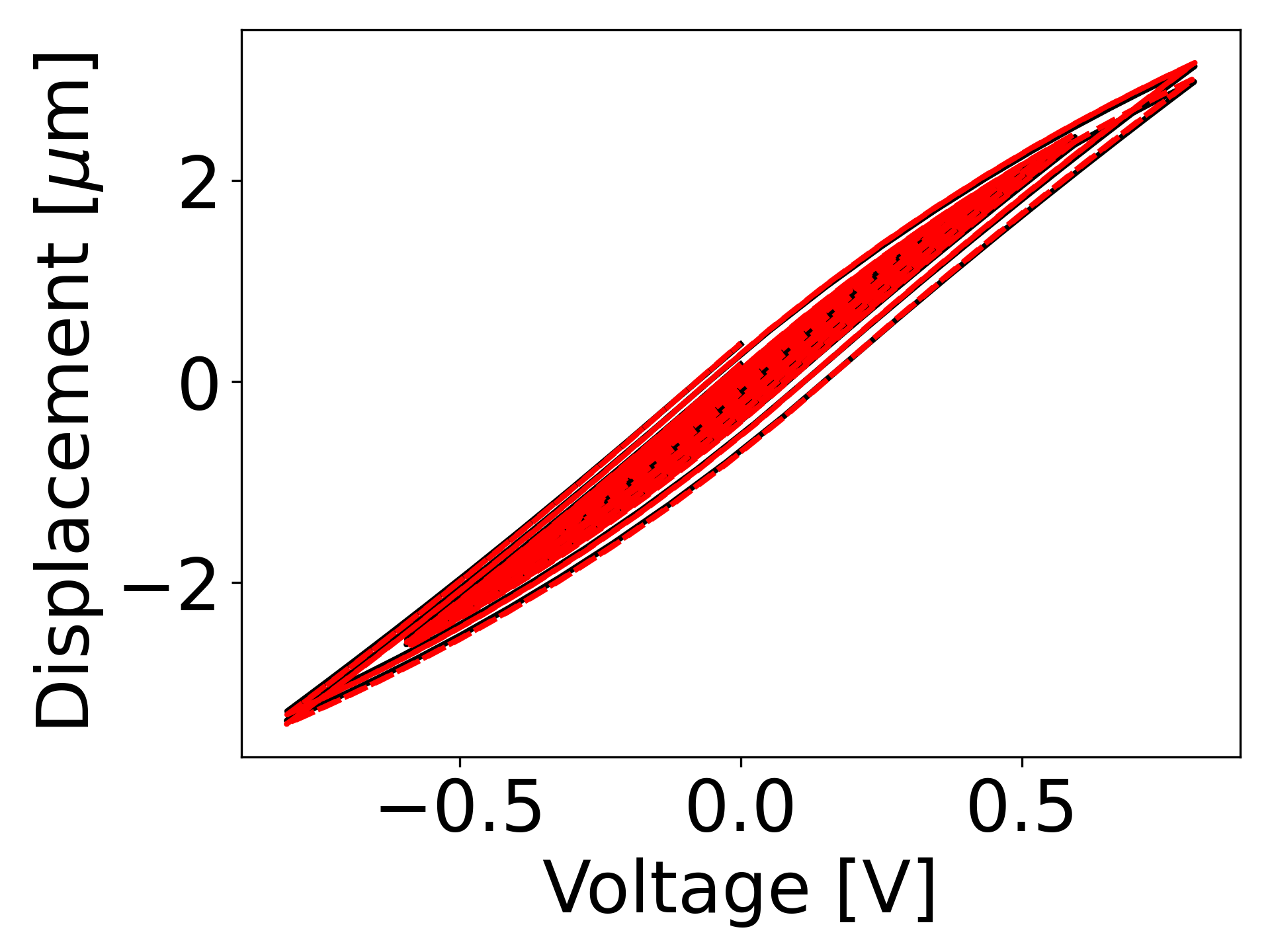}
\includegraphics[width=0.30\columnwidth]{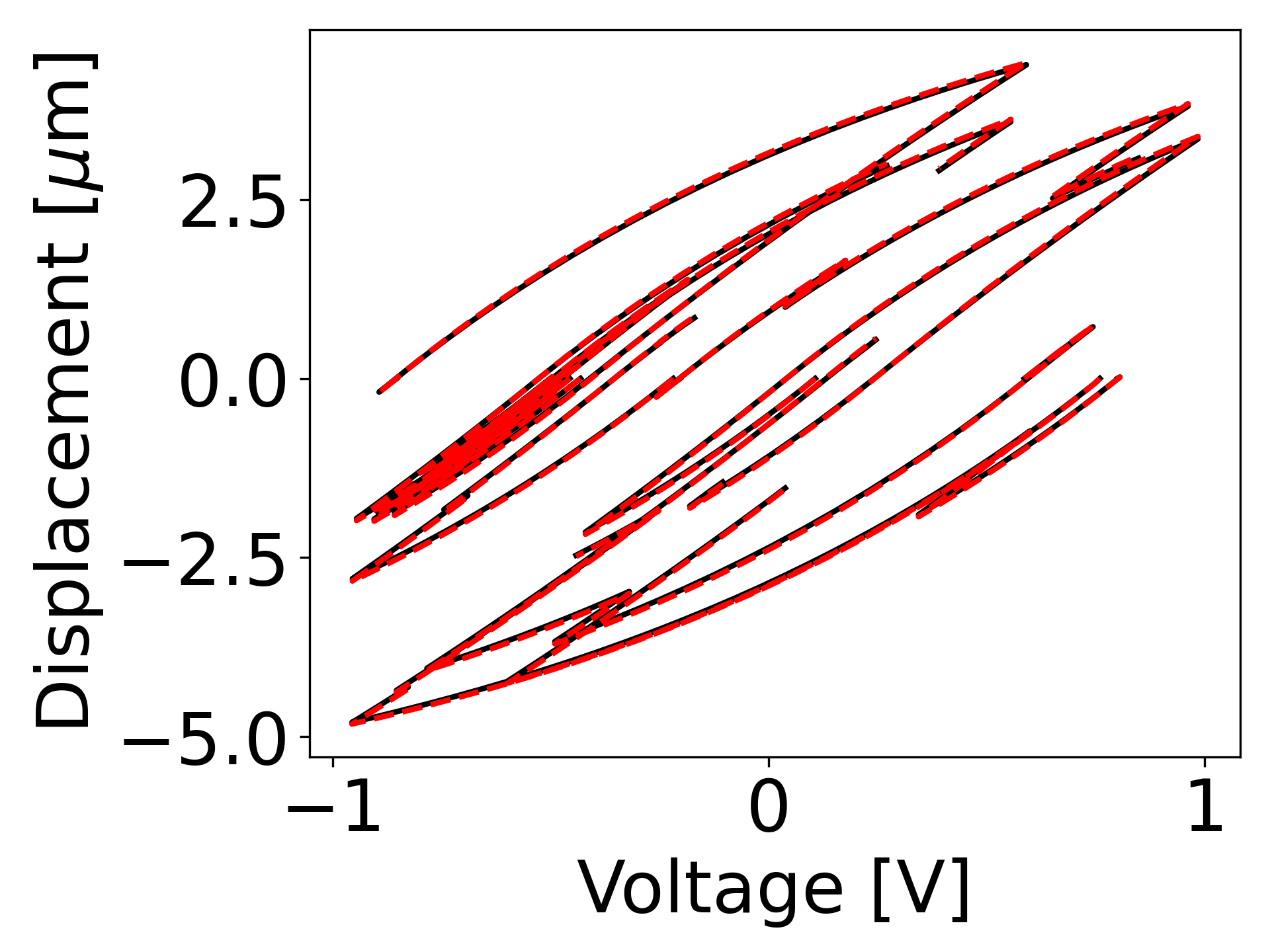}
\includegraphics[width=0.30\columnwidth]{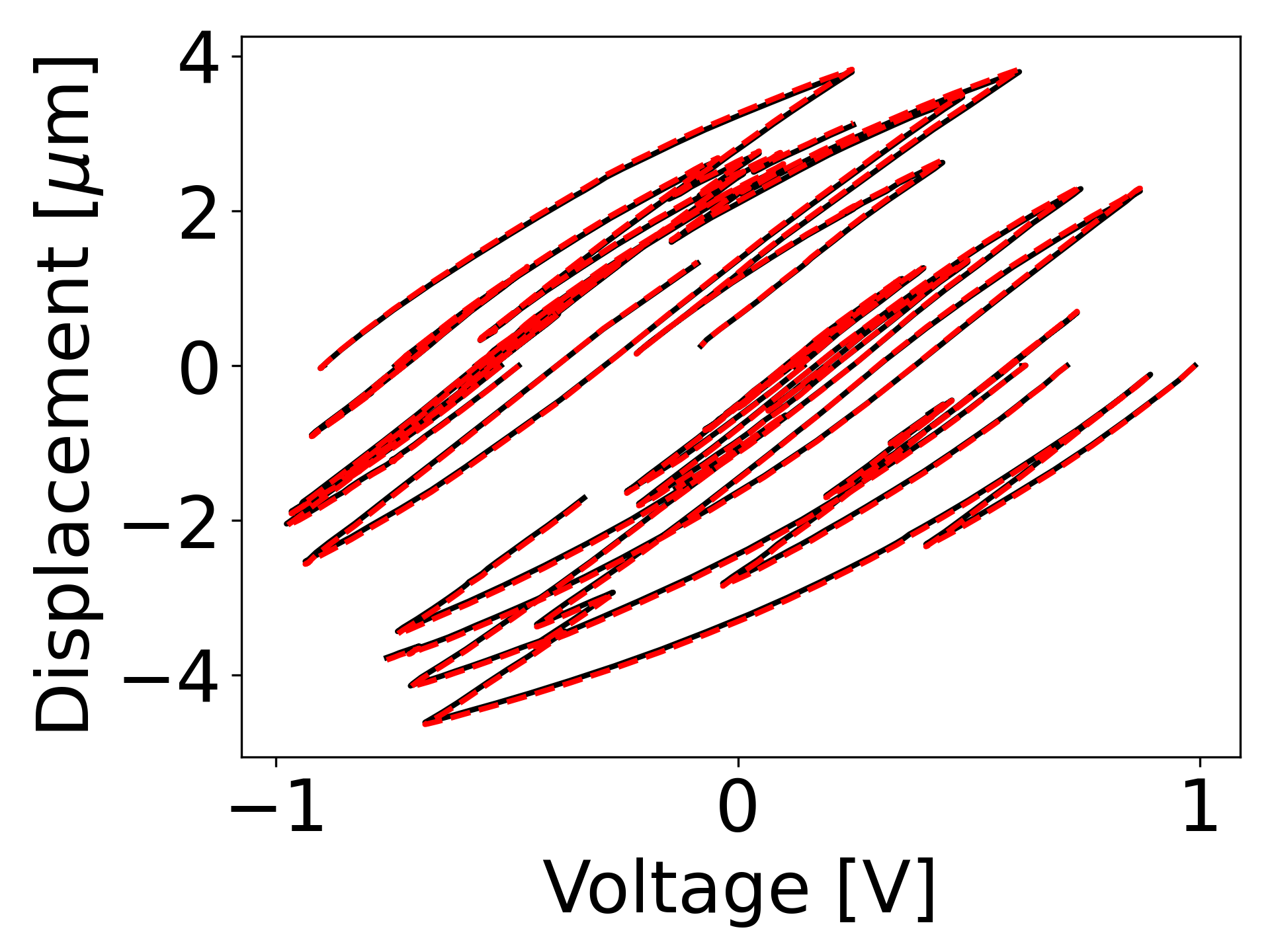}
\caption{Results for Experiment 2: \textbf{Top row: }Testing voltage fields sampled from Sine, RBF, and Matern52 kernels for FNO (first three columns) and NSO (last three columns). \textbf{Bottom row: }Hysteresis responses for FNO (first three columns) and NSO (last three columns) corresponding to the voltage fields on the Top row.}
\label{Nfig2}
\end{figure*}

Like FNO, CNO also concatenates the inputs and lifts to a latent space of size 64. The core architecture follows an encoder-bottleneck-decoder structure with four down-sampling and up-sampling blocks. Each block performs convolution operations followed by LeakyReLU activation and spatial interpolation using batch normalization. Residual connections are implemented at each encoder level using a ResNet consisting of 4 residual blocks per level and 4 in the bottleneck. The number of channels evolves with depth, starting from 8 based on a channel multiplier of 16. The decoder mirrors the encoder in reverse, concatenating skip connections through an additional convolutional block in the expansion path. The final output is projected back to a single channel. The model is trained using the Adam optimizer to optimize mean square error loss with a learning rate of 0.001, a batch size of 100, and 500 epochs.

For NSO, FNO is first trained to predict displacement with its hyperparameter configurations. Once FNO predictions are obtained, STLSQ uses voltage fields, their derivatives, absolute valued fields and their derivatives, and the displacement profiles as the basis functions. A polynomial feature expansion up to degree 2 is used for these basis functions to construct the library $\boldsymbol{\Theta}$ without including a bias term. This setup enables the generation of linear, squared, and interaction terms. The STLSQ algorithm is subjected to $\lambda\!=\!0.01$ to prune insignificant terms and a maximum of 10 iterations to ensure convergence. The resulting sparse model is then embedded into a first-order ODE and solved using \texttt{odeint} module of SciPy, with zero as initial condition. 

\subsection{Experiment 1}
\label{exp1}
For this experiment, three voltage input fields are generated: Sine, RBF, and Matern52. The Sine fields are of the form \( v(t) = A \sin(4\pi t) \), where amplitudes \( A \) are drawn uniformly from \([0, 1]\), sampled at 100 equispaced time points in \([0, 1]\). RBF and Matern52 fields are sampled from Gaussian processes defined over the same interval, with RBF and Matern52 kernels (variance \(1.0\), length scale \(0.2\)), respectively, yielding bounded and diverse non-periodic field realizations.

The sampled Sine voltages are simulated with an ODE governing the hysteresis dynamics \cite{chandra2023discovery}: $\dot{d} = 0.4 |\dot{v}| v - 0.85 |\dot{v}| d + 0.2 \dot{v}$. The data generated by simulating the ODE is used to train and test the neural operators. The equation discovered by NSO closely resembles the structure of the actual equation, $\dot{d} = 0.39 |\dot{v}| v - 0.83 |\dot{v}| d + 0.2 \dot{v}$. This identified model offers interpretability owing to its white-box nature, facilitating the analysis of the system under different scenarios. This interpretability is challenging with the baseline operators. The identified ODE is further used to test its generalization ability under RBF and Matern test field conditions.

\subsection{Experiment 2}
\label{exp2}
The voltage fields used in this experiment are generated similarly to experiment 1. However, the randomness inherent in sampling leads to different realizations of the fields across the two experiments.

For this experiment, the reference hysteresis system is governed by the equation $\dot{d} = 5 \dot{v} - 0.25 |\dot{v}| d - 0.5 \dot{v} |d|$, incorporating both linear and nonlinear terms. The model identified by NSO is given by $\dot{d} = 4.91 \dot{v} - 0.25 |\dot{v}| d - 0.48 \dot{v} |d|$, which preserves the same mathematical structure and captures the physical behavior of the system. Importantly, this model, being a white-box ODE, offers interpretability compared to traditional black-box operators. This identified equation is further used for testing the RBF and Matern test fields.

\begin{figure*} 
\centering
\includegraphics[width=0.30\columnwidth]{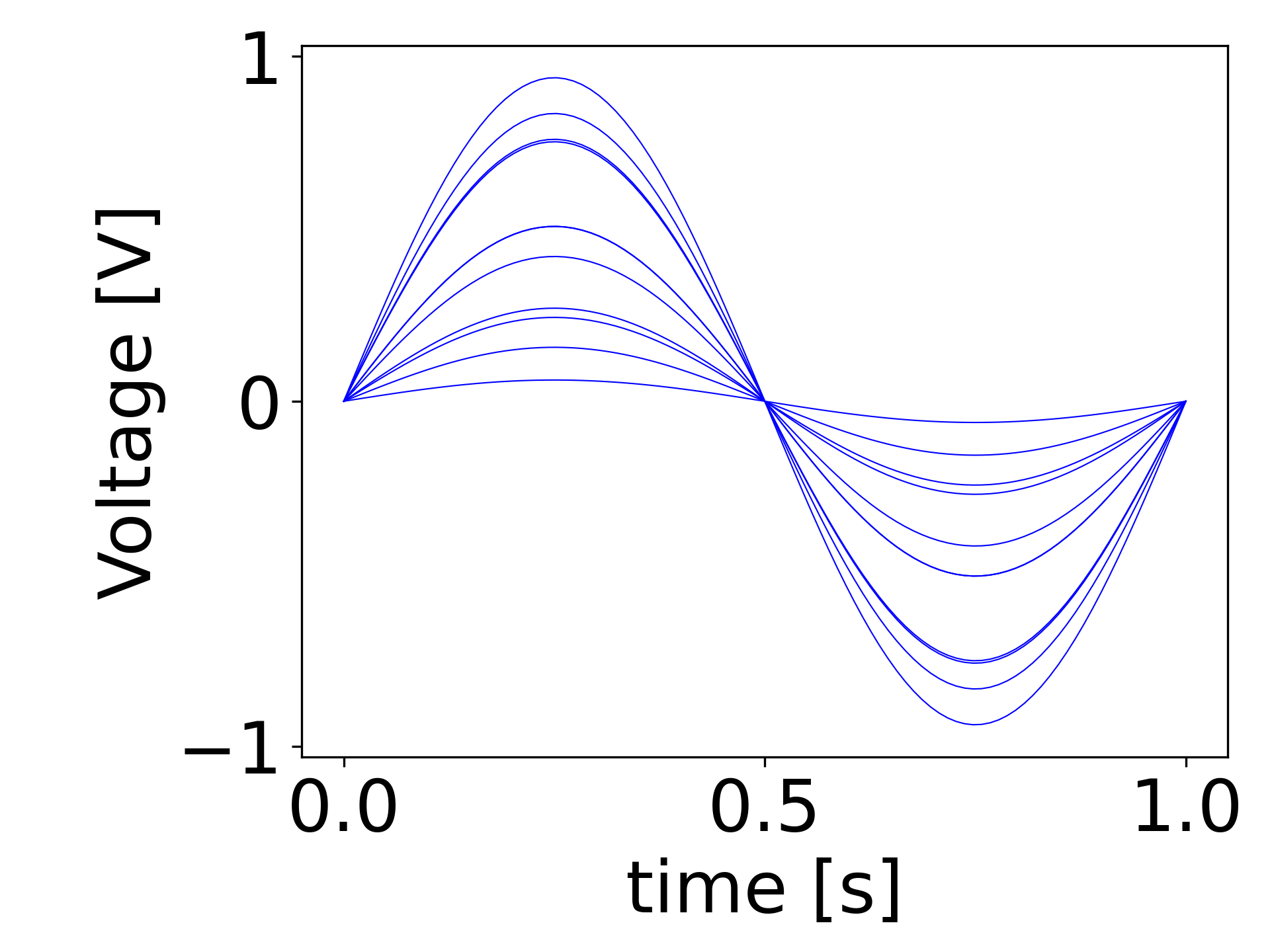}
\includegraphics[width=0.30\columnwidth]{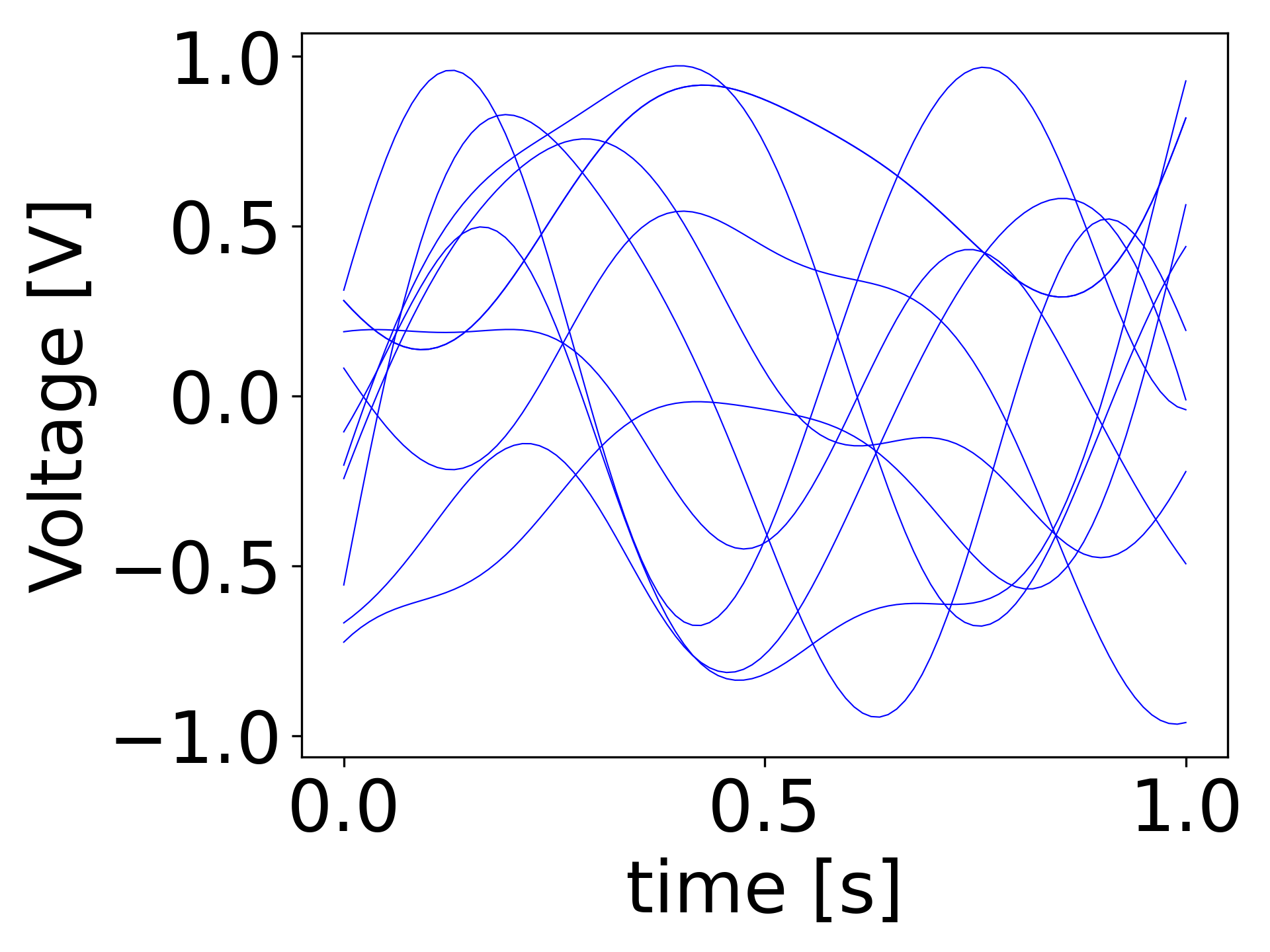}
\includegraphics[width=0.30\columnwidth]{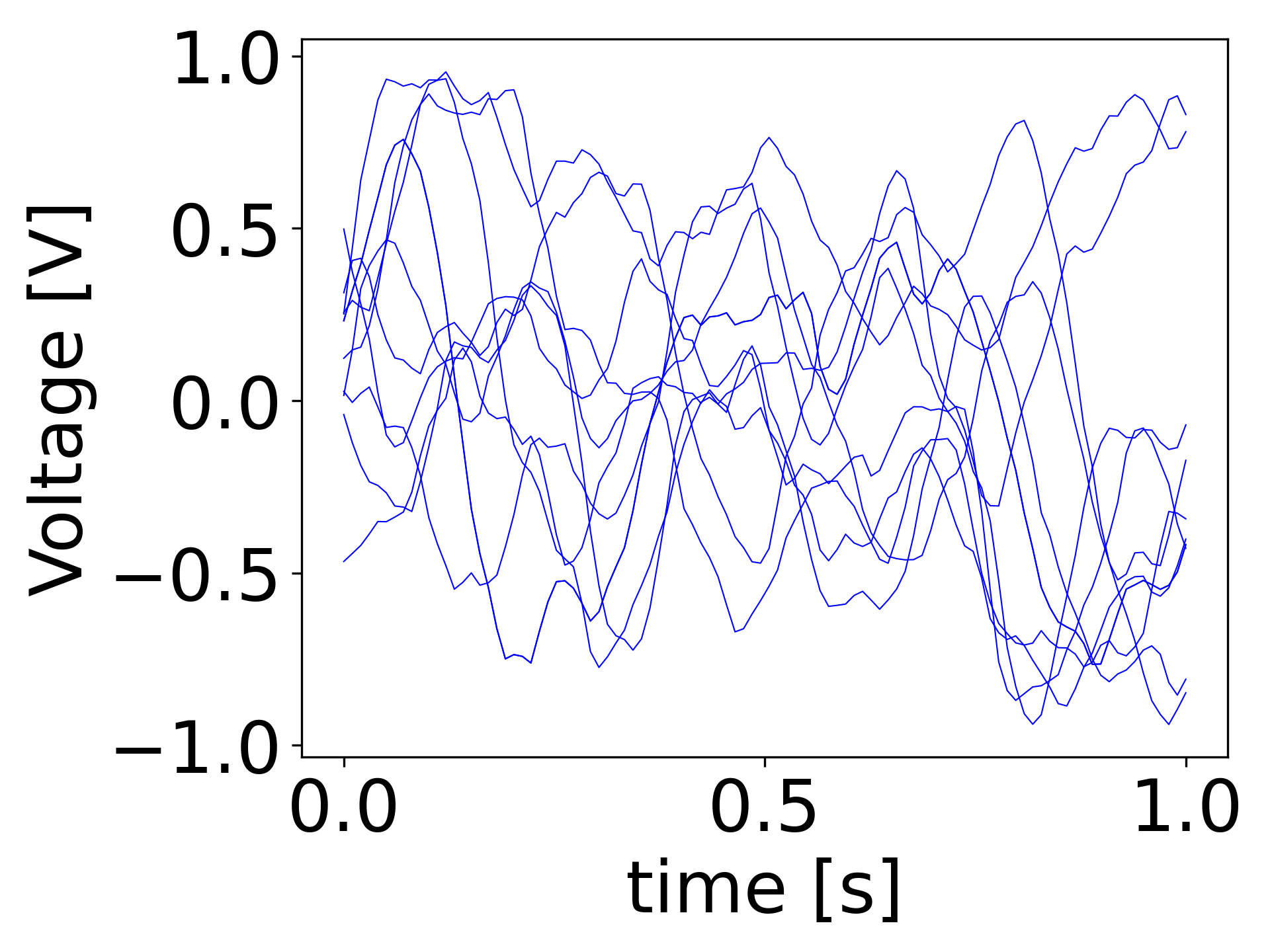}
\includegraphics[width=0.30\columnwidth]{Figures/Exp3/Exp3_Sine.png}
\includegraphics[width=0.30\columnwidth]{Figures/Exp3/Exp3_RBF.png}
\includegraphics[width=0.30\columnwidth]{Figures/Exp3/Exp3_Matern32.png}

\includegraphics[width=0.30\columnwidth]{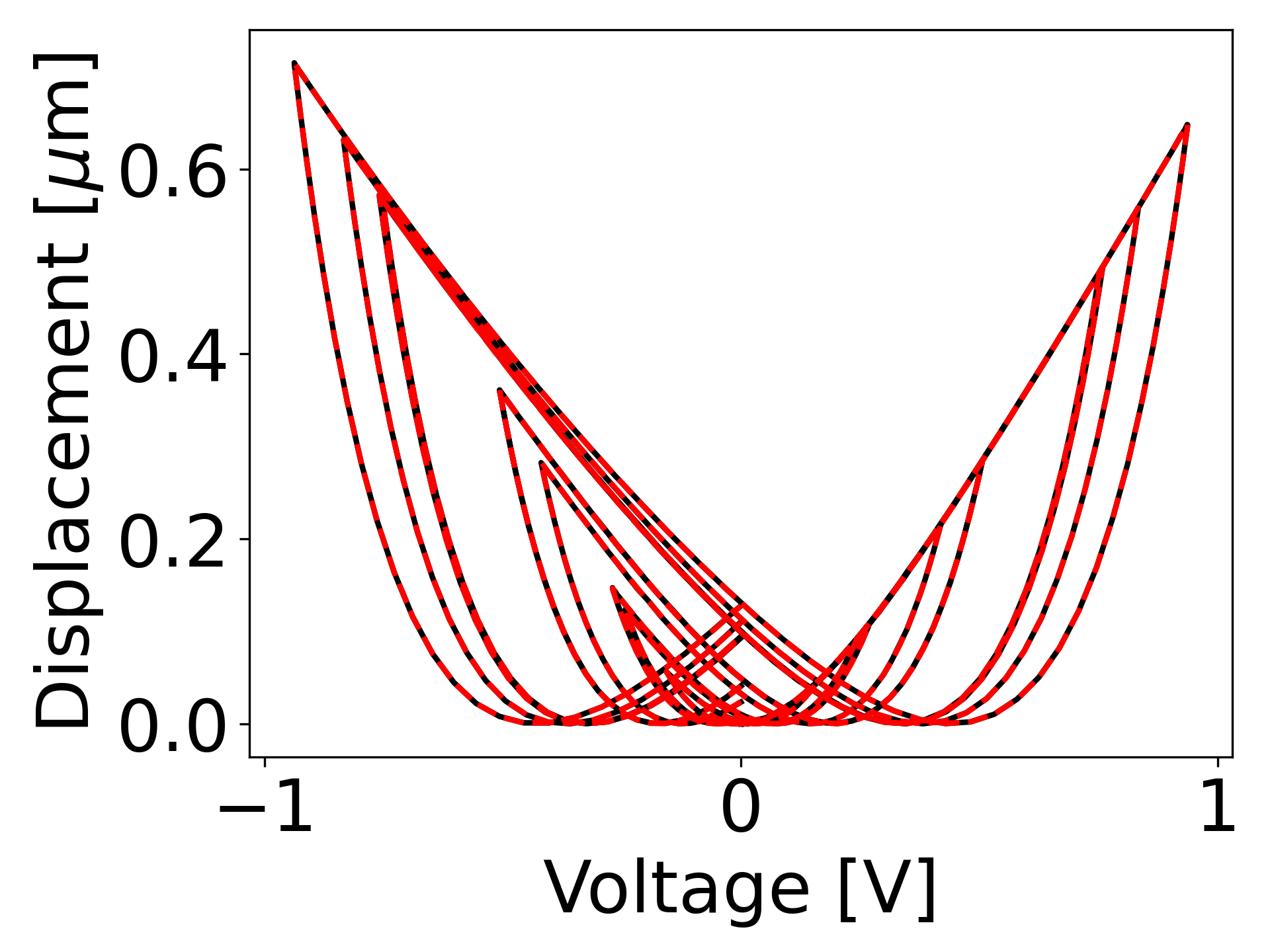}
\includegraphics[width=0.30\columnwidth]{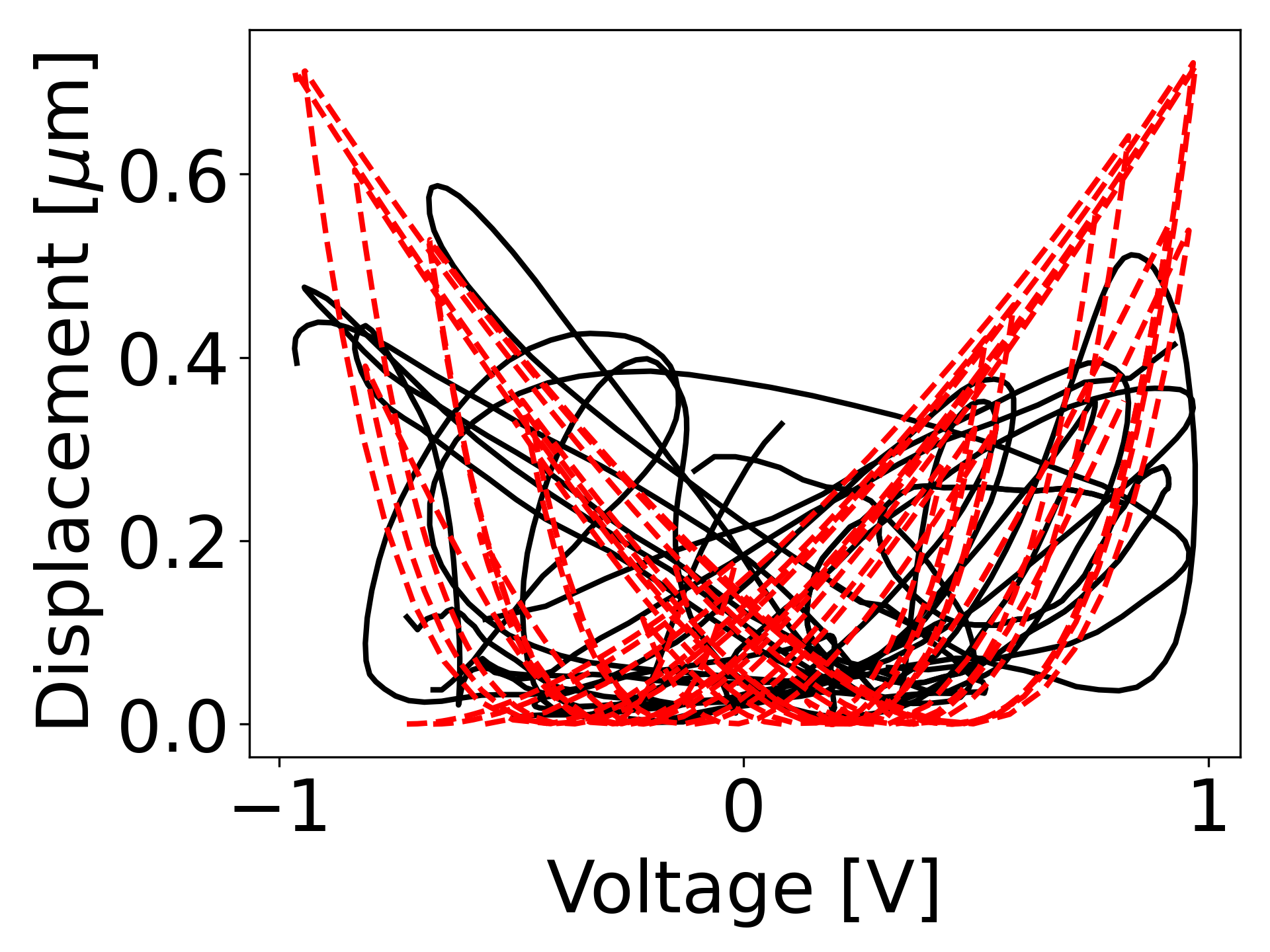}
\includegraphics[width=0.30\columnwidth]{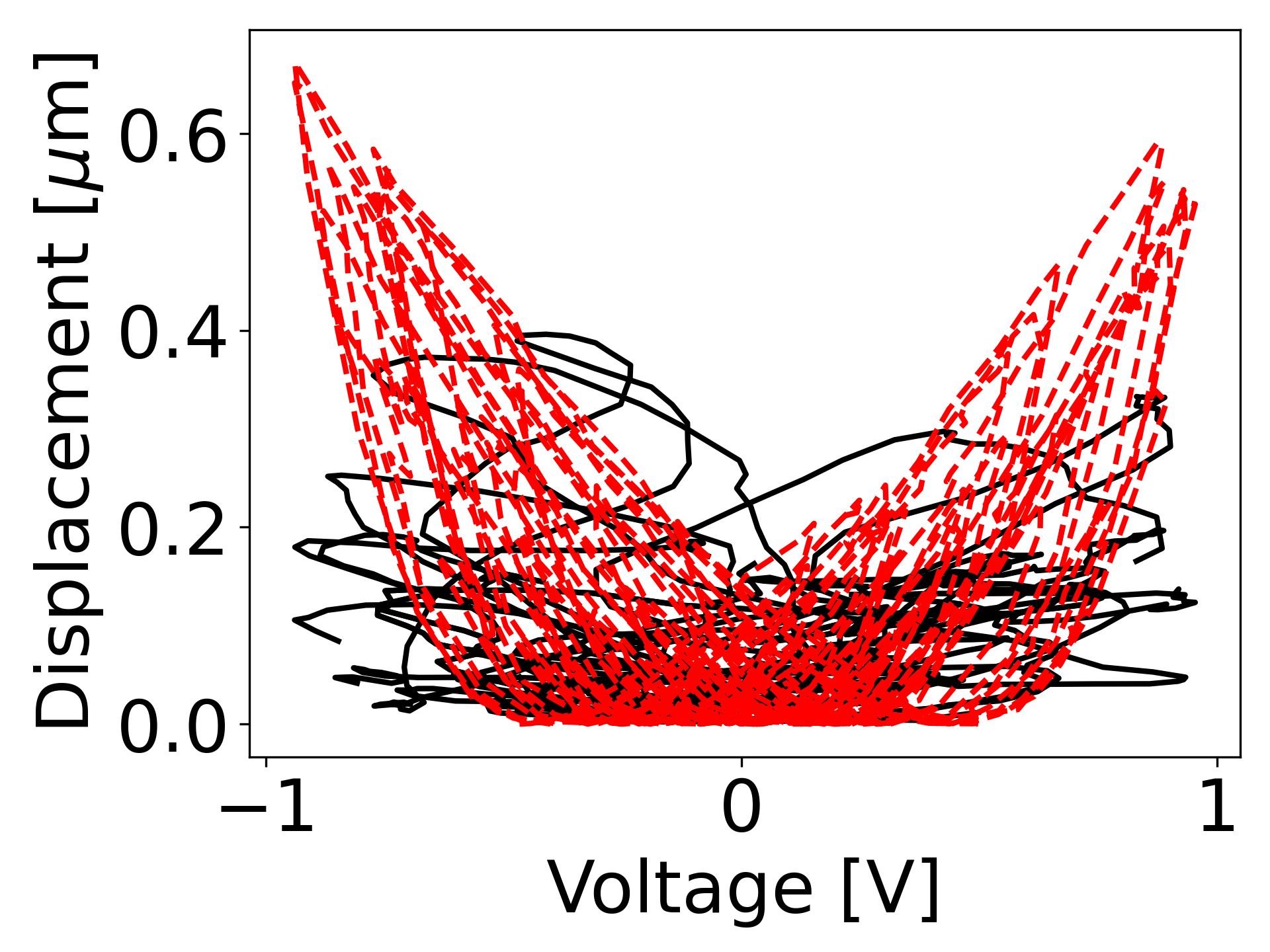}
\includegraphics[width=0.30\columnwidth]{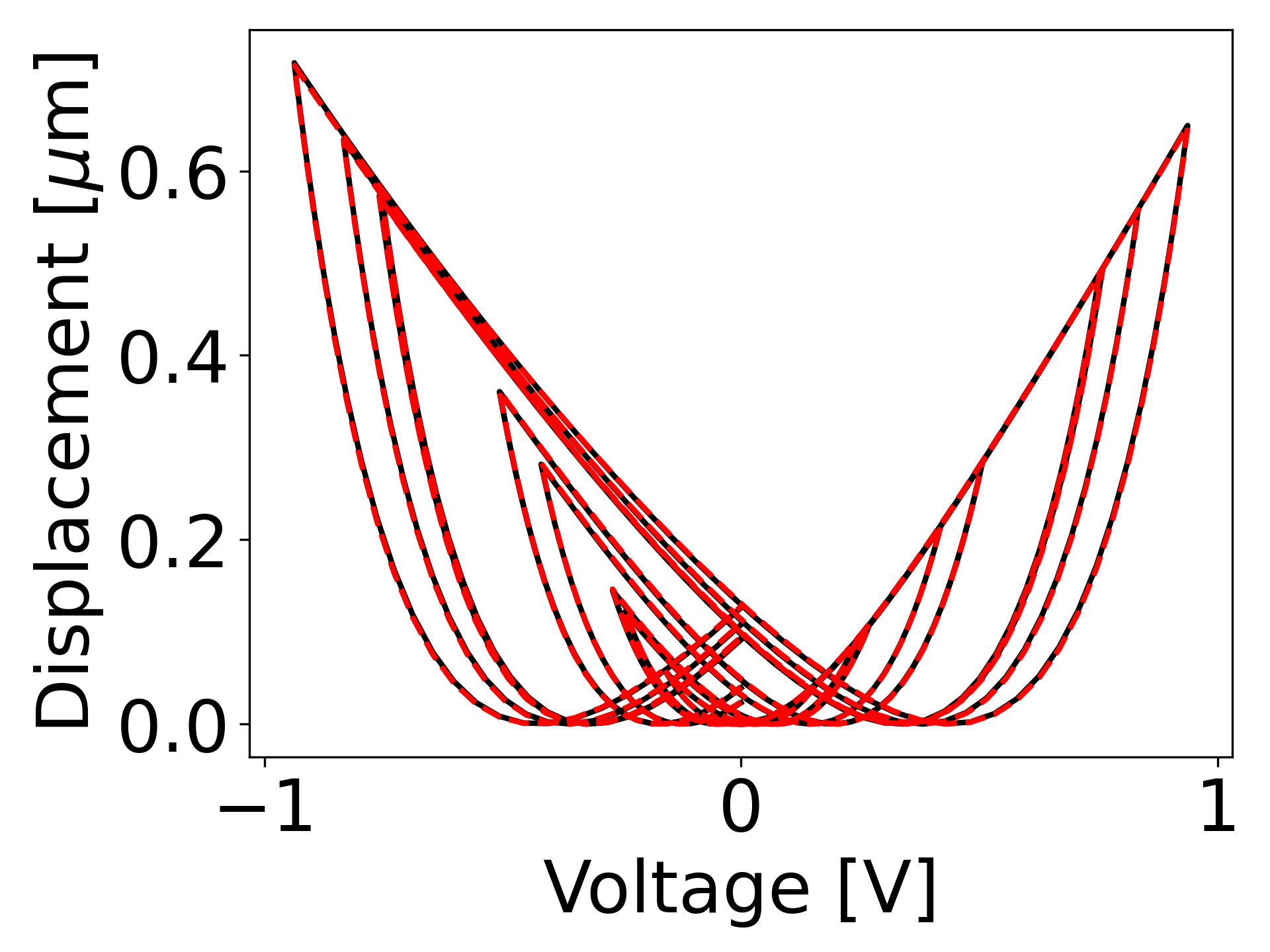}
\includegraphics[width=0.30\columnwidth]{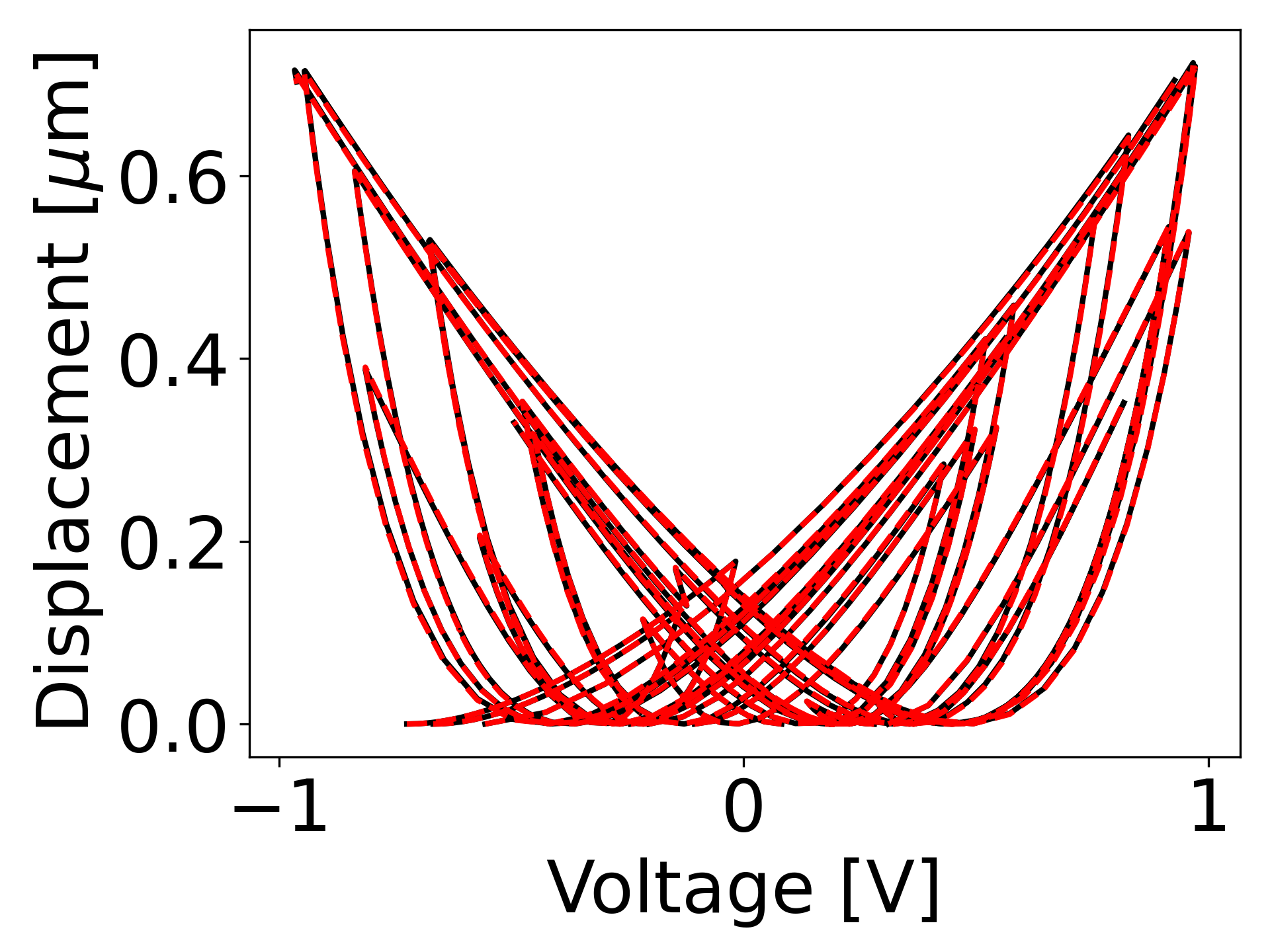}
\includegraphics[width=0.30\columnwidth]{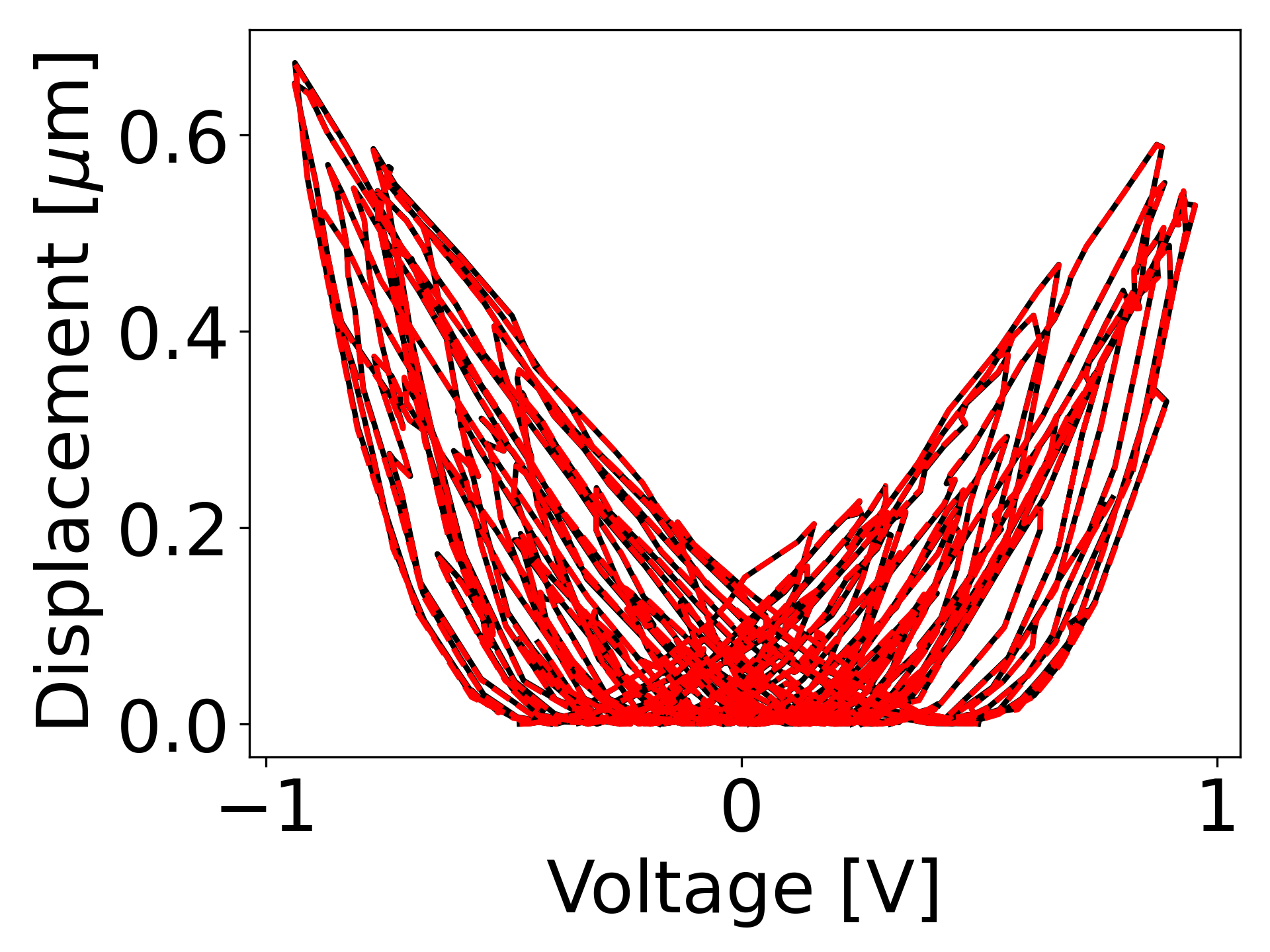}
\caption{Results for Experiment 3: \textbf{Top row: }Testing voltage fields sampled from Sine, RBF, and Matern32 kernels for FNO (first three columns) and NSO (last three columns). \textbf{Bottom row: }Hysteresis responses for FNO (first three columns) and NSO (last three columns) corresponding to the voltage fields on the Top row.}
\label{Nfig3}
\end{figure*}

\begin{figure*} 
\centering
\includegraphics[width=0.30\columnwidth]{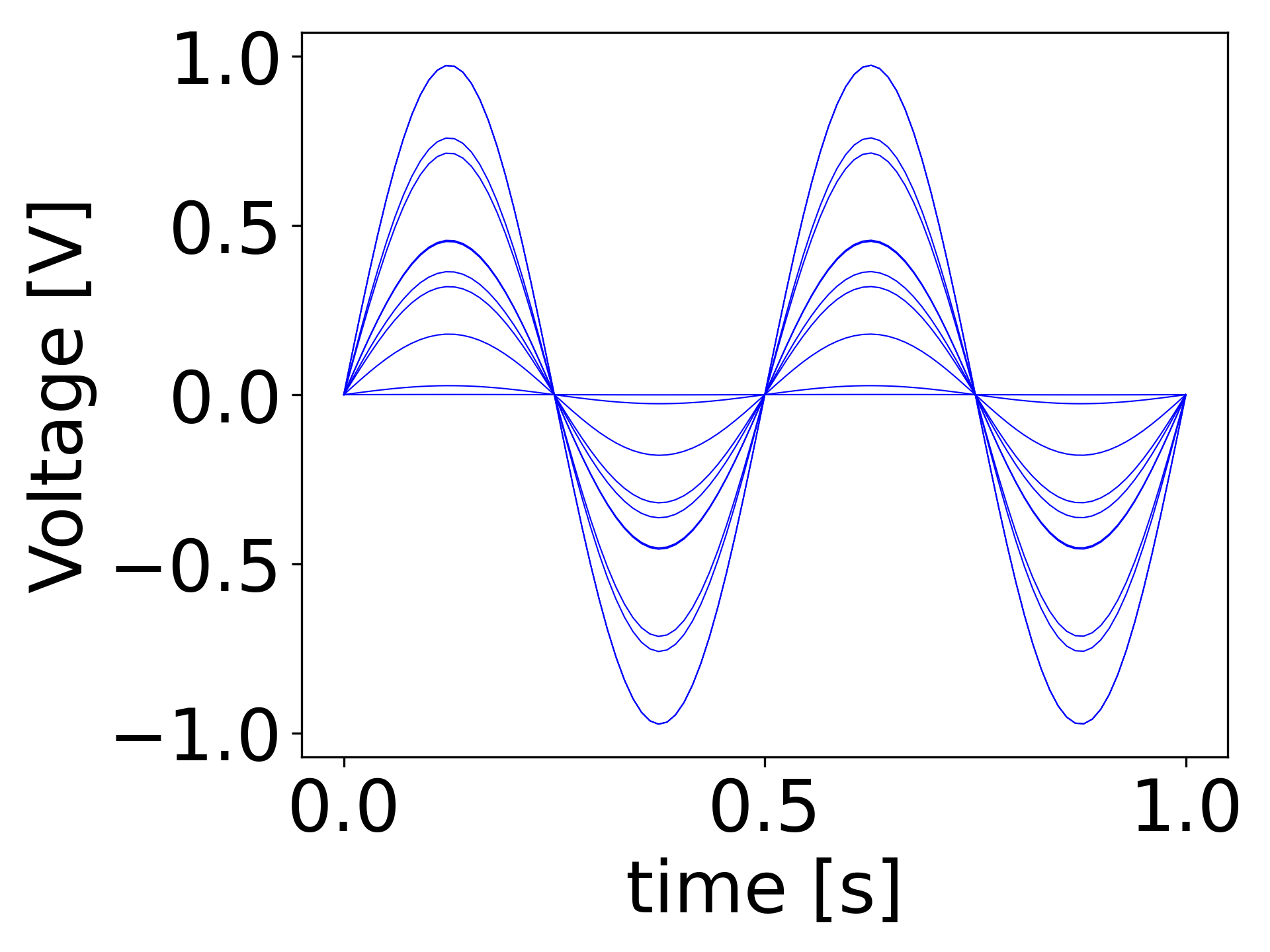}
\includegraphics[width=0.30\columnwidth]{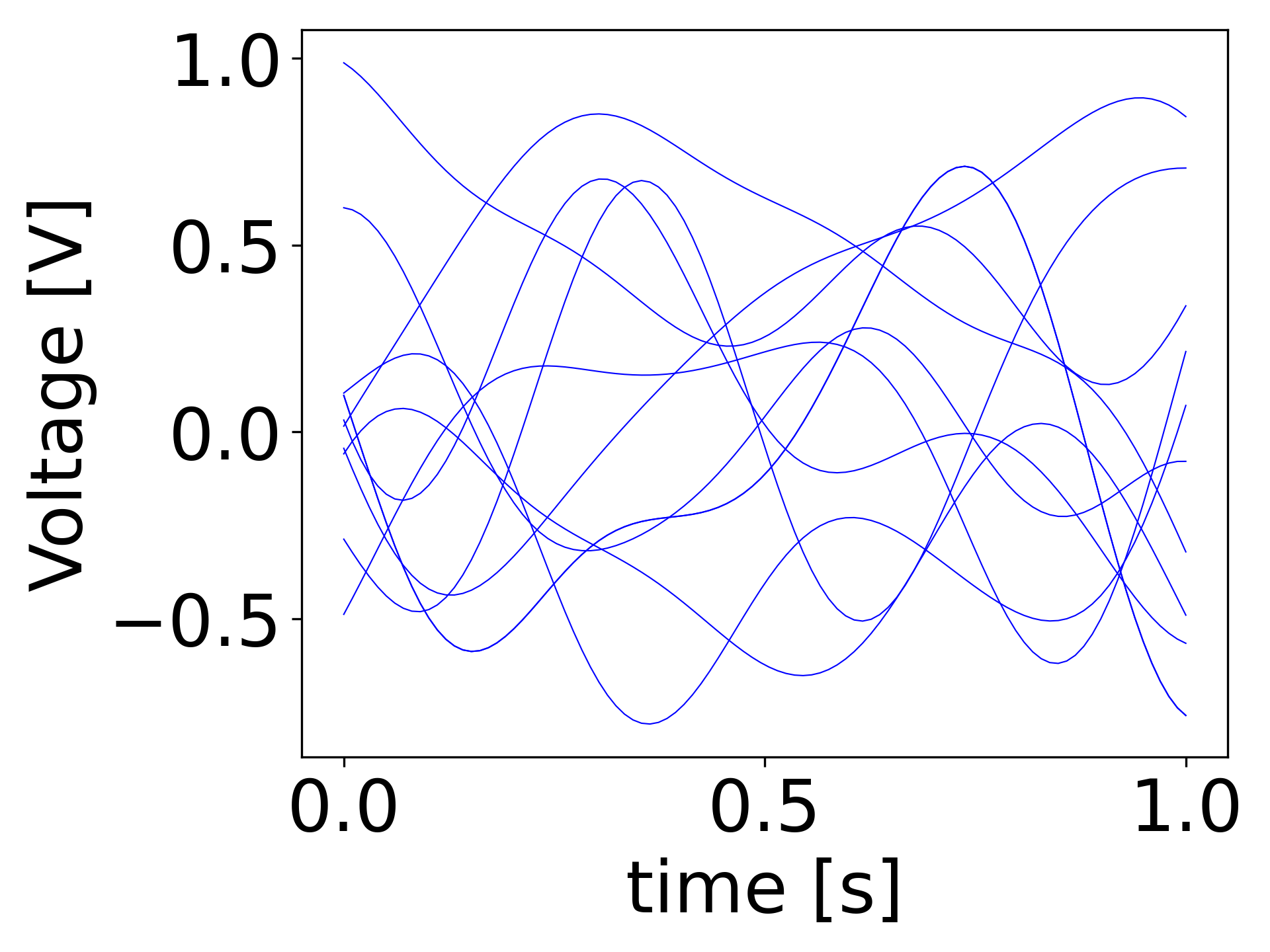}
\includegraphics[width=0.30\columnwidth]{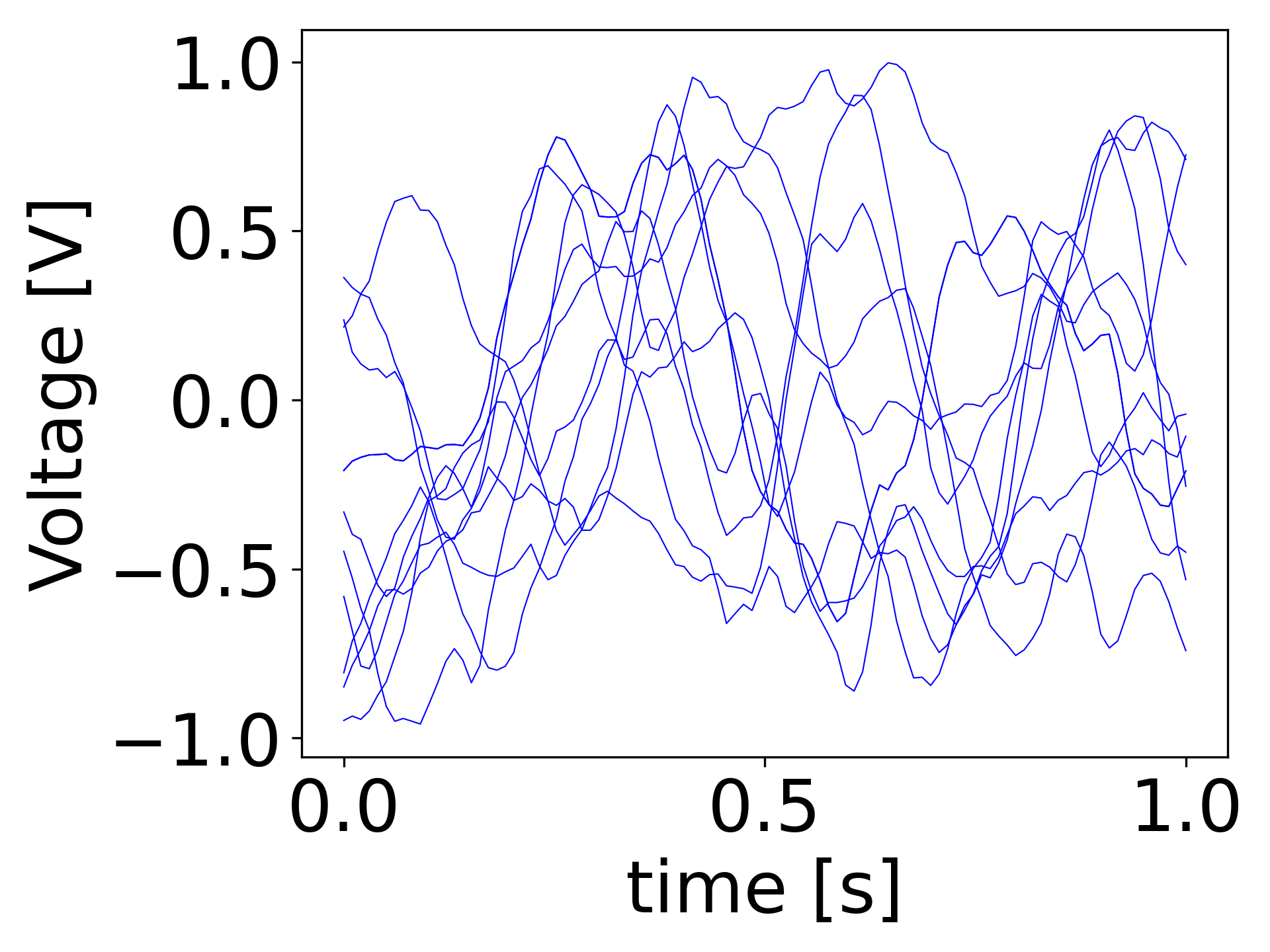}
\includegraphics[width=0.30\columnwidth]{Figures/Exp4/Exp4_Sine.png}
\includegraphics[width=0.30\columnwidth]{Figures/Exp4/Exp4_RBF.png}
\includegraphics[width=0.30\columnwidth]{Figures/Exp4/Exp4_Matern32.png}

\includegraphics[width=0.30\columnwidth]{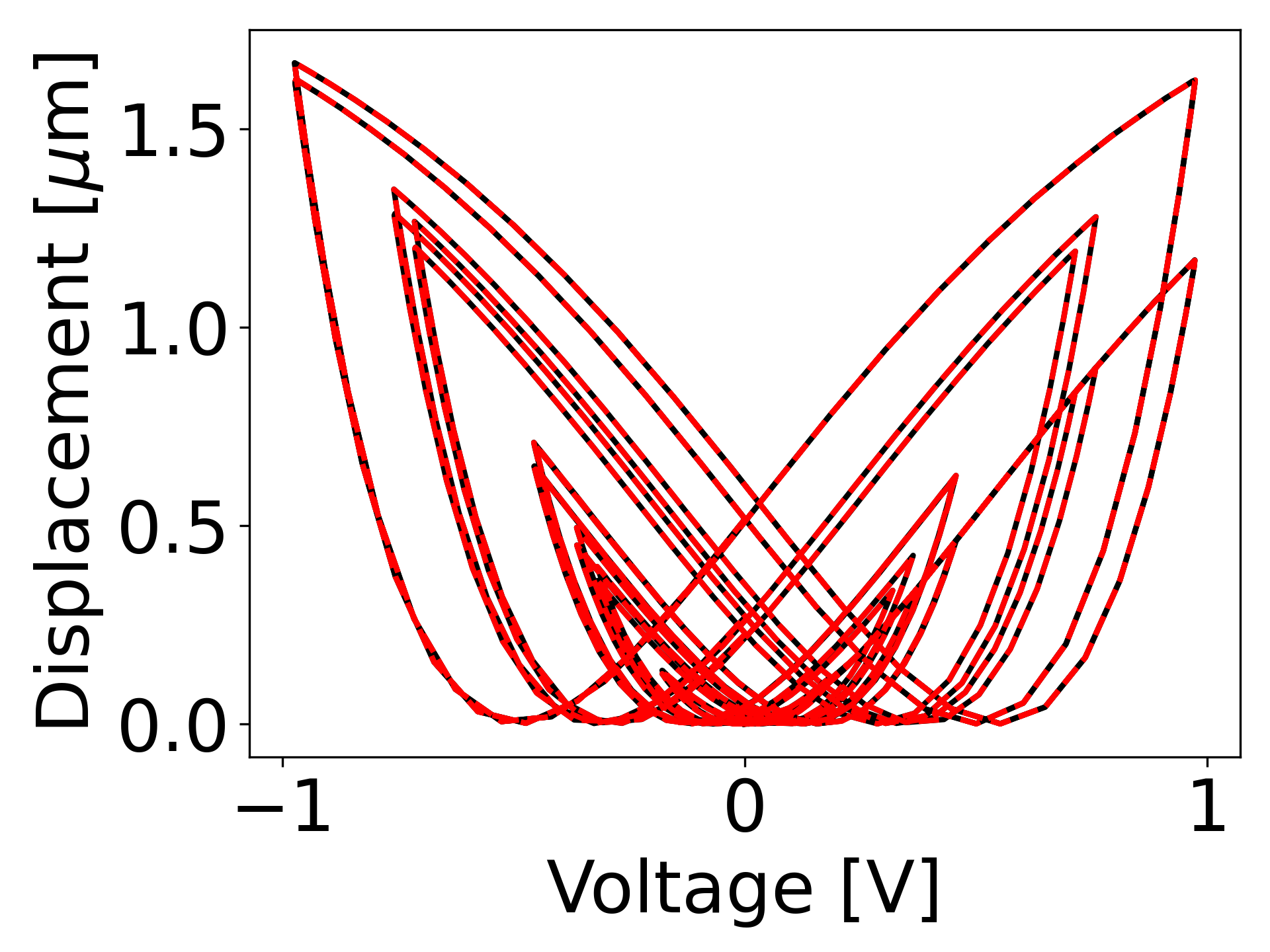}
\includegraphics[width=0.30\columnwidth]{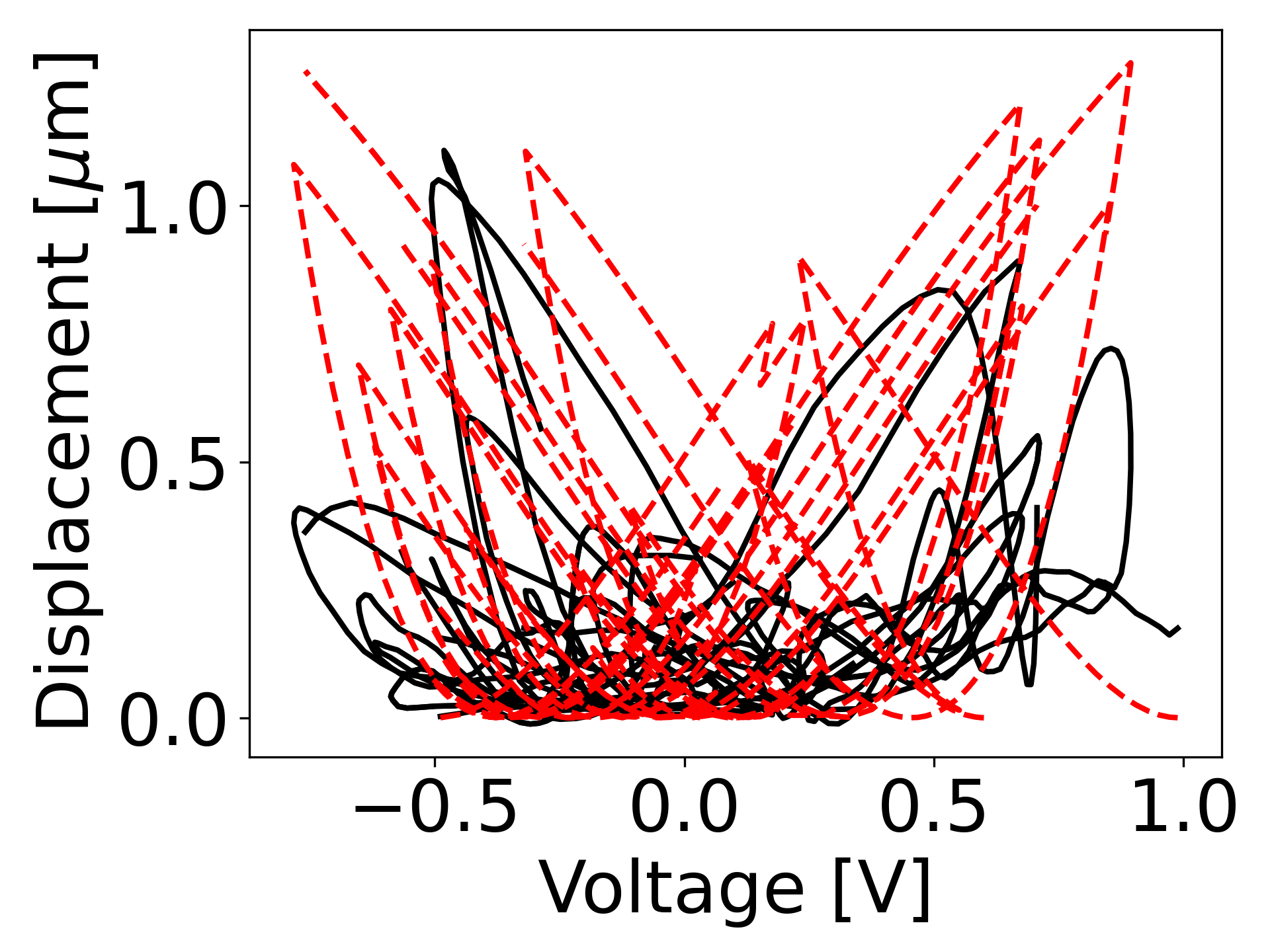}
\includegraphics[width=0.30\columnwidth]{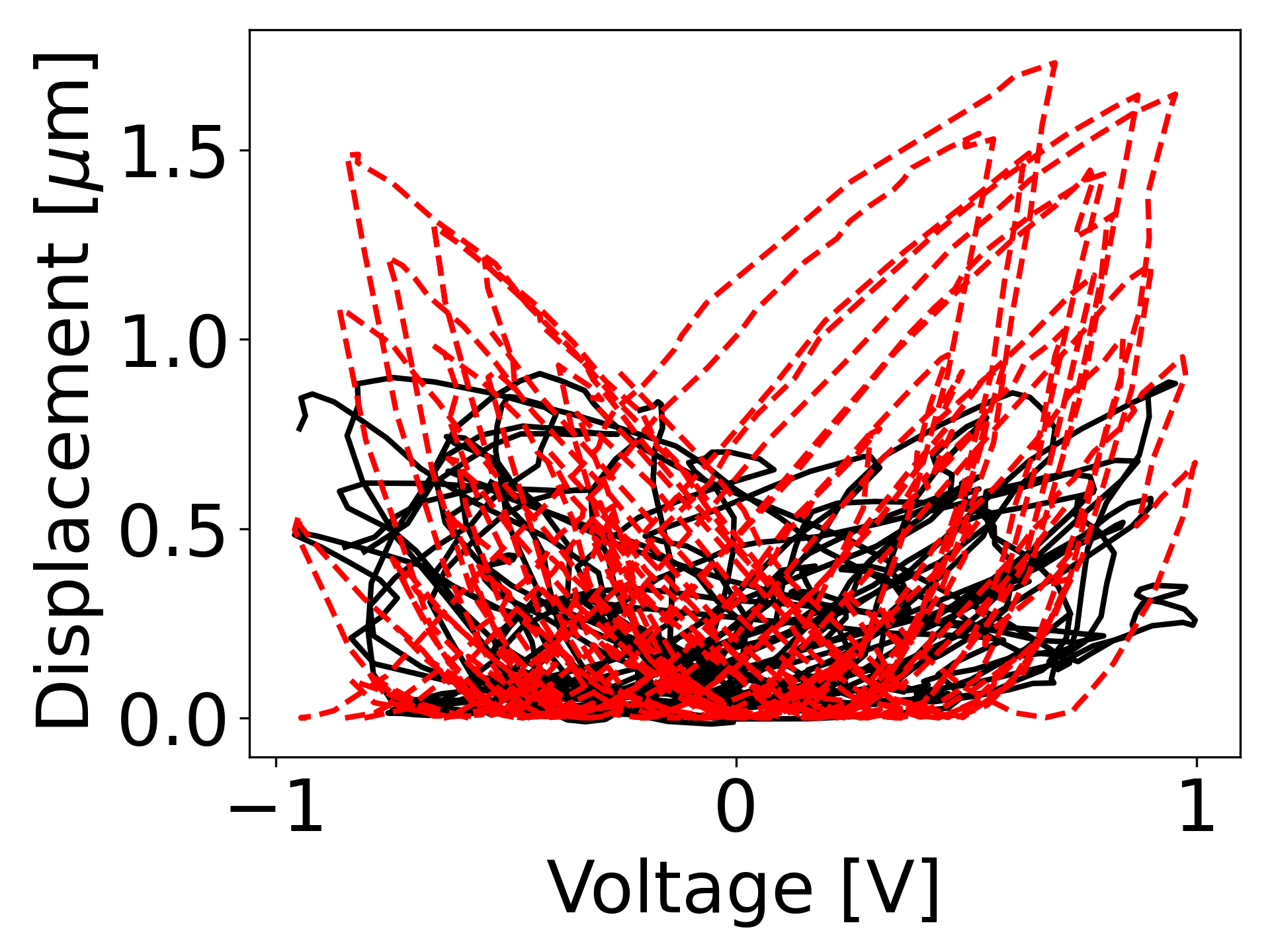}
\includegraphics[width=0.30\columnwidth]{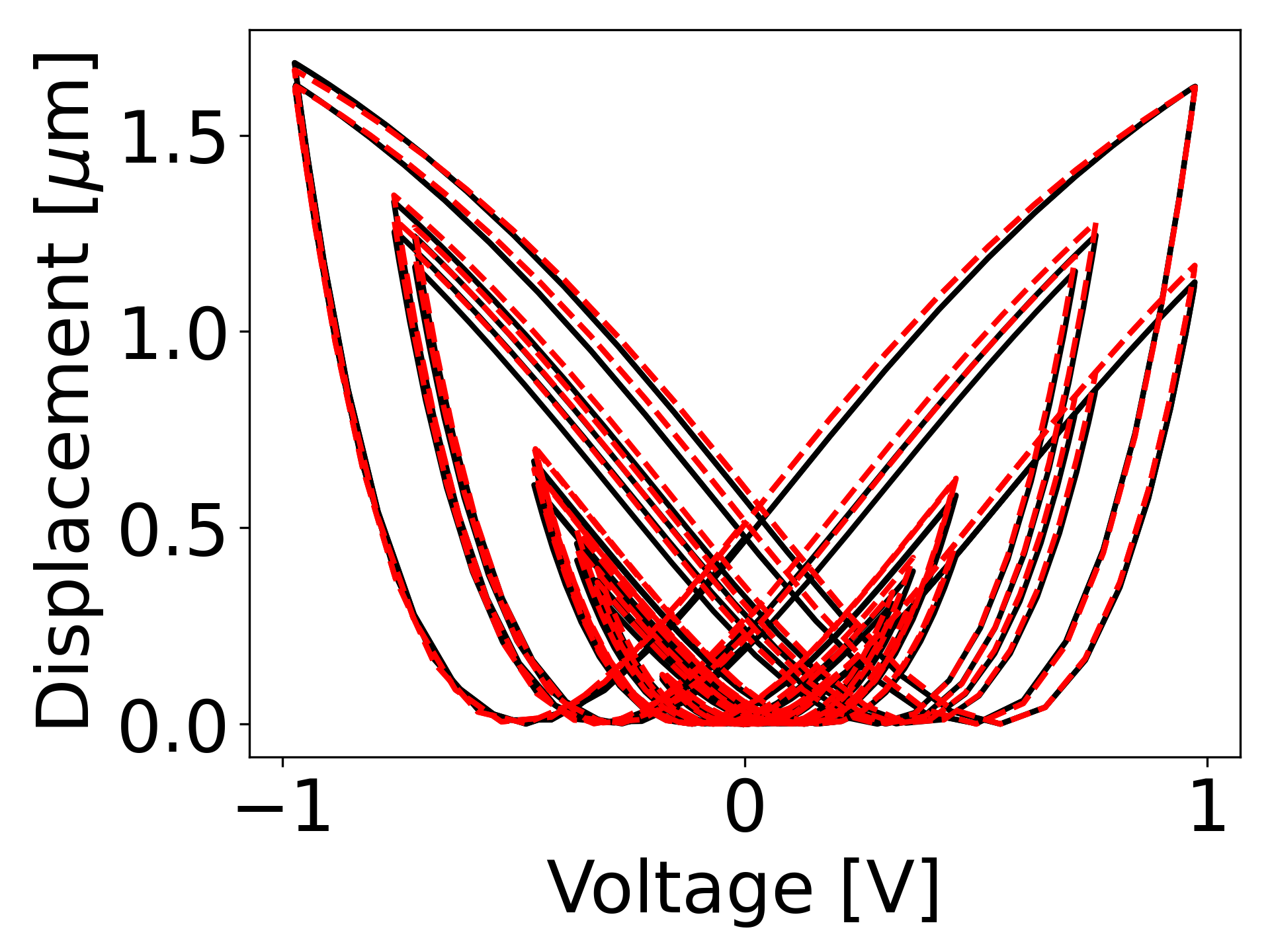}
\includegraphics[width=0.30\columnwidth]{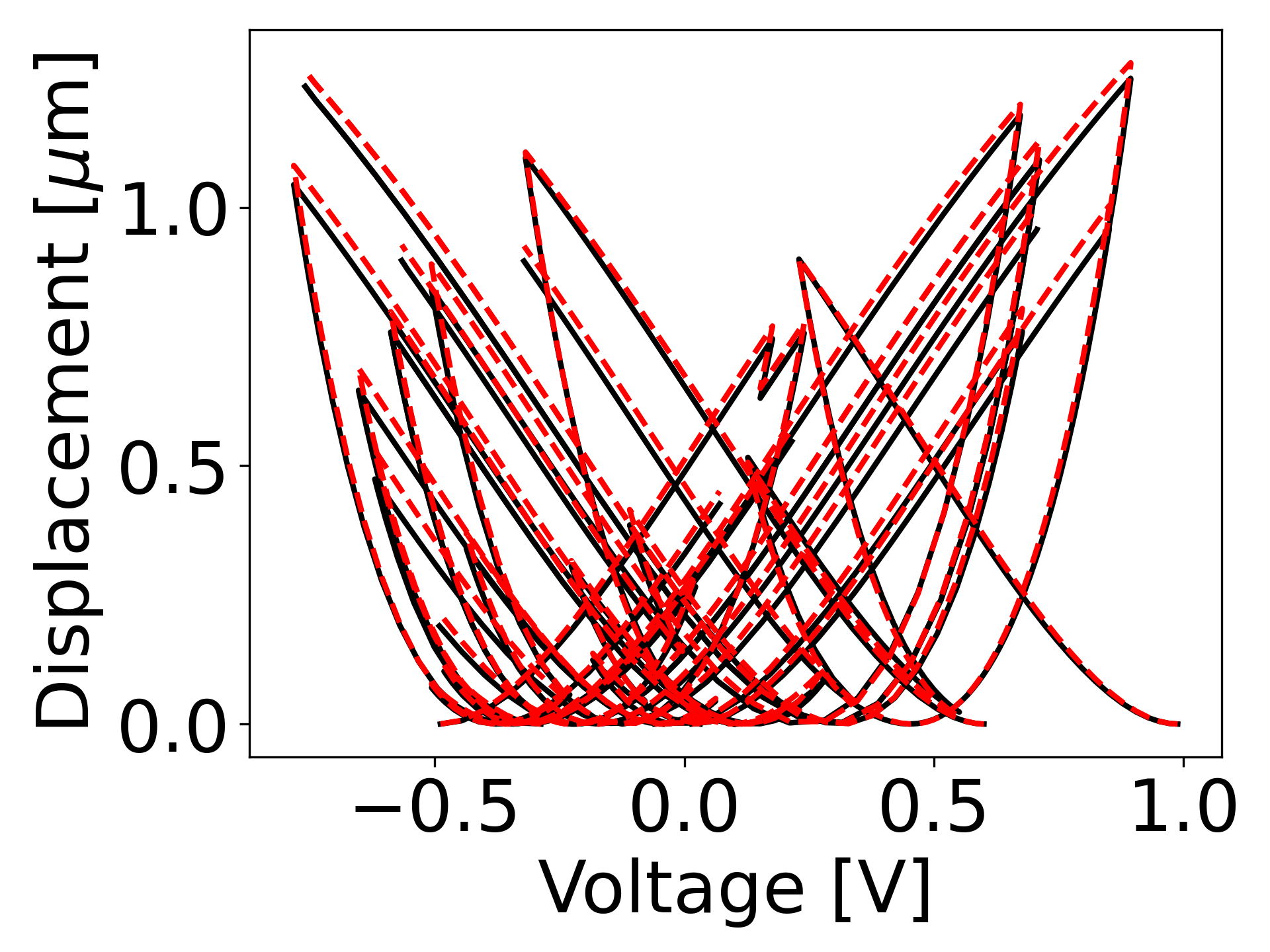}
\includegraphics[width=0.30\columnwidth]{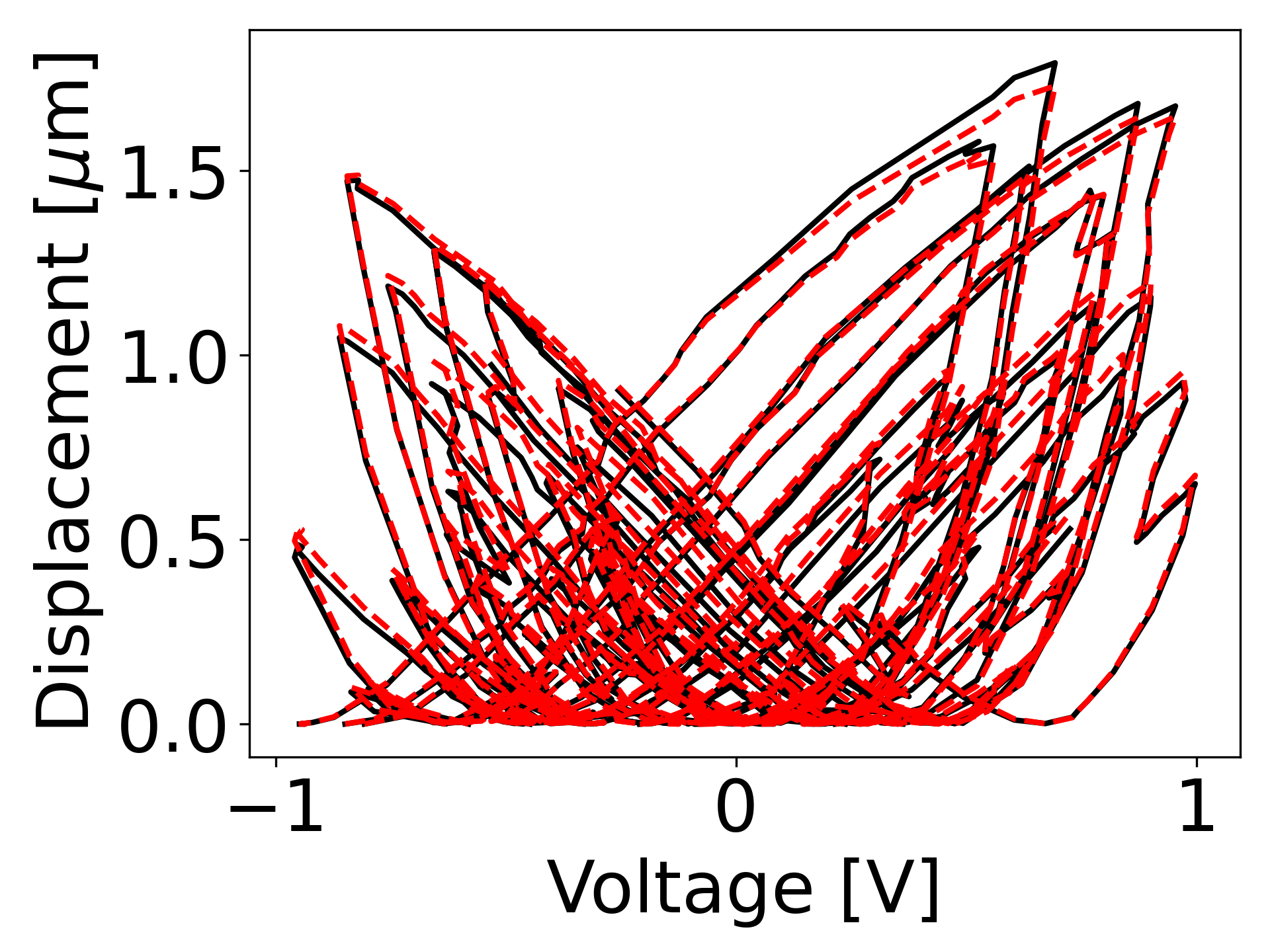}
\caption{Results for Experiment 4: \textbf{Top row: }Testing voltage fields sampled from Sine, RBF, and Matern32 kernels for FNO (first three columns) and NSO (last three columns). \textbf{Bottom row: }Hysteresis responses for FNO (first three columns) and NSO (last three columns) corresponding to the voltage fields on the Top row.}
\label{Nfig4}
\end{figure*}

\subsection{Experiment 3}
\label{exp3}
This experiment assesses the potential of the proposed framework on butterfly-shaped hysteresis. The voltage fields are generated similar to those in previous experiments, with minor modifications. Specifically, the sine fields use a reduced frequency of \(2\pi\) instead of \(4\pi\), and the Matern kernel is changed from Matern52 to Matern32 to increase the complexity of voltage fields.

The dynamics governing the butterfly-shaped hysteresis are taken to be a system of equation \cite{chandra2023discovery}: $\dot{d} = 2 |\dot{v}| v y - 4.70 |\dot{v}| d y + 3 \dot{v} y$; $\dot{y} = |\dot{v}| v - 2.35 |\dot{v}| y + 1.5 \dot{v}$. Here $y$ denotes a latent variable. The model identified by NSO is: $\dot{d} = 1.98 |\dot{v}| v y - 4.67 |\dot{v}| d y + 2.98 \dot{v}$ y; $\dot{y} = 0.99 |\dot{v}| v - 2.32 |\dot{v}| y + 1.49 \dot{v}$, which preserves the original structure of a white-box ODE.

\subsection{Experiment 4}
\label{exp4}
For the fourth experiment, the sine fields are generated with frequency \(4\pi\), and the RBF and Matern32 fields are generated similarly to experiment 3. The experiment explores the butterfly-shaped hysteresis system governed by the relation $\dot{d} = 4 \dot{v} y - 2.5 |\dot{v}| d y - 0.2 \dot{v} |d| y$; $\dot{y} = 2 \dot{v} - 1.25 |\dot{v}| y - 0.1 \dot{v} |y|$, which includes both linear and nonlinear terms. Here $y$ denotes a latent variable. The identified ODE by NSO is given by: $\dot{d} = 3.74 \dot{v} y - 2.4 |\dot{v}| d y - 0.1 \dot{v} |d| y$; $\dot{y} = 1.87 \dot{v} - 1.20 |\dot{v}| y - 0.05 \dot{v} |y|$, which maintains the mathematical form and interpretability of the actual dynamics and is used for simulating the RBF and Matern fields.

Table~\ref{tbl1} presents error metrics for four operators, DON, FNO, CNO, and NSO, across all experiments and kernel types. Training for all models is performed on the Sine field; testing on that field yields good performance generally, and the methods can predict the hysteresis loops from the Sine field with varying errors. However, when models are tested on different fields on which training has not been performed, such as RBF and Matern fields, all the traditional neural operators, DON, FNO, and CNO, fail by an order of magnitude behind the proposed NSO. Hence, NSO is generalizable to varying fields, essentially important for monitoring and maintaining piezoelectric material-based devices and machines.

In Figures~\ref{Nfig1}, ~\ref{Nfig2}, \ref{Nfig3}, and~\ref{Nfig4}, the left three plots show results from FNO and the right three from NSO. The red dashed lines in each plot represent the ground truth, while the black curves show the model predictions. Figures show that FNO and NSO accurately predict for the Sine kernel. FNO predictions deviate substantially from the ground truth for the RBF and Matern kernels. The NSO predictions, on the other hand, remain closely aligned across all kernels. These results show that NSO is accurate in training and generalizes effectively to different fields, which is important for real-world applications involving piezoelectric materials.

\subsection{Experiment 5: Ablation on noise and fidelity}
The rationale for the following two experiments is to assess the potential of NSO on noisy and low-fidelity data. In the following two subsections, NSO is compared with Lasso and SINDy. All methods are employed on the same candidate library of basis functions. Lasso incorporates an L1 penalty in the loss function, while SINDy utilizes STLSQ. Moreover, unlike NSO, which is trained on the denoised and upsampled predictions from FNO, both Lasso and SINDy are applied to the raw data. The experiments assist in assessing the benefits of neural operator pre-processing in enhancing symbolic regression performance.

\subsubsection{Experiment 5a (Noisy data)}
\label{exp5a}
This experiment utilizes the data used for experiment 1 and corrupts the displacement profiles with 20\% Gaussian noise to simulate measurement uncertainty. The actual hysteresis dynamics are governed by: $\dot{d} = 0.4 |\dot{v}| v - 0.85 |\dot{v}| d + 0.2 \dot{v}$.The equations discovered by the three methods are as follows, NSO: $\dot{d} = 0.39 |\dot{v}| v - 0.83 |\dot{v}| d + 0.2 \dot{v}$. Lasso: $\dot{d} = 0.15 |\dot{v}| v - 0.001 |\dot{v}| d + 0.23 \dot{v}$. SINDy: $\dot{d} = 0.28 |\dot{v}| v - 0.35 |\dot{v}| d + 0.22 \dot{v}$. 

Among the three approaches, NSO achieves the most accurate model structure, effectively capturing the dominant physical interactions and coefficients even in the presence of noise. While both Lasso and SINDy identify some relevant terms, their coefficients deviate significantly from the ground truth, and in Lasso's case, the model severely underestimates the second term. This suggests that symbolic regression methods applied directly to noisy data can yield unreliable models, whereas NSO, trained on denoised operator-enhanced predictions, demonstrates strong robustness.

\begin{figure} 
\centering
\includegraphics[width=0.30\columnwidth]{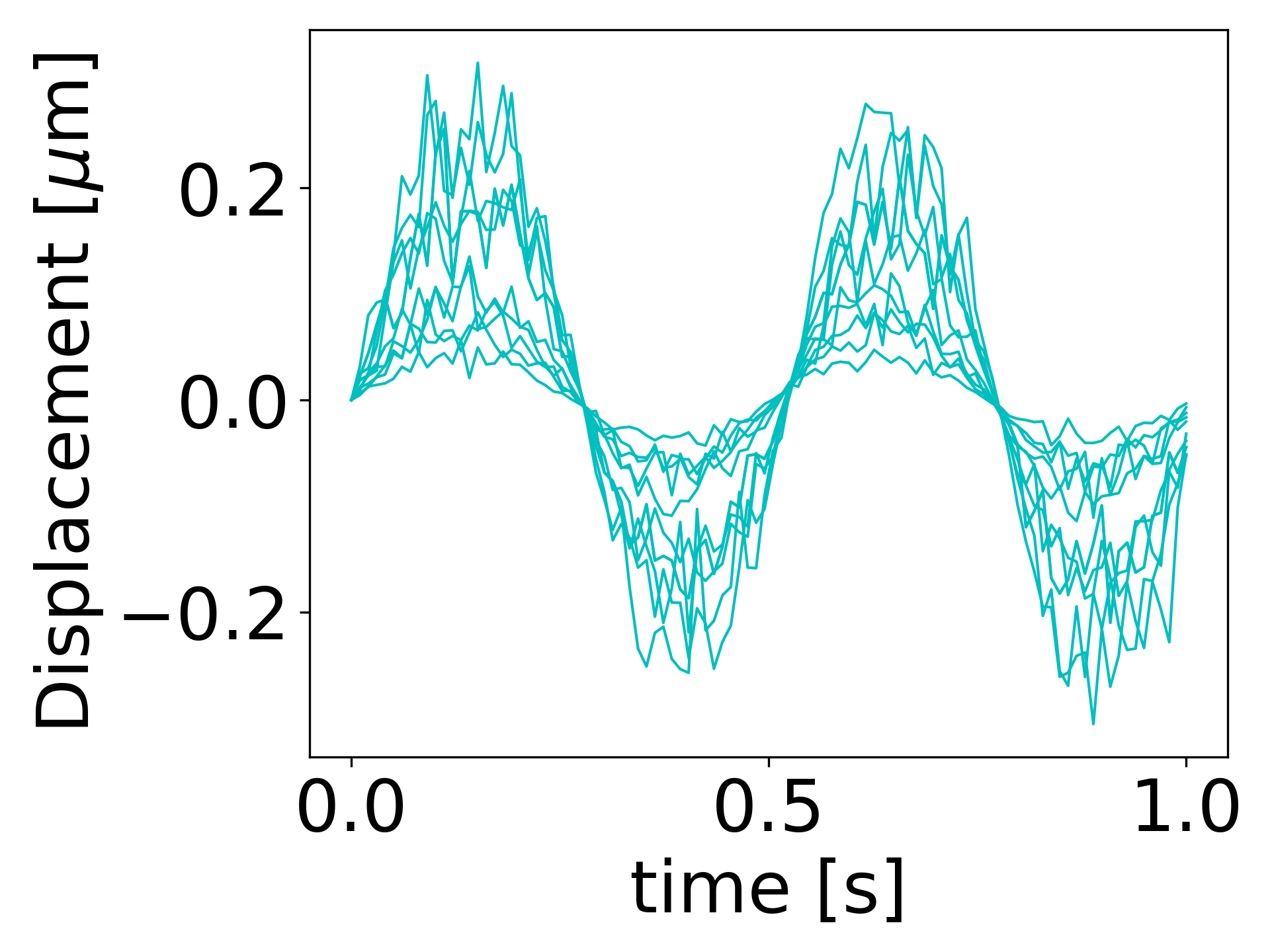}
\includegraphics[width=0.30\columnwidth]{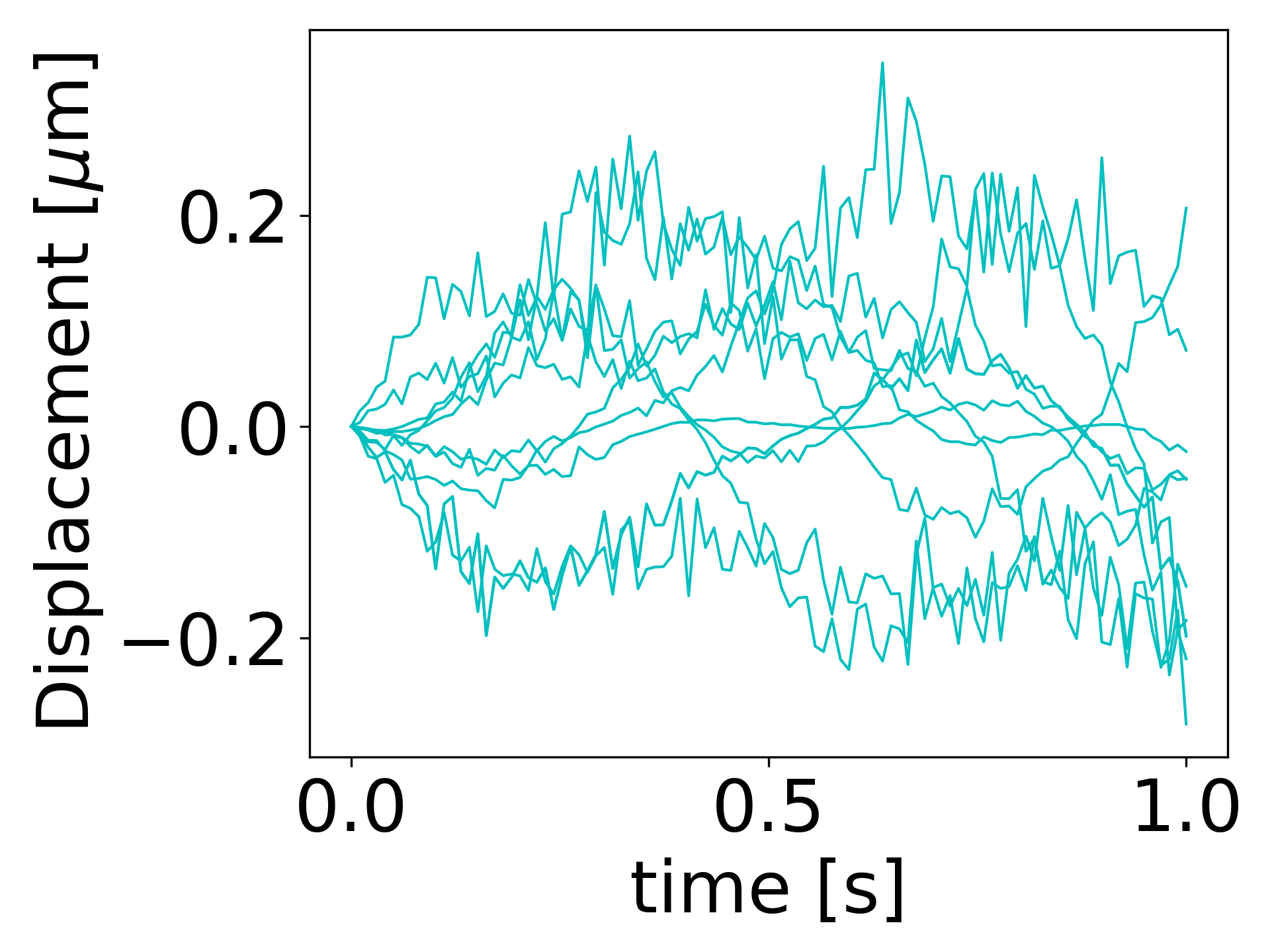}
\includegraphics[width=0.30\columnwidth]{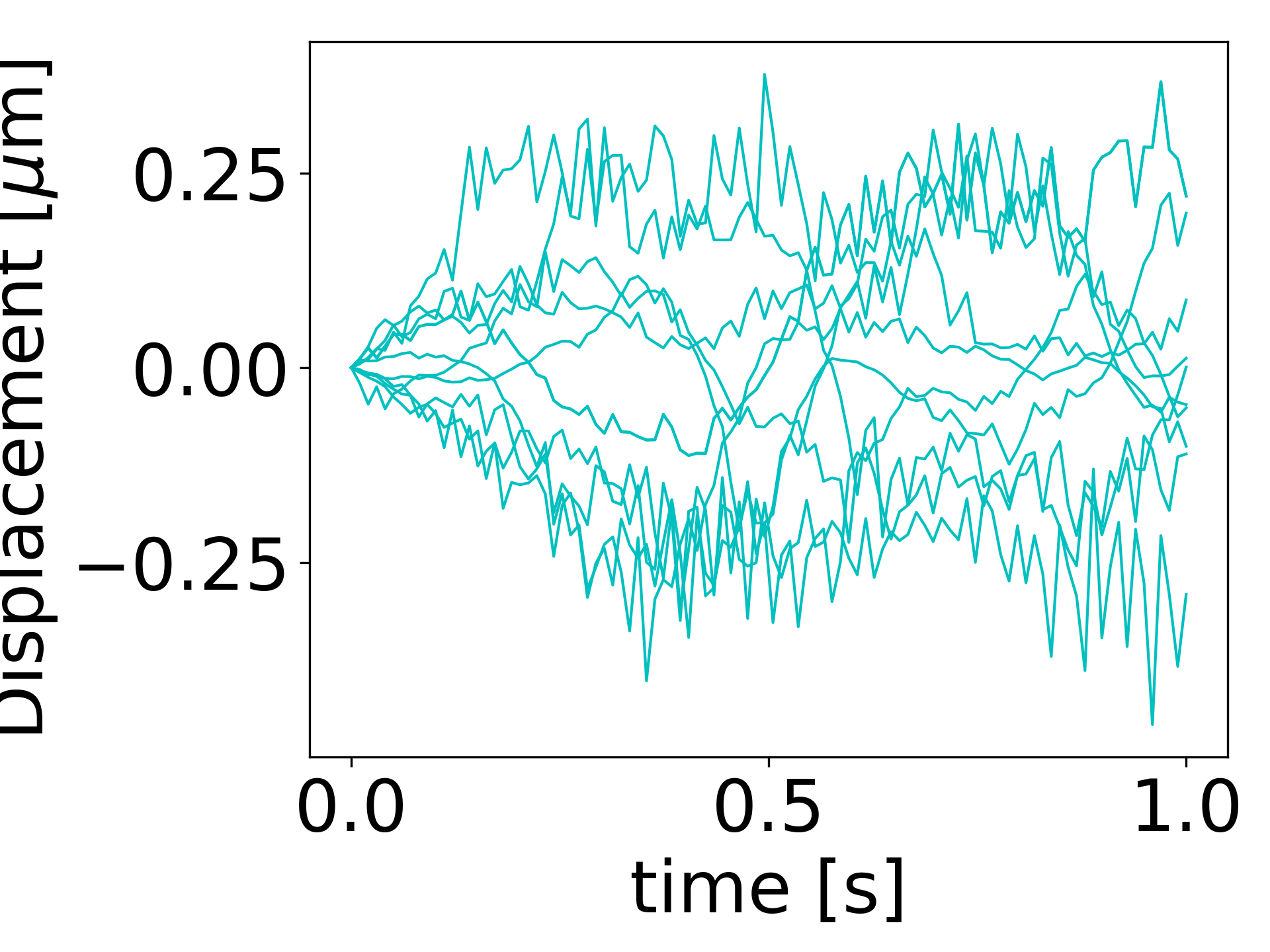}

\includegraphics[width=0.30\columnwidth]{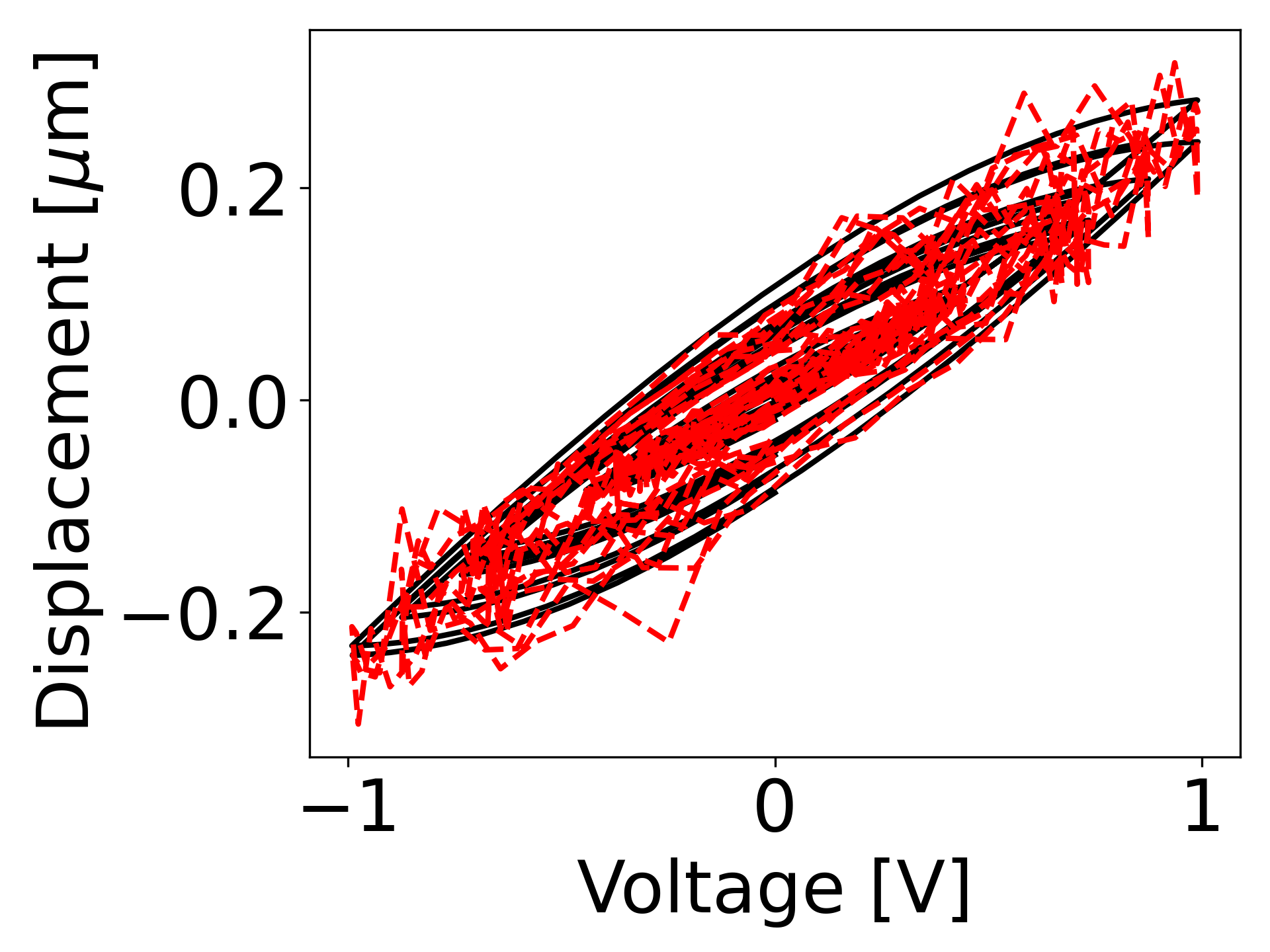}
\includegraphics[width=0.30\columnwidth]{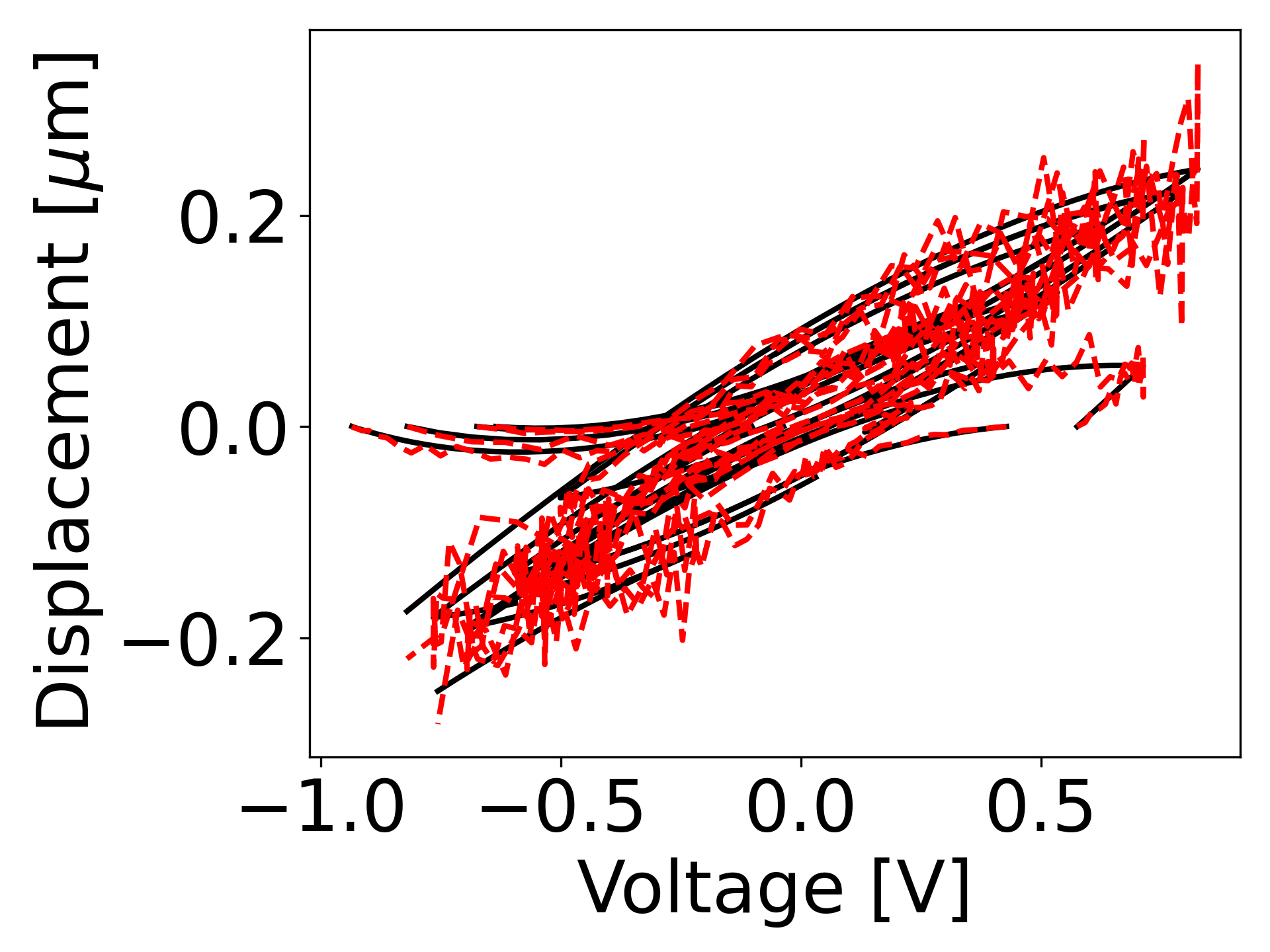}
\includegraphics[width=0.30\columnwidth]{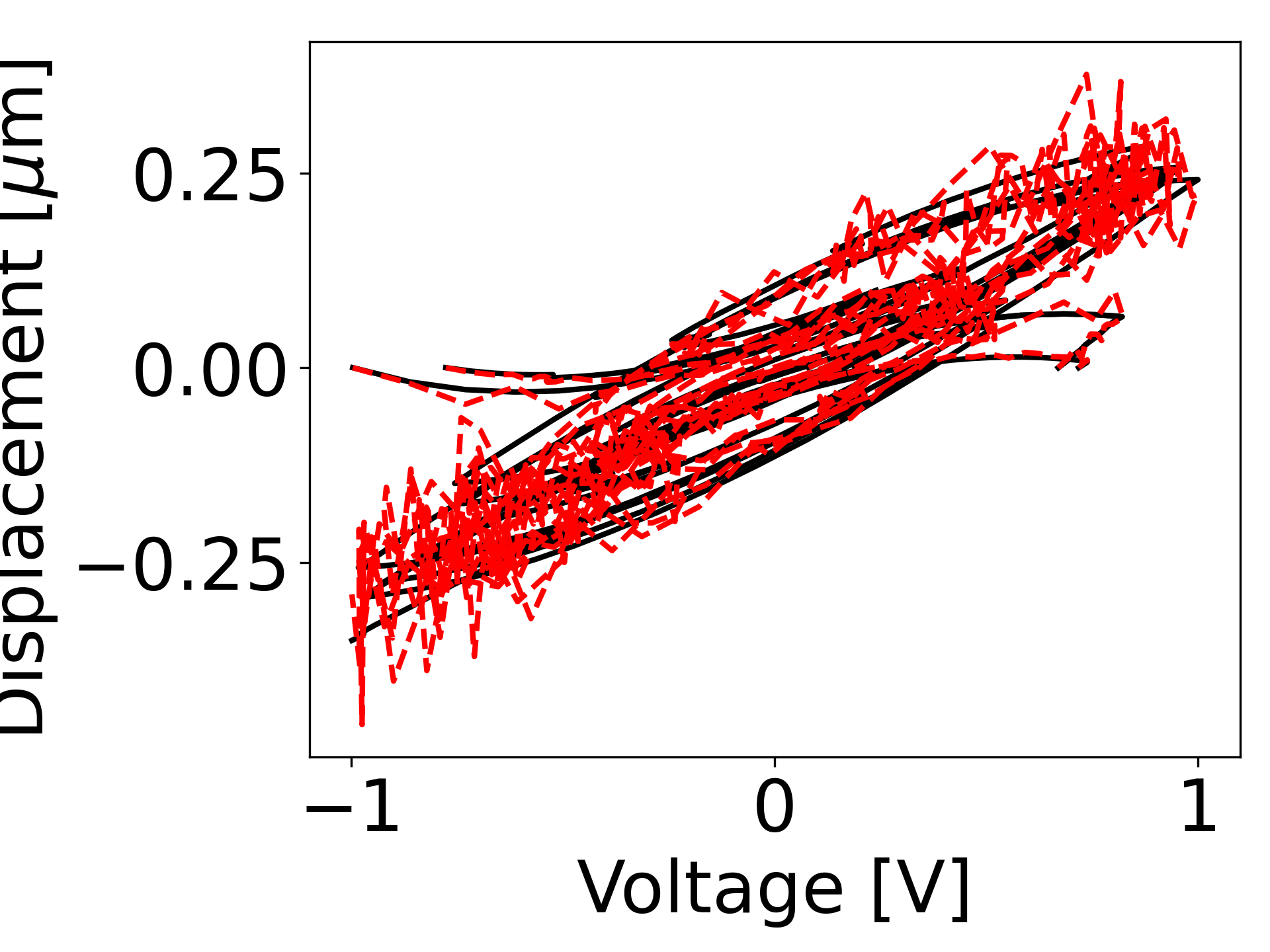}
\caption{\textbf{Top row:} Displacement profiles with 20\% Gaussian noise on Sine, RBF and Matern52 kernels; \textbf{Bottom row:} NSO predictions under corresponding noisy data.}
\label{Nfig5a}
\end{figure}

\begin{figure} 
\centering
\includegraphics[width=0.30\columnwidth]{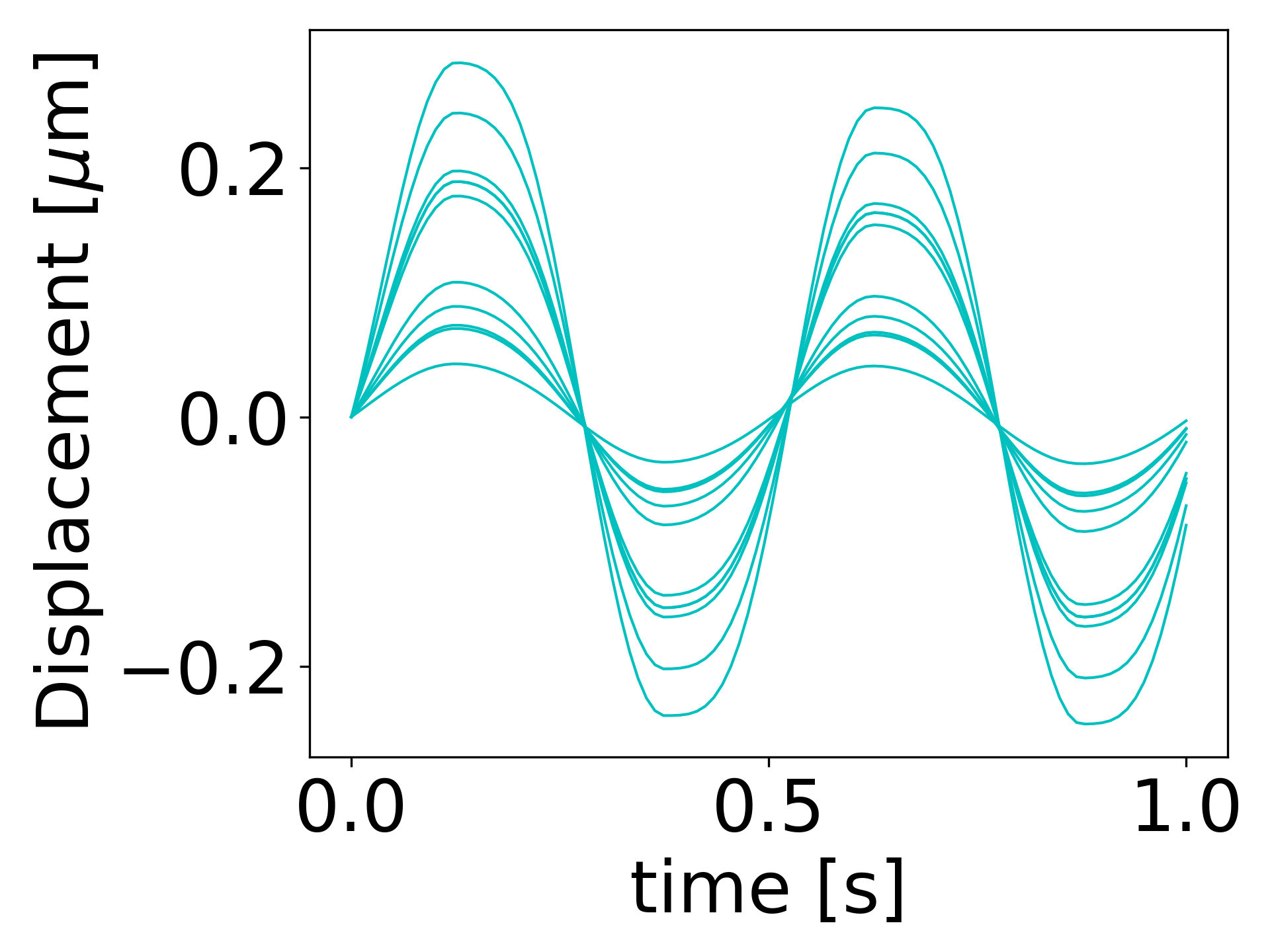}
\includegraphics[width=0.30\columnwidth]{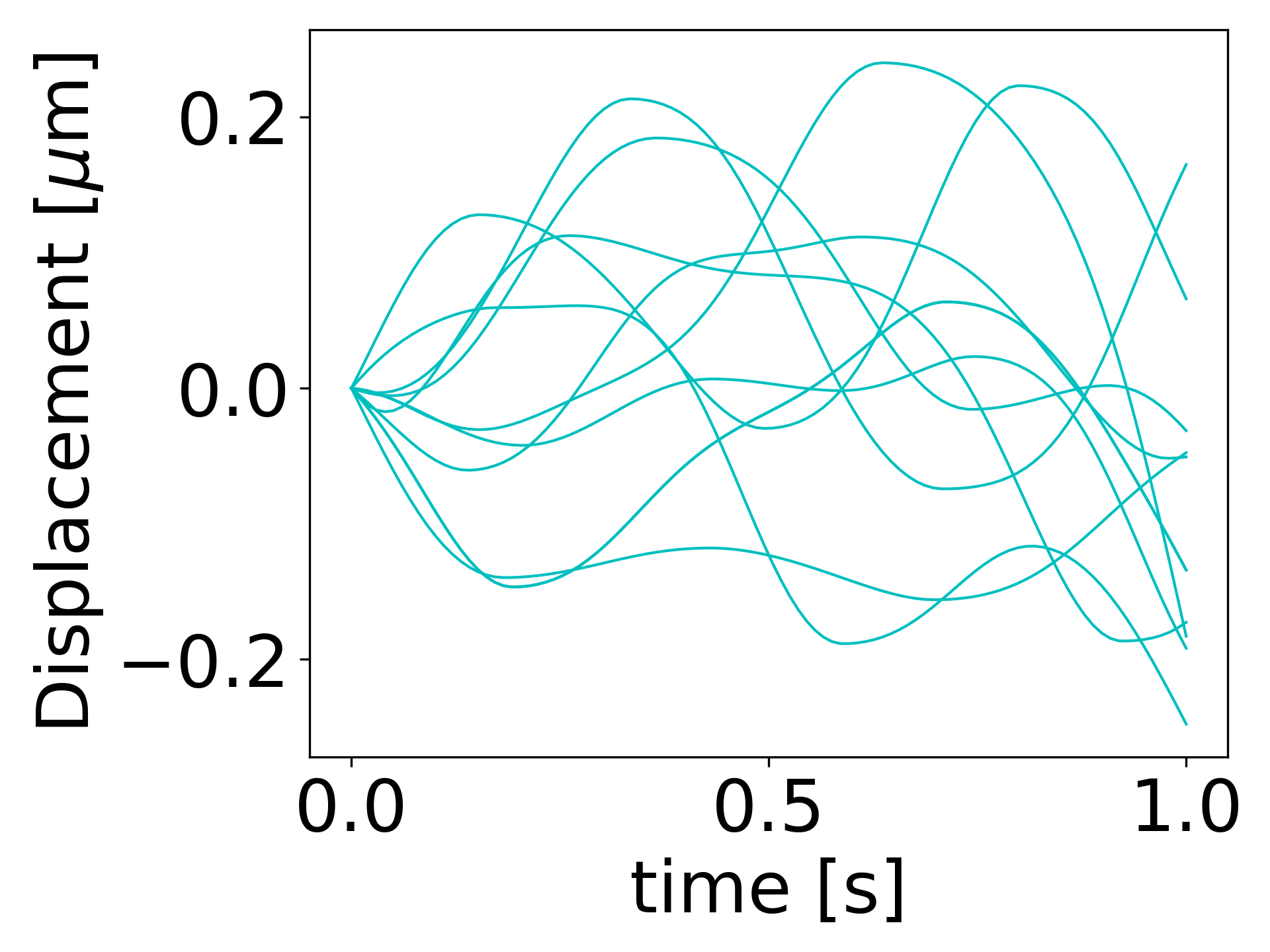}
\includegraphics[width=0.30\columnwidth]{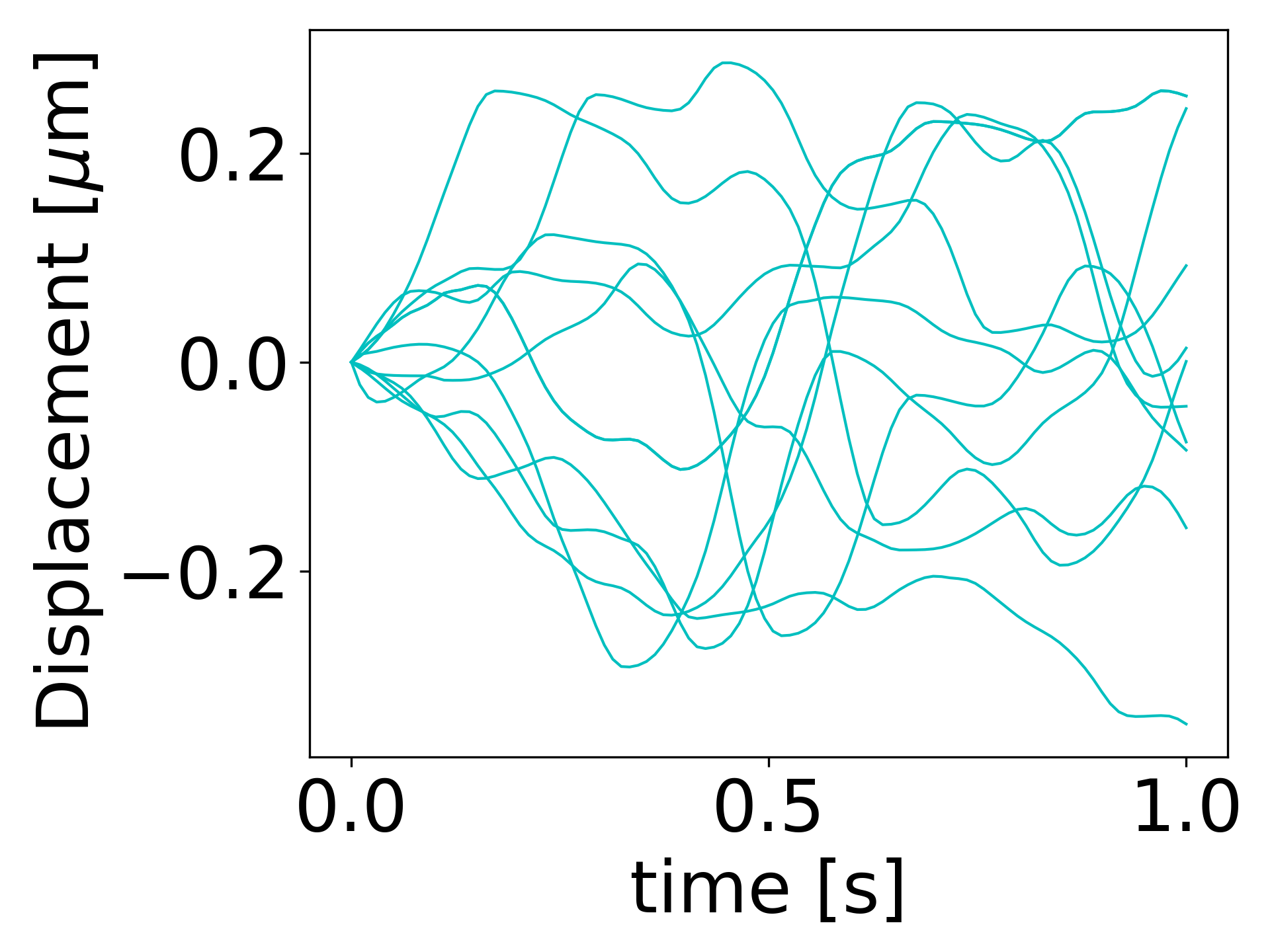}

\includegraphics[width=0.30\columnwidth]{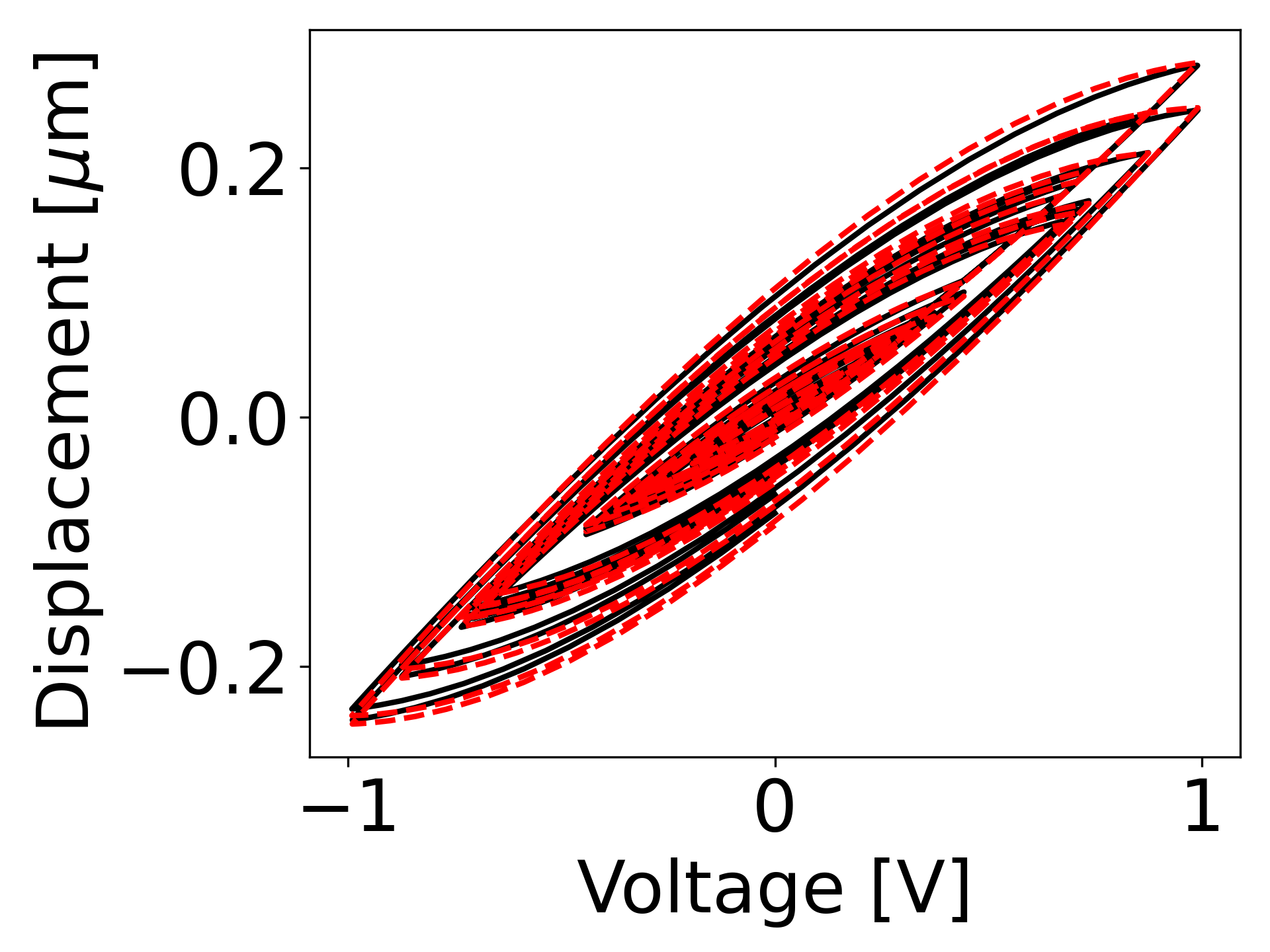}
\includegraphics[width=0.30\columnwidth]{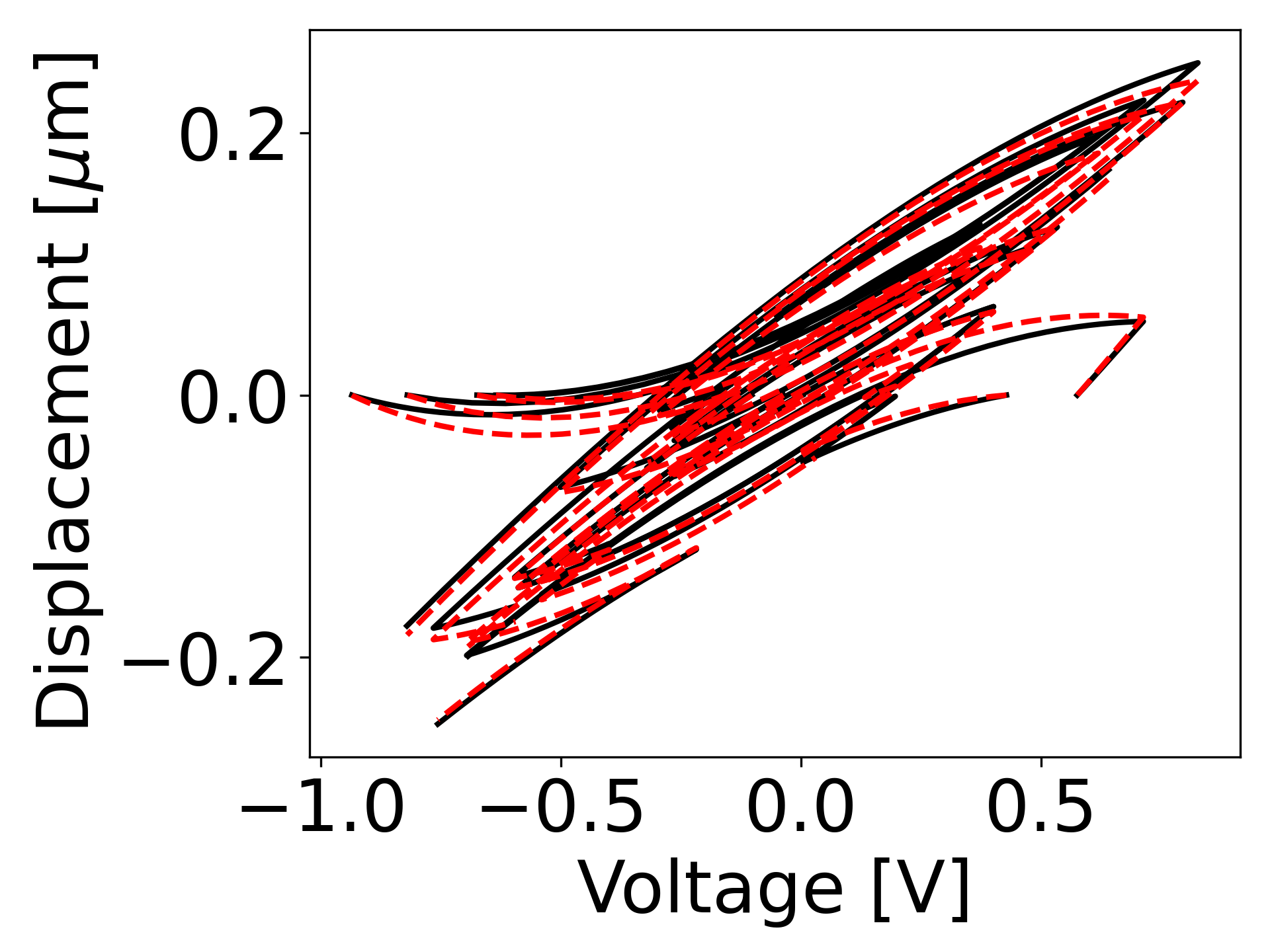}
\includegraphics[width=0.30\columnwidth]{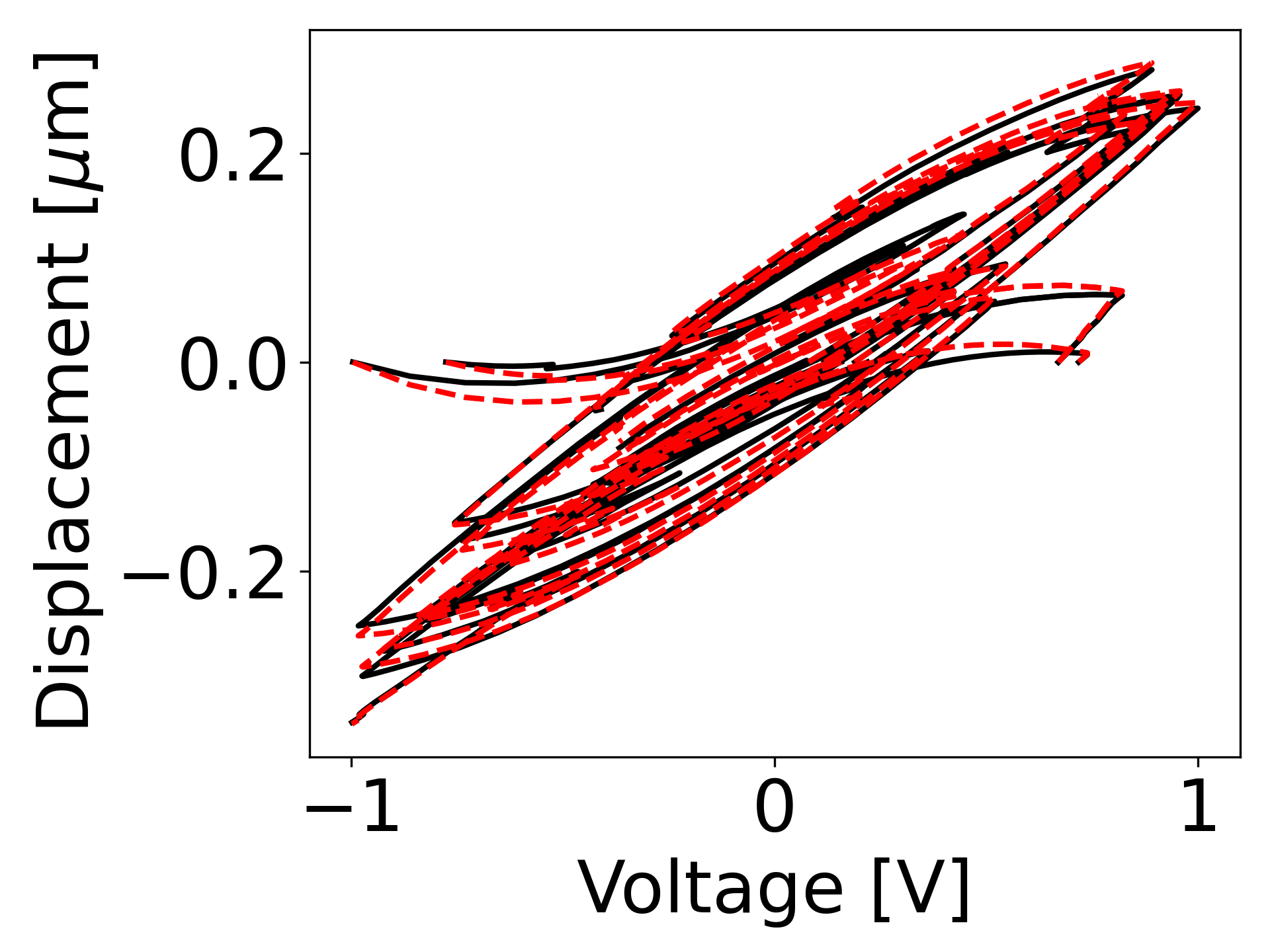}
\caption{\textbf{Top row: }Low-fidelity displacement profiles with 20 points each for Sine, RBF, and Matern52 kernels; \textbf{Bottom row:} NSO predictions under corresponding data.}
\label{Nfig5b}
\end{figure}

\begin{table}[t]
\centering
\setlength{\tabcolsep}{2.8pt}
\caption{Performance of model discovery methods for noisy data (Exp 5(a)), and low-fidelity data (Exp 5(b)), evaluated across diverse kernels under different metrics.}
\label{tbl2}
\resizebox{\columnwidth}{!}{%
\begin{tabular}{llcccccc}
\toprule
\textbf{Kernel} & \textbf{Metric} & \multicolumn{2}{c}{\textbf{LASSO}} & \multicolumn{2}{c}{\textbf{SINDy}} & \multicolumn{2}{c}{\textbf{NSO}} \\
 & & 5(a) & 5(b) & 5(a) & 5(b) & 5(a) & 5(b) \\
\midrule
\multirow{3}{*}{Sine}
& $\mathcal{R}$ & 2.29e-01 & 2.33e-01 & 2.19e-01 & 5.71e-02 & \textbf{1.97e-01} & \textbf{3.42e-02} \\
& RMSE          & 2.41e-02 & 2.41e-02 & 2.31e-02 & 5.90e-03 & \textbf{2.07e-02} & \textbf{3.53e-03} \\
& MAE           & 1.52e-02 & 1.43e-02 & 1.46e-02 & 3.65e-03 & \textbf{1.26e-02} & \textbf{2.52e-03} \\
\midrule
\multirow{3}{*}{RBF}
& $\mathcal{R}$ & 6.31e-01 & 5.79e-01 & 3.10e-01 & 1.36e-01 & \textbf{1.99e-01} & \textbf{7.22e-02} \\
& RMSE          & 7.94e-02 & 7.14e-02 & 3.90e-02 & 1.68e-02 & \textbf{2.50e-02} & \textbf{8.90e-03} \\
& MAE           & 6.15e-02 & 5.59e-02 & 2.96e-02 & 1.37e-02 & \textbf{1.62e-02} & \textbf{7.00e-03} \\
\midrule
\multirow{3}{*}{Matern52}
& $\mathcal{R}$ & 6.74e-01 & 6.19e-01 & 3.12e-01 & 1.26e-01 & \textbf{1.98e-01} & \textbf{6.43e-02} \\
& RMSE          & 8.75e-02 & 7.88e-02 & 4.05e-02 & 1.61e-02 & \textbf{2.58e-02} & \textbf{8.18e-03} \\
& MAE           & 6.87e-02 & 6.30e-02 & 3.10e-02 & 1.32e-02 & \textbf{1.69e-02} & \textbf{6.40e-03} \\
\bottomrule
\end{tabular}}
\end{table}

\subsubsection{Experiment 5b (Low-fidelity data)}
\label{exp5b}
This experiment utilizes the data used for experiment 1 and downsamples it by a factor of five to reflect sparse sensor measurements. The actual hysteresis dynamics are governed by: $\dot{d} = 0.4 |\dot{v}| v - 0.85 |\dot{v}| d + 0.2 \dot{v}$. The equations discovered by the three methods are as follows, NSO: $\dot{d} = 0.39 |\dot{v}| v - 0.83 |\dot{v}| d + 0.2 \dot{v}$. Lasso: $\dot{d} = 0.13 |\dot{v}| v - 0.001 |\dot{v}| d + 0.22 \dot{v}$. SINDy: $\dot{d} = 0.29 |\dot{v}| v - 0.52 |\dot{v}| d + 0.21 \dot{v}$.

Similar to the noisy case, NSO extracts a parsimonious and interpretable model with accurate coefficients despite reduced data fidelity. SINDy maintains moderate performance, capturing relevant terms but with less precise weights. Lasso again tends to produce error and retains terms with insignificant influence, reducing interpretability. The results emphasize the benefit of using learned neural operator representations as pre-processing tools. By leveraging the expressivity of FNO to reconstruct high-fidelity dynamics from low-fidelity training data, NSO facilitates effective symbolic discovery of governing equations.

Figures~\ref{Nfig5a} and~\ref{Nfig5b} illustrate the predictive performance of the NSO under noise and low-fidelity data. Specifically, Figure \ref{Nfig5a} predicts displacement-voltage responses under 20$\%$ Gaussian noise on displacement profiles, demonstrating NSO's robustness across Sine, RBF, and Matern52 kernels. Figure \ref{Nfig5b} shows the displacement-voltage responses in low-fidelity settings, validating the generalization of NSO. Table~\ref{tbl2} provides quantitative results for Experiments 5(a) and 5(b), showing that NSO consistently achieves the lowest errors across all metrics and kernels.

\subsection{Experiment 6: Ablation on sparsity threshold}
\label{exp6}
The experiment aims to assess the impact of sparsity threshold $\lambda$ on the training of NSO. This experiment investigates the effect of varying the \( \lambda \in \{0.1, 0.01, 0.001\} \) on the performance of the NSO. The data used for this ablation study is from experiment 4. The ODE discovered for $\lambda=0.1$ is $\dot{d} = -2.46 |\dot{v}| d y + 3.76 \dot{v} y$; $\dot{y} = -1.23 |\dot{v}| y + 1.88 \dot{v}$. This model is highly sparse, with only two terms in each equation. It is concise and interpretable. However, its simplicity may omit finer interactions, potentially limiting predictive accuracy. Whereas, the ODE discovered for $\lambda=0.01$ is $\dot{d} = 4 \dot{v} y - 2.5 |\dot{v}| d y - 0.2 \dot{v} |d| y$; $\dot{y} = 2 \dot{v} - 1.25 |\dot{v}| y - 0.1 \dot{v} |y|$. With a slightly relaxed sparsity constraint, more terms are retained. This model captures additional nonlinear interactions, improving accuracy while maintaining interpretability. The ODE discovered for $\lambda\!=\!0.001$ contains nine terms per equation and is not mentioned here for brevity but is not interpretable owing to the large number of terms. 

\begin{figure} 
\centering
\includegraphics[width=0.30\columnwidth]{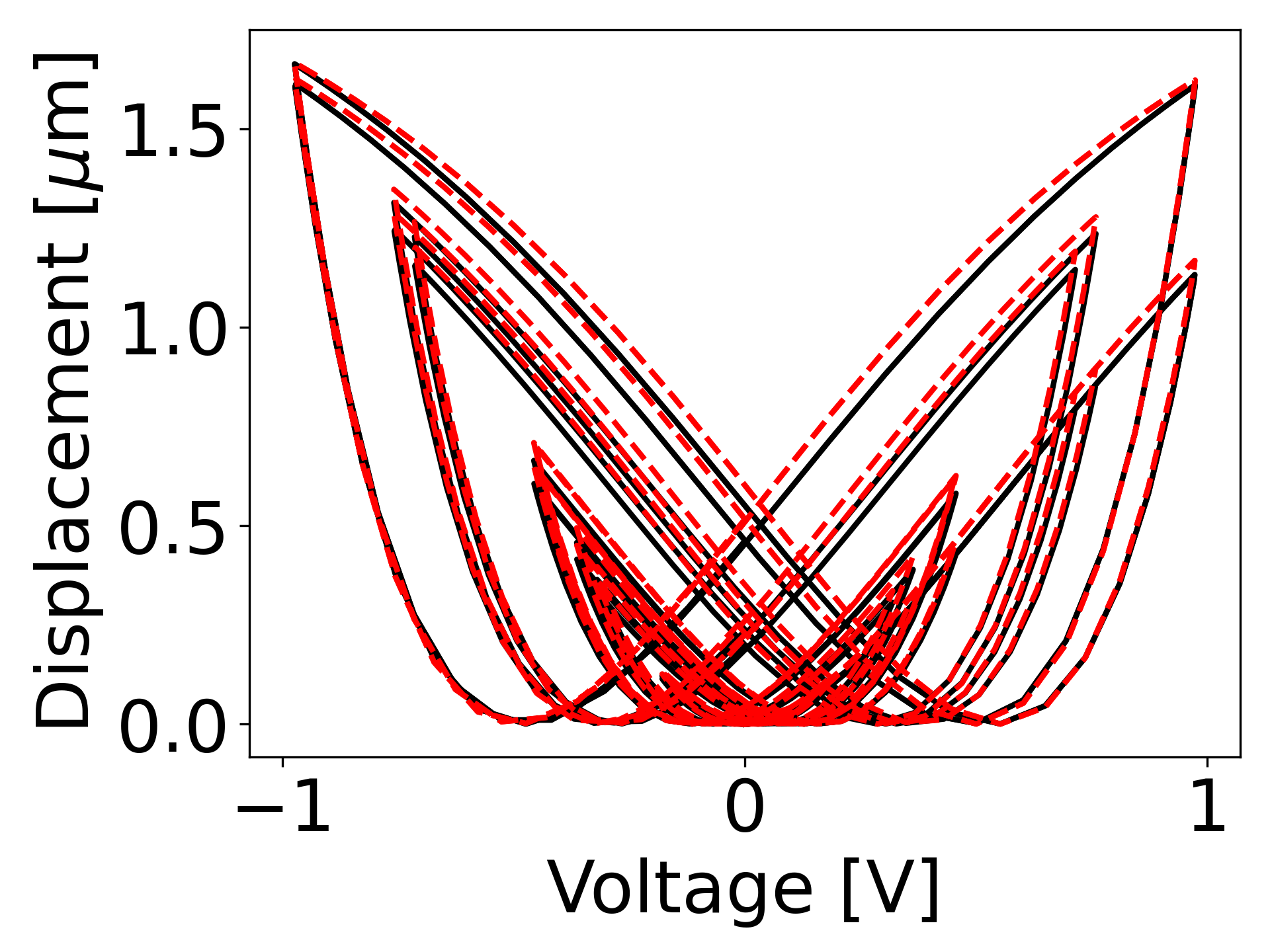}
\includegraphics[width=0.30\columnwidth]{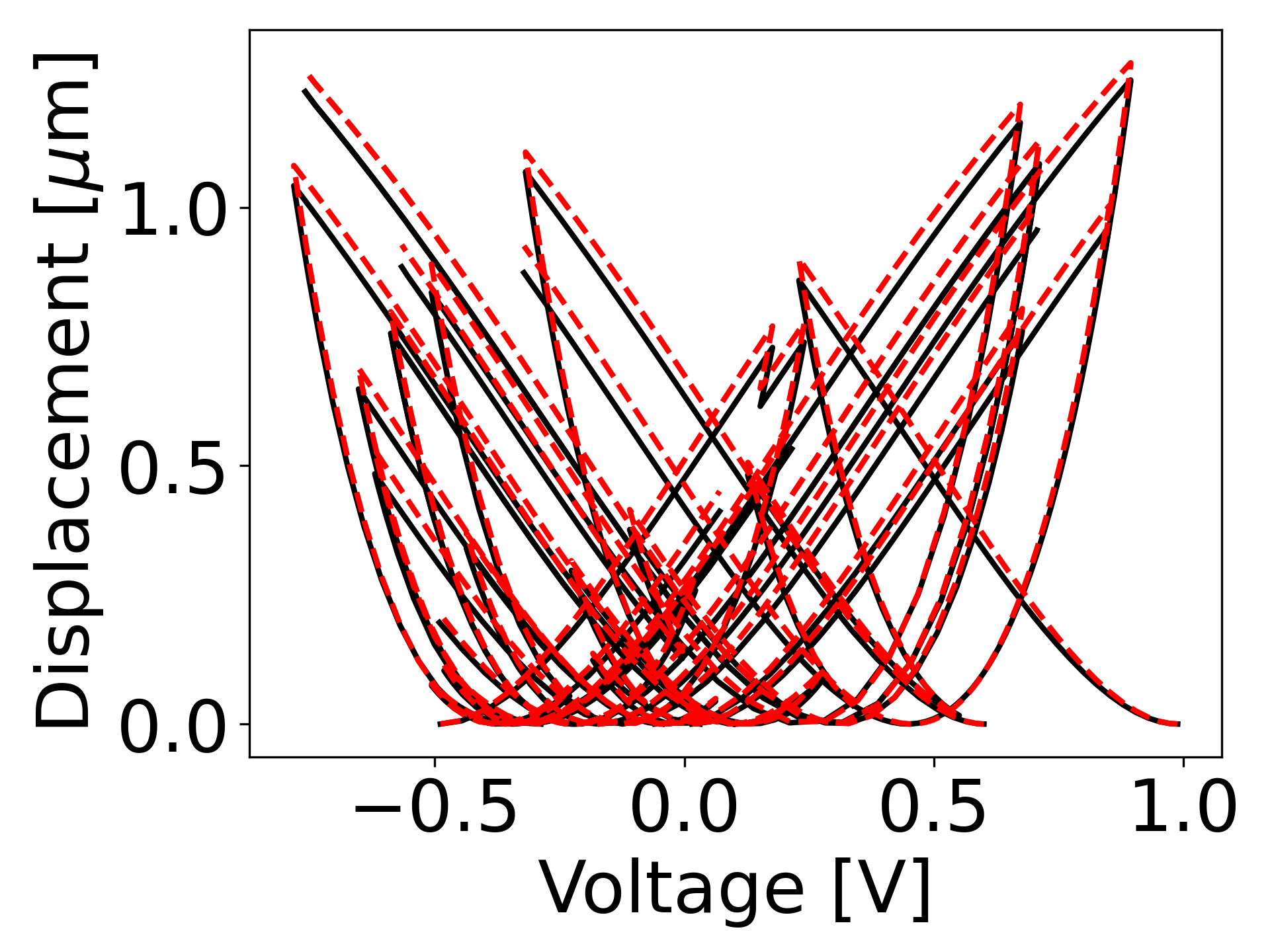}
\includegraphics[width=0.30\columnwidth]{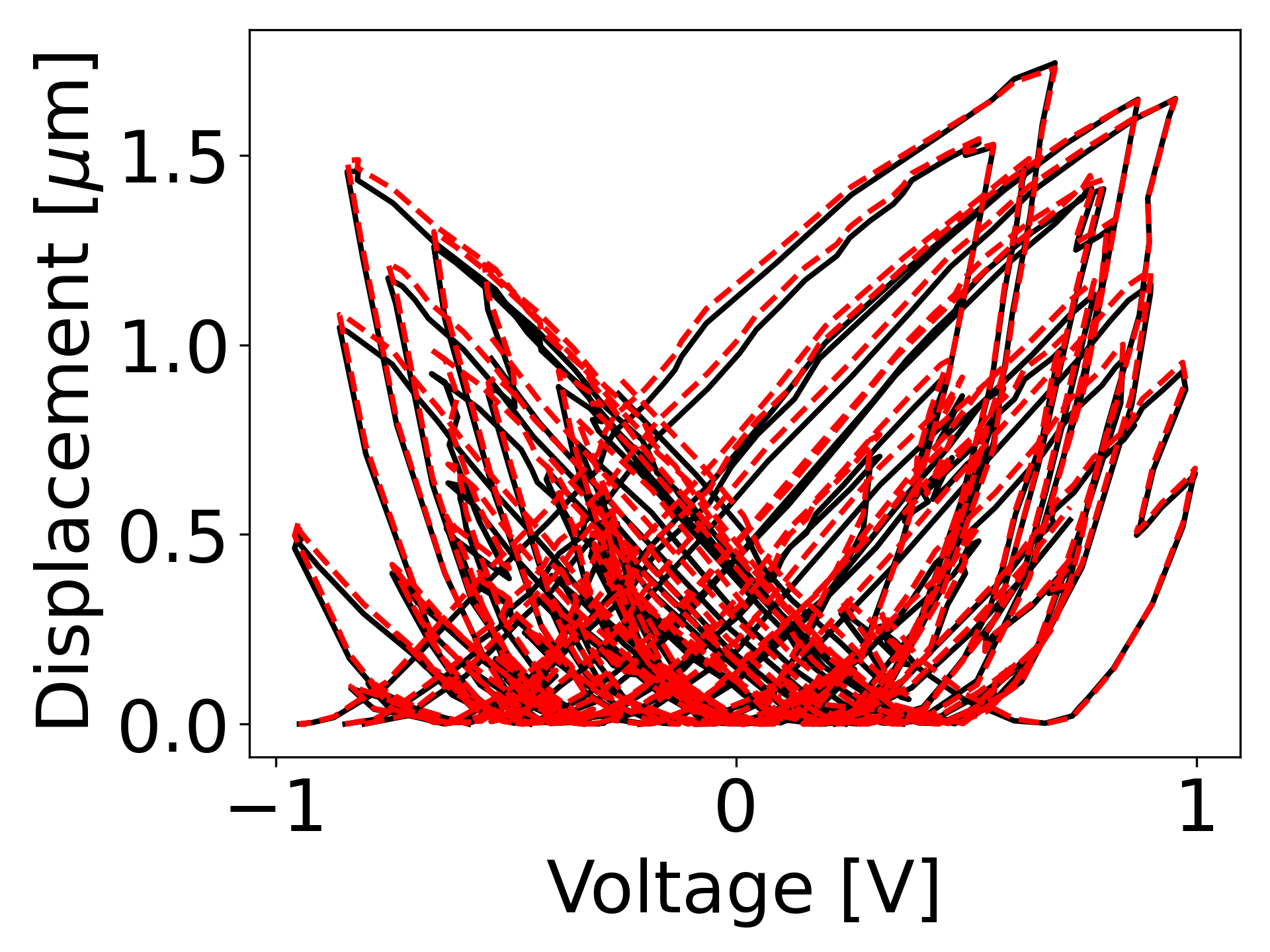}
\includegraphics[width=0.30\columnwidth]{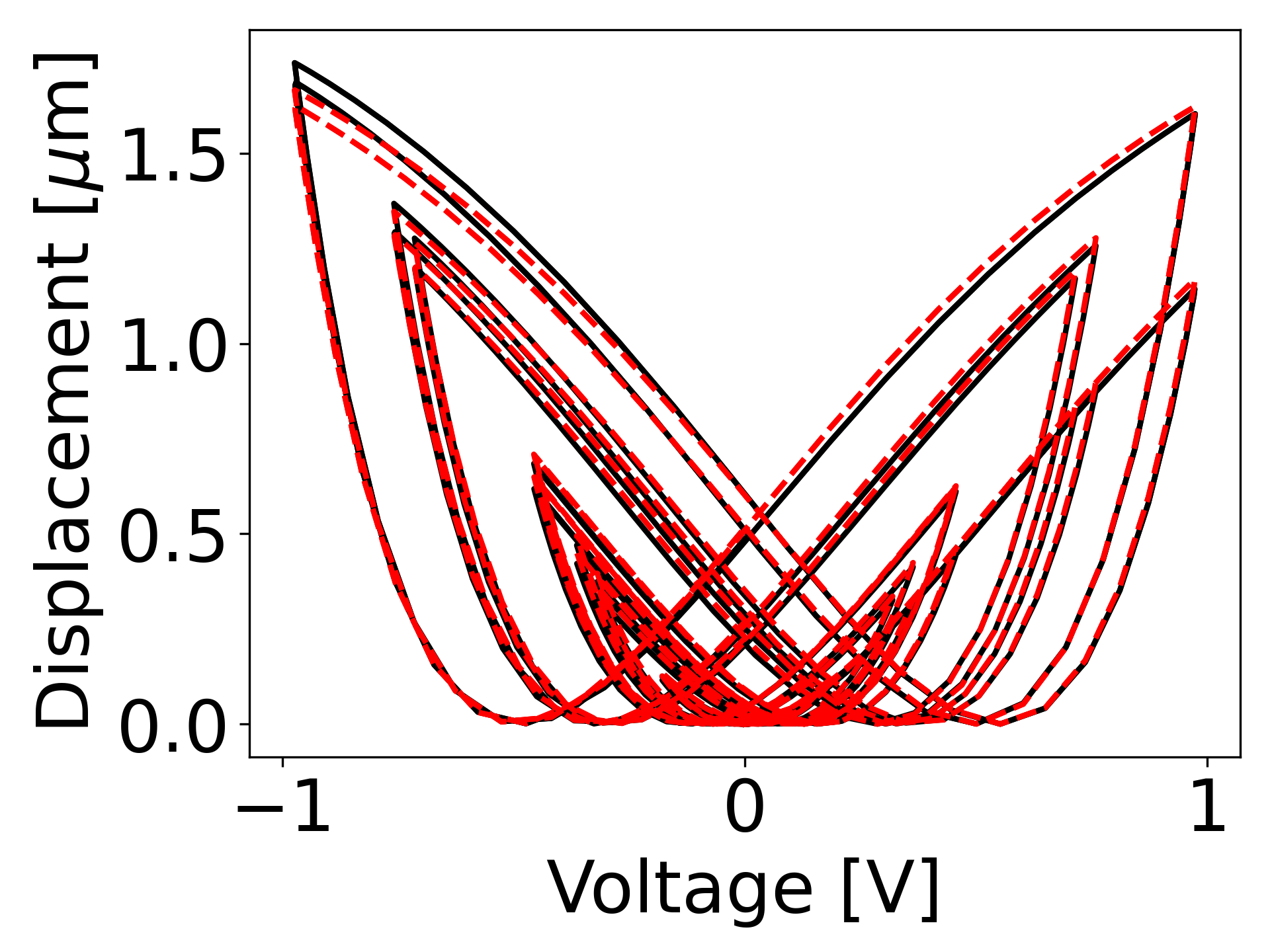}
\includegraphics[width=0.30\columnwidth]{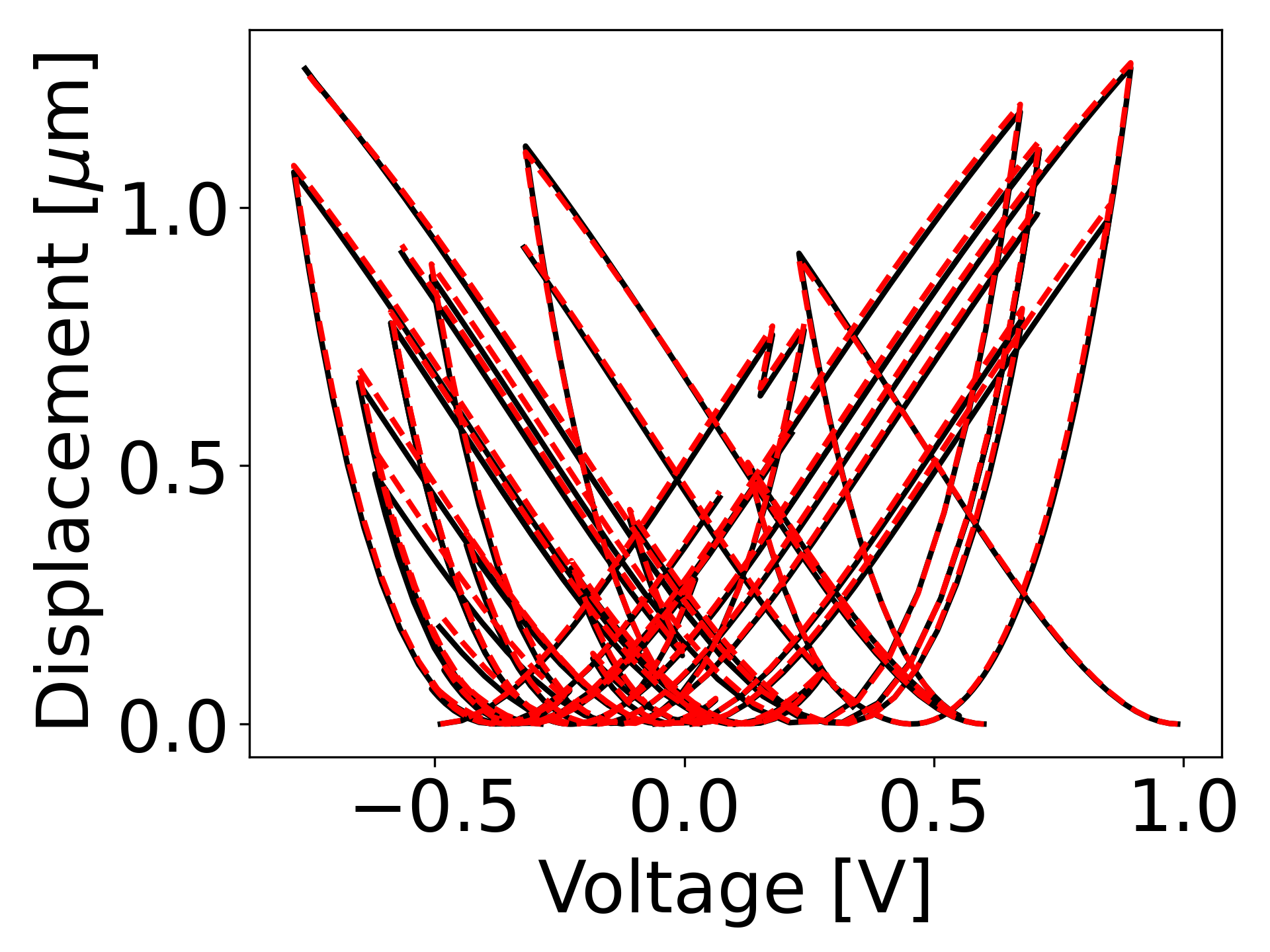}
\includegraphics[width=0.30\columnwidth]{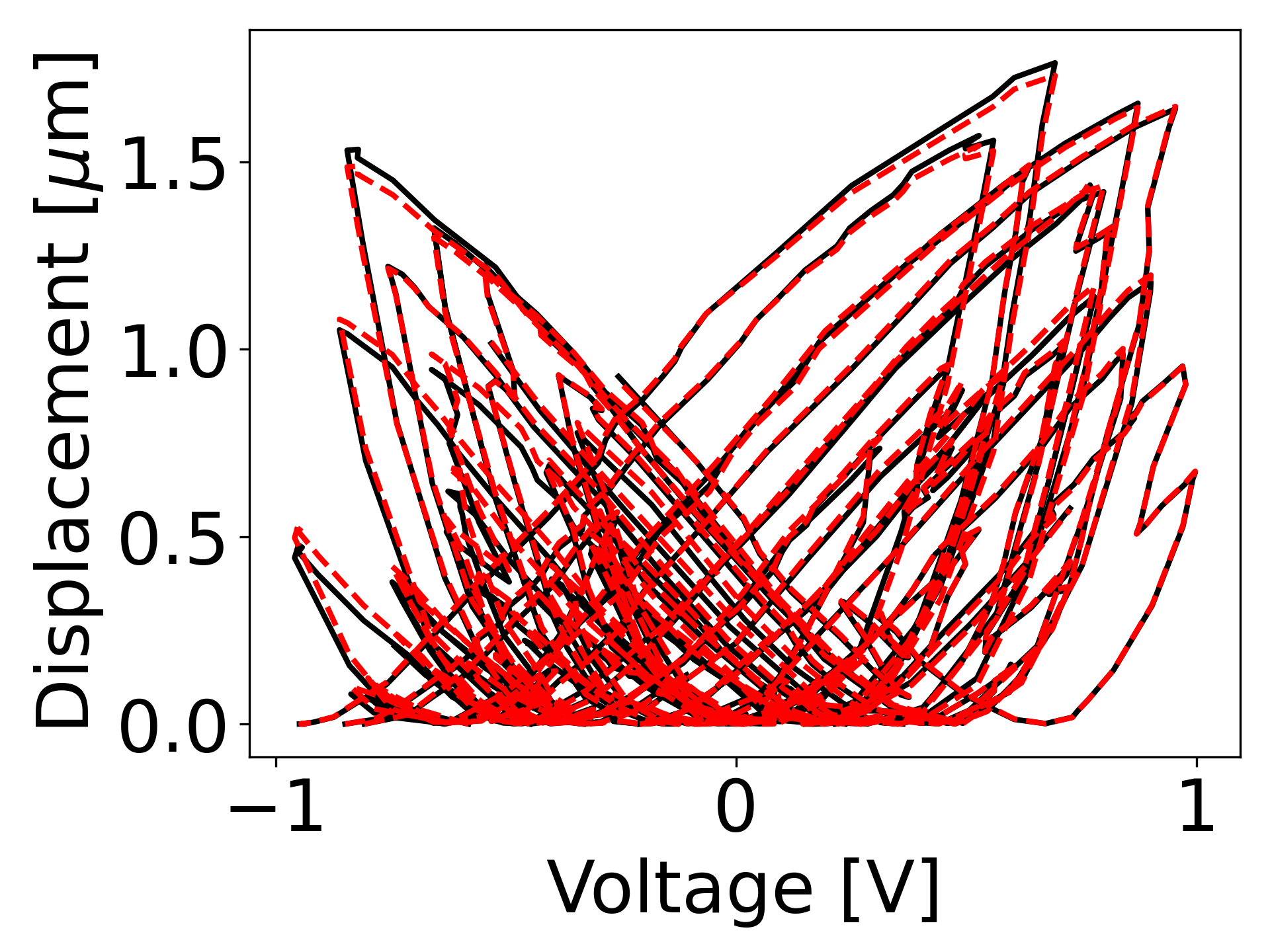}
\caption{NSO predictions with \textbf{Top: } $\lambda=0.1$; \textbf{Bottom: } $\lambda=0.001$, on Sine, RBF, and Matern32 kernels.}
\label{Nfig6}
\end{figure}

Figure~\ref{Nfig6} presents the results of this experiment. Trained on a Sine field with two different threshold values ($\lambda = 0.1$ for the top row and $\lambda = 0.001$ for the bottom row), the model accurately predicts displacement-voltage responses when tested on Sine, RBF, and Matern52 kernels. Table~\ref{tbl3} quantifies NSO predictions for different $\lambda$ values. For example, with $\lambda = 0.001$, NSO achieves L2 errors of $0.0272$, $0.0258$, and $0.0348$ for the Sine, RBF, and Matern52 kernels, respectively. Corresponding RMSE values remain below $0.0229$, and MAE values stay within $0.0107-0.0126$, highlighting that the lower values of $\lambda$ improve the predictive performance of NNO and generalization across different kernels. However, this improved accuracy has a trade-off with interpretability as observed from the discovered equations.

The experiment presents the importance of selecting an appropriate \( \lambda \) to balance accuracy and interpretability. While smaller \( \lambda \) values yield accurate models, they risk overfitting and prioritize less to the underlying physics. Conversely, larger \( \lambda \) values produce more parsimonious models, albeit with reduced expressiveness. The results suggest that intermediate values of \( \lambda \), such as \( 0.01 \), offer the best trade-off, retaining meaningful dynamics while maintaining interpretability and generalizability.

\begin{table}[t]
\centering
\renewcommand{\arraystretch}{1.1}
\setlength{\tabcolsep}{5pt}
\caption{Performance of NSO for varying $\lambda$.}
\label{tbl3}
\begin{tabular}{llccc}
\toprule
Kernel & Metric & $\lambda{=}0.1$ & $\lambda{=}0.01$ & $\lambda{=}0.001$ \\
\midrule
\multirow{3}{*}{Sine} 
& $\mathcal{R}$ & 4.62e-2 & 3.91e-2 & \textbf{2.72e-2} \\
& RMSE          & 2.61e-2 & 2.21e-2 & \textbf{1.54e-2} \\
& MAE           & 1.92e-2 & 1.63e-2 & \textbf{1.07e-2} \\
\midrule
\multirow{3}{*}{RBF} 
& $\mathcal{R}$ & 4.18e-2 & 3.69e-2 & \textbf{2.58e-2} \\
& RMSE          & 2.60e-2 & 2.29e-2 & \textbf{1.60e-2} \\
& MAE           & 2.03e-2 & 1.70e-2 & \textbf{1.11e-2} \\
\midrule
\multirow{3}{*}{Matern52} 
& $\mathcal{R}$ & 4.67e-2 & 3.98e-2 & \textbf{3.48e-2} \\
& RMSE          & 2.49e-2 & 2.12e-2 & \textbf{1.85e-2} \\
& MAE           & 1.90e-2 & 1.56e-2 & \textbf{1.26e-2} \\
\bottomrule
\end{tabular}
\end{table}

\section{Discussion}
\label{sec6}
This section presents a brief discussion of the results and their implications. Across all experiments, NSO consistently achieves lower $\mathcal{R}$, RMSE, and MAE than baseline operators when testing in untrained spaces, as shown in Table~\ref{tbl1}. For instance, in Experiment 1 for the RBF kernel, NSO achieves an RMSE of 1.25e-3 and MAE of 9.97e-4, outperforming DON (8.63e-2, 3.42e-1), FNO (1.11e-1, 9.00e-2), and CNO (5.72e-2, 4.41e-2). This trend holds across other experiments as well, such as Experiment 4 with the Matern32 kernel, where NSO yields an RMSE of 2.12e-2 and MAE of 1.56e-2, significantly lower than DON (3.11e-1, 4.45e-1), FNO (3.90e-1, 2.81e-1), and CNO (4.20e-1, 3.15e-1).

\begin{table}[t]
    \centering
    \caption{Operators' comparison on various attributes}
    \label{tbl4}
    \resizebox{\columnwidth}{!}{%
    \begin{tabular}{|l|c|c|c|c|}
        \hline
        \multirow{2}{*}{Attributes} & \multicolumn{4}{c|}{Methods} \\ \cline{2-5} 
                                    & DON & FNO & CNO & NSO \\ \hline
        Predictions in trained input space & \checkmark & \checkmark &  \checkmark & \checkmark \\ \hline
        Preserves history in trained input space   & \checkmark & \checkmark  & \checkmark & \checkmark \\ \hline
        Interpretable        & $\times$ & $\times$  & $\times$ & \checkmark \\ \hline
        Predictions in novel input space & $\times$ & $\times$  & $\times$ & \checkmark \\ \hline
        Preserves history in novel input space                  & $\times$ & $\times$ &  $\times$ & \checkmark \\ \hline
    \end{tabular}
    }
    \label{tab:attributes}
\end{table}
Another characteristic NSO exhibits is model agnosticism, showcasing limited reliance on prior model assumptions. NSO does not require explicitly including candidate terms governing the dynamics in the library and discovers the dynamics without embedded bias toward the data source. The proposed NSO framework demonstrates two critical attributes for real-world deployment: noise robustness and fidelity invariance. In Experiment 5(a), conducted with 20\% Gaussian noise, and 5(b), with only 20 data points per function, NSO provides high predictive accuracy across all kernels.

As shown in Table~\ref{tbl2}, for the Sine kernel, NSO achieves RMSE values of 2.07e-02 in 5(a) and 3.53e-03 in 5(b), outperforming LASSO (2.41e-02 and 2.41e-02) and SINDy (2.31e-02 and 5.90e-03). For the RBF kernel, NSO achieves RMSEs of 2.50e-02 in 5(a) and 8.90e-03 in 5(b), outperforming LASSO (7.94e-02 and 7.14e-02) and SINDy (3.90e-02 and 1.68e-02). The trend holds for the Matern52 kernel as well, where NSO attains RMSEs of 2.58e-02 (5a) and 8.18e-03 (5b), lower than LASSO (8.75e-02 and 7.88e-02) and SINDy (4.05e-02 and 1.61e-02). These results highlight NSO's ability to maintain low prediction error under noisy and data-scarce conditions. This robustness supports its integration into simulation-based modeling workflows, especially in industrial or scientific domains where generating high-fidelity data is computationally expensive or experimentally infeasible\textemdash such as in sensor-driven systems, experimental measurements, or field observations. Moreover, Table~\ref{tbl3} shows that symbolic models derived through NSO mitigate the trade-off between accuracy and interpretability. Even with few terms (higher threshold values), the models retain high accuracy and exhibit structural resemblance to classical hysteresis formulations such as the Bouc--Wen and Duhem models.

Finally, Table~\ref{tbl4} compares neural operators studied in the paper\textemdash DON, FNO, CNO, and NSO\textemdash across multiple attributes. While conventional operators provide predictions and preserve history within the trained input space, their interpretability and generalization differ significantly. In hysteresis modeling, preserving history\textemdash retaining and utilizing past input states to capture path-dependent behaviors\textemdash is essential, as the system's output depends on the whole input trajectory. NSO exhibits interpretability and predicts and preserves history for unseen input spaces, facilitating superior generalization. Conversely, DON, FNO, and CNO face generalization challenges, as indicated by their performance beyond the training distribution.

\section{Conclusion}
\label{sec7}
This paper introduced a neuro-symbolic operator (NSO) framework for characterizing complex, nonlinear, and history-dependent piezoelectric hysteresis. Traditional neural operator architectures are effective in function-to-function mapping but lack interpretability and exhibit limited generalizability. The proposed NSO framework addressed these challenges by integrating the expressivity of Fourier neural operators with the parsimony of sparse regression. NSO extracted interpretable differential models from black-box neural operator outputs, providing insights into the underlying physics of piezoelectric hysteresis. The discovered white-box models enabled robust and accurate prediction of displacement profiles across various voltage fields, including those unseen during training. Experimental results across diverse hysteresis models demonstrated NSO's superior performance, even with noisy and low-fidelity input data. Furthermore, the framework demonstrated model agnosticism and independence from data source bias. The results demonstrate that our proposed method is accurate, robust, explainable, and generalizable for complex piezoelectric systems essential for designing, monitoring, and maintaining industrial systems.

\bibliographystyle{IEEEtran}
\bibliography{ref}

\end{document}